\documentclass[a4paper,10pt]{article}
\usepackage[english]{babel}
\usepackage[a4paper, top=1.0in, bottom=1.0in, left=0.85in, right=0.85in]{geometry}
\usepackage[english]{babel}
\usepackage{amsmath}
\usepackage{amssymb}
\usepackage{bbm}
\usepackage{mathrsfs}
\usepackage{mathtools}
\usepackage{color}
\usepackage{textcomp}
\usepackage{amsfonts}
\usepackage{dsfont}
\usepackage{lettrine}
\usepackage{units}
\usepackage{comment}
\usepackage{tikz-network}
\usepackage{rotating}
\usepackage{algorithm} 
\usepackage{algpseudocode}
\usepackage{array}
\usepackage{hhline}
\usepackage{braket}
\usepackage{graphicx}
\usepackage[colorlinks,citecolor=blue,urlcolor=blue]{hyperref}

\newcommand{\rvline}{\hspace*{-\arraycolsep}\vline\hspace*{-\arraycolsep}}
\newcommand{\bigzero}{\mbox{\normalfont\Large\bfseries 0}}

\newcommand{\Xinner}[2]{\left\langle #1,\,#2\right\rangle_{\mathcal X}}
\newcommand{\e}{\mathrm e}

\title{Fisher Information, Training and Bias in Fourier Regression Models}

\author{Lorenzo Pastori$^{1}$, Veronika Eyring$^{1,2}$, and Mierk Schwabe$^{1}$} 

\date{
    \small{$^1$ \emph{Deutsches Zentrum für Luft- und Raumfahrt (DLR), Institut f\"ur Physik der Atmosph\"are, Oberpfaffenhofen, Germany} \\
    $^2$ \emph{Institute of Environmental Physics (IUP), University of Bremen, Bremen, Germany}}
}

\begin{document}

\maketitle

\begin{abstract}
    Motivated by the growing interest in quantum machine learning, in particular quantum neural networks (QNNs), we study how recently introduced evaluation metrics based on the Fisher information matrix (FIM) are effective for predicting their training and prediction performance.
    We exploit the equivalence between a broad class of QNNs and Fourier models, and study the interplay between the \emph{effective dimension} and the \emph{bias} of a model towards a given task, investigating how these affect the model's training and performance. 
    We show that for a model that is completely agnostic, or unbiased, towards the function to be learned, a higher effective dimension likely results in a better trainability and performance. On the other hand, for models that are biased towards the function to be learned a lower effective dimension is likely beneficial during training.
    To obtain these results, we derive an analytical expression of the FIM for Fourier models and identify the features controlling a model's effective dimension. This allows us to construct models with tunable effective dimension and bias, and to compare their training. We furthermore introduce a tensor network representation of the considered Fourier models, which could be a tool of independent interest for the analysis of QNN models.
    Overall, these findings provide an explicit example of the interplay between geometrical properties, model-task alignment and training, which are relevant for the broader machine learning community.
\end{abstract}

\section{Introduction}
A popular approach for developing quantum machine learning (QML) models for the analysis of classical data is to use parameterized quantum circuits (PQCs) as trainable machine learning models \cite{PerdomoOrtiz2018,Benedetti2019,Cerezo2022}. In these models, also known as quantum neural networks (QNNs) \cite{Farhi2018}, the classical input data and the trainable parameters are encoded as angles in the quantum gates of the circuit, and the outputs are extracted as expectation values of some observables at the end of the PQC \cite{PerdomoOrtiz2018,Benedetti2019,Cerezo2022,Farhi2018}. The variational nature of QNNs, where the parameters are typically trained in a quantum-classical feedback loop \cite{Cerezo2021}, makes them viable approaches for near-term quantum devices \cite{Bharti2022,Preskill2018}. 

In the last years, several works theoretically investigated how QNNs differ from their classical counterparts, with particular focus in understanding their expressivity \cite{Du2020,Yu2023} and their generalization capability \cite{Caro2021,Banchi2021,Caro2022a,Haug2024}. While no general consensus exists as to whether these variational QML models can offer rigorous advantages compared to classical approaches on classical `real-world' datasets, there is a growing body of literature proposing applications of QNNs in several areas of classical data analysis and machine learning \cite{Chen2022,Hur2022,Suenkel2023,Shen2024,Belis2024,Corli2024,Duneau2024,Aizpurua2024,Sakhnenko2024,Slabbert2024,Pastori2025}. 

In parallel to these investigations, there are also several attempts to develop general evaluation metrics for such architectures. Finding such metrics assessing the quality of a model on a given task, \emph{before} the model has been trained for it, is indeed a key challenge for any machine learning practitioner. While some of the proposed metrics, such as the quantum expressivity \cite{Sim2019,Hubregtsen2021} or entangling capability \cite{Sim2019,Hubregtsen2021} are only relevant in the case of variational quantum models, other metrics, such as the effective dimension \cite{Abbas2021} can be calculated for both classical and quantum models. This latter quantity is the main focus of our work.

The effective dimension (ED), calculated from the Fisher information matrix (FIM), has been recently introduced in a seminal work \cite{Abbas2021} as a measure for the capacity of a model to effectively explore all of its degrees of freedom, i.e., make use of its full parameter space. In \cite{Abbas2021}, the authors not only use the ED for constructing theoretical generalization bounds, but also show numerical examples of QNNs having larger ED than classical networks used for comparison, which they relate to their faster training ability for a chosen learning task. These encouraging results then prompt the question: do models with a high ED \emph{always} have faster training abilities?

\begin{figure*}
    \centering
    \includegraphics[width=\linewidth]{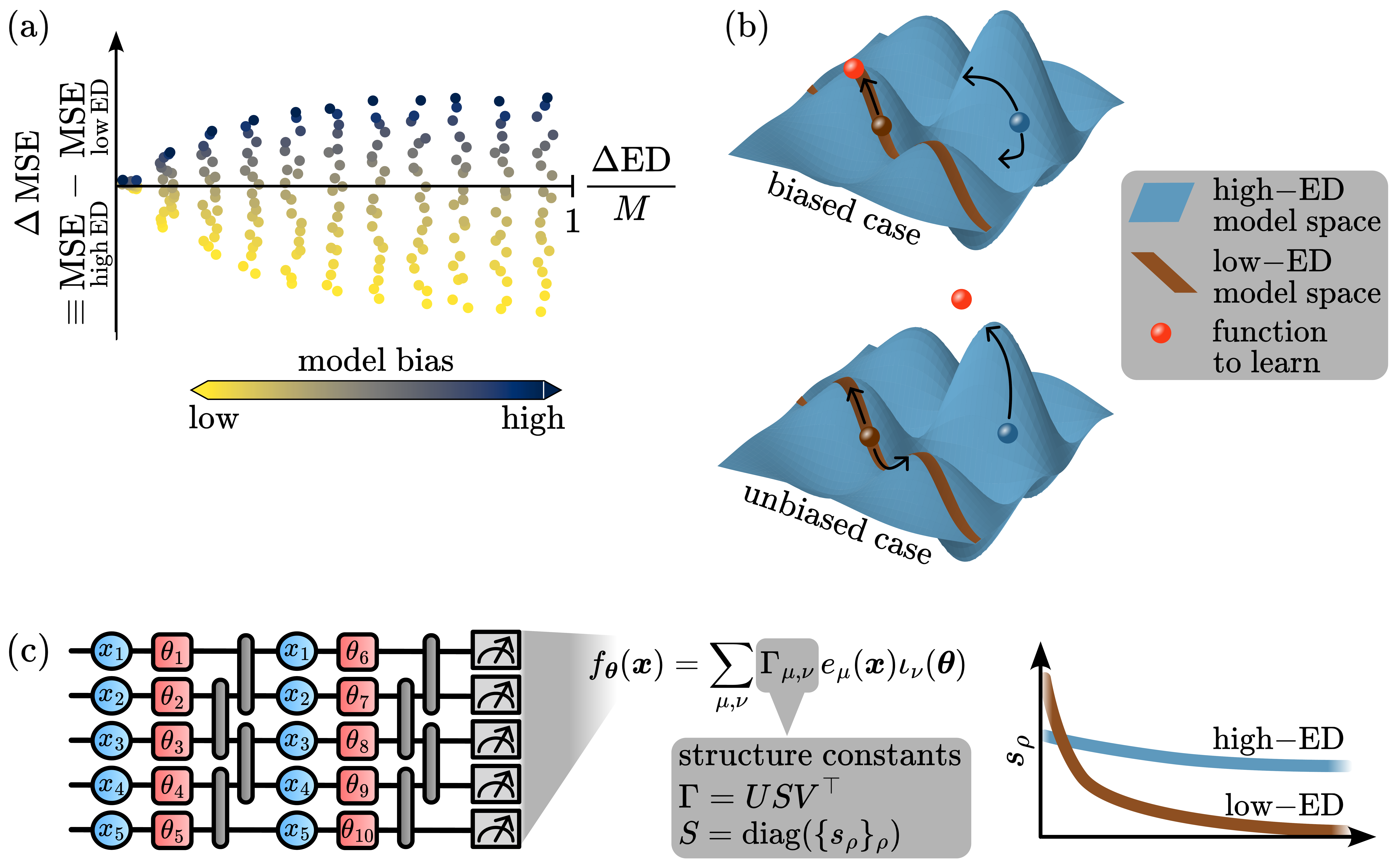}
    \caption{Illustration of main results. (a) Schematic behavior of the difference in the training loss (here the mean squared error --- MSE) $\Delta\mathrm{MSE}$, between models with high and low effective dimension (ED), vs.~the corresponding ED difference $\Delta\mathrm{ED}$ (between models with high and low ED --- normalized by the number of trainable parameters $M$), for models with different bias towards the function to be learned (represented by the color scale). Each point represents the average behavior over several model realizations and training experiments. Models with low ED have better training performance than models with high ED (positive $\Delta\mathrm{MSE}$) in the biased case. The converse is true (negative $\Delta\mathrm{MSE}$) in the unbiased case. (b) Visualization of model spaces in the high- (blue surface) and low-ED (brown line) cases, for biased and unbiased case. Points in these spaces represent functions obtained for specific choices of trainable parameters. In the biased case, the data-generating function (red point) belongs to the model space (to good approximation): a model with low ED (brown point) trained with gradient descent is more likely to converge to the data-generating function since there are effectively less dimensions to explore (only one direction leads to minimizing the loss, as represented by the black arrow). A model with high ED (blue point) is instead more likely to incur in local minima (there are multiple directions, represented by the black arrows, leading to similar loss minimization). In the unbiased case the data-generating function is outside the model space: a model with high ED (blue point) is likely to yield better results, as more directions are available for reaching a better approximation to the data-generating function. (c) Illustration of a QNN and the expansion of its output $f_{\boldsymbol{\theta}}(\boldsymbol{x})$ in the basis functions $e_{\mu}(\boldsymbol{x})$ and $\iota_{\nu}(\boldsymbol{\theta})$ (with $\boldsymbol{x}$ denoting the inputs and $\boldsymbol{\theta}$ the trainable parameters). The coefficient matrix $\Gamma$ (structure constants) can be decomposed in orthogonal matrices $U$ and $V$ and a diagonal matrix $S$ of singular values $s_{\rho}$. The ED of the model is controlled by the decay properties of the singular values: a faster decay results in a lower ED.}
    \label{figapp:Figure_Intro} 
\end{figure*}

In this work, we provide a negative answer to this question. Focusing on models for regression, we show that whether a model with high ED has better training performance compared to one with low ED depends on how \emph{biased} the models are towards the specific regression task. In particular, for models that are completely agnostic, or unbiased, towards the function to be learned (the data-generating function), a high ED likely results in a better trainability. On the other hand, for models that are biased towards the function to be learned a lower ED is beneficial during training. Maybe unsurprisingly, these results quantitatively confirm the intuitive expectation that when the effective space a model can explore is constrained around the task's data-generating function, training the model to a good performance becomes easier. The goal of this work is to explain the basic mechanisms behind this phenomenon, understand them on an analytical level as schematically illustrated in Fig.~\ref{figapp:Figure_Intro}(a) and (b), and provide numerical results supporting the findings. More general, our work hints towards the difficulty, and perhaps the impossibility, of finding a data- and task-independent evaluation metric that can assess a ML model's performance prior to its training.

\subsection{Related literature in quantum and classical ML}
The ED has become a standard information-geometric diagnostic for comparing the capacity of QNNs and related parametrized quantum models, following its introduction in \cite{Abbas2021} as expressivity metric and trainability proxy. It is used as an evaluation metric in several QML studies as a means of comparing different (quantum or classical) models and as guiding principle for model design \cite{Kiss2022,Wilkinson2022,Otgonbaatar2023b,Dragan2022,Monnet2024,Zhu2026,Chen2025,Roseler2026,Mckiernan2026}. This literature establishes ED as a useful capacity metric for QNN architecture assessment. At the same time, these studies also leave open how ED should be interpreted when the architecture is not only more or less expressive, but also more or less aligned with the target task.

A related question has been studied in classical ML under the notions of task-model alignment, target-kernel alignment, spectral bias, and symmetry-induced inductive bias. In particular, in \cite{Canatar2021} a notion of model-task alignment is introduced in the context of kernel regression, as the alignment of the target function with the kernel eigenfunctions. In this context, directly related to comparisons between biased and unbiased model classes are the works of \cite{Amini2022,Wang2024}, which study truncated kernel ridge regression and show that when the target is sufficiently aligned with the retained truncated kernel, the truncated estimator can achieve better learning than full kernel ridge regression. These results further strengthen the intuition that task-aligned model can outperform a less biased or more expressive model when the target lies in the subspace favored by the architecture.

Further classical works connect alignment to training dynamics. For example. in \cite{Shan2022} the authors study neural tangent kernel (NTK) alignment during training and argue that the evolution of the NTK toward the target function provides a mechanism for accelerated learning outside the static-kernel regime. Also, symmetry-constrained models provide an important physics-motivated example of useful task alignment \cite{Elesedy2021}.

These works support the broader view that trainability cannot be assessed from model capacity alone, but depend crucially on how the model's bias matches the structure of the target task. Here we investigate this issue in a controlled class of models that maps naturally to QNNs, studying ED and model-task alignment jointly rather than as separate diagnostics. This allows us to give an analytical account of the interplay between ED, task bias, and trainability, and to construct explicit examples showing when architecture design should prioritize task-aligned bias over raw expressivity.

\subsection{Summary of results}
The findings presented in this paper are obtained by exploiting the equivalence between a broad class of QNNs and Fourier models \cite{GilVidal2020,Schuld2021,Wierichs2022,Casas2023}, which we extend to include the dependence on the trainable QNN parameters. Based on this equivalence, we derive an analytical expression for the Fisher information matrix (FIM), which allows us to identify the relevant features of a Fourier model that control the FIM spectrum, and thereby the effective dimension (ED). Specifically, we derive an explicit relationship between the FIM spectrum and ED of a model and the dimension of the space of functions it has access to. This is schematically illustrated in Fig.~\ref{figapp:Figure_Intro}(c). 

These analytical results enable the practical construction of Fourier models with tunable ED, as well as the construction of models that are more or less biased (as we quantify later in this work) towards a given data-generating function for a regression task. This allows us to study the interplay between ED and bias and their effect on the training ability of a model with numerical examples, with the aforementioned main finding: for models that are biased towards the function to be learned, a lower ED is beneficial during training, whereas for unbiased models a high ED is likely to achieve better performance (see Fig.~\ref{figapp:Figure_Intro}(a) and (b) for a schematic representation).

As a secondary result, in order to numerically investigate larger problem instances (i.e., with larger number of input features and parameters) we introduce a tensor network representation of the structure of Fourier models, called \emph{tensorized} Fourier models. These could be a tool of independent interest for the analysis of QNN models.

\subsection{Organization of the paper}
The remainder of the paper is structured as follows. In Section \ref{sec:models_and_structure} we define the general structure of the regression models we focus on in this work, making the connection with QNNs explicit. In Section \ref{sec:FIM_and_ED} we recap the notions of FIM and ED and, via analytical calculations and numerical examples, we show how these depend on the models' characteristics. In Section \ref{sec:Biased_Unbiased} we give our working definition of model bias, and in Section \ref{sec:tensorized_models} we introduce the concept of tensorized models. With these definitions at hand, in Section \ref{sec:Training_FIM_Bias} we present our results on the interplay between ED and model bias and their effect on training regression models via gradient descent. Finally, we conclude and provide an outlook on possible future investigations in Section \ref{sec:conclusions}.

\section{Methods}

\subsection{Preliminaries: models and structure constants} \label{sec:models_and_structure}
In this section we define the general structure of the regression models that we focus on throughout the paper. Besides providing their general definition, we discuss their relation with functions parameterized by quantum neural networks (QNNs), and introduce the concept of their \emph{structure constants}, which is the central object of our subsequent analysis. For clarity, a list of symbols used throughout this work is provided in Table \ref{table:ListSymbols}.

\subsubsection{Definition of regression models used.} \label{subsec:regr_model_def}
In this work we consider real regression models taking as input a vector $\boldsymbol{x}\in\mathbb{R}^N$, with $N$ denoting the number of input components (features), and parameterized by $M$ trainable (variational) parameters $\boldsymbol{\theta}\in\mathbb{R}^M$. For simplicity in the presentation and derivation of the results, we focus on the case of a single real output, although our results can be easily extended to the case of multi-output models. The general expression of the regression models studied here is
\begin{equation}
    f_{\boldsymbol{\theta}}(\boldsymbol{x})=\sum_{\mu=1}^D c_{\mu}(\boldsymbol{\theta})\,e_{\mu}(\boldsymbol{x}) \,\,.
    \label{eq:general_model_def}
\end{equation}
The functions $e_{\mu}(\boldsymbol{x})\in\mathbb{R}$ are taken to form an orthonormal basis in the $D$-dimensional space of input functions, i.e.,
\begin{equation}
    \mathbb{E}_{\boldsymbol{x}\sim p}\big[e_{\mu}(\boldsymbol{x})e_{\mu'}(\boldsymbol{x})\big]=\int e_{\mu}(\boldsymbol{x})e_{\mu'}(\boldsymbol{x})\,p(\boldsymbol{x})\,\mathrm{d}\boldsymbol{x}=\delta_{\mu,\mu'} \,\,,
    \label{eq:ortho_basis_funs_inputs}
\end{equation}
with $p(\boldsymbol{x})$ the probability density function for the inputs and $\mathbb{E}_{\boldsymbol{x}}$ the expected value over this input distribution. The coefficients $c_{\mu}(\boldsymbol{\theta})\in\mathbb{R}$ encode the dependence on the variational parameters $\boldsymbol{\theta}$. This form of regression model exactly encompasses that of quantum neural networks (QNNs) which, as we discuss later, are known to be Fourier models \cite{GilVidal2020,Schuld2021,Wierichs2022,Casas2023}.

Going a step further, we may expand the coefficients $c_{\mu}(\boldsymbol{\theta})$ in a finite orthonormal basis in the space of parameters' functions, under the assumption that the parameter space is effectively bounded, or that the $c_{\mu}(\boldsymbol{\theta})$ are square-integrable functions. In this case we have
\begin{equation}
    c_{\mu}(\boldsymbol{\theta})=\sum_{\nu=1}^K\Gamma_{\mu,\nu}\,\iota_{\nu}(\boldsymbol{\theta}) \,\,,
    \label{eq:inputbasis_coeffs_expansion}
\end{equation}
where $\iota_{\nu}(\boldsymbol{\theta})\in\mathbb{R}$ are a set of $K$ orthonormal basis functions satisfying
\begin{equation}
    \frac{1}{V_{\Theta}}\int_{\Theta} \iota_{\nu}(\boldsymbol{\theta})\iota_{\nu'}(\boldsymbol{\theta})\,\mathrm{d}\boldsymbol{\theta}=\delta_{\nu,\nu'} \,\,,
    \label{eq:ortho_basis_funs_pars}
\end{equation}
with $\Theta$ denoting the parameter space and $V_{\Theta}$ its volume. As discussed later, this form encompasses the case of QNNs, where both input features and parameters are encoded as rotation angles in quantum gates.

The coefficients $\Gamma_{\mu,\nu}\in\mathbb{R}^{D\times K}$ depend only on the model architecture, and we therefore call them the \emph{structure constants} of the model. The structure constants $\Gamma_{\mu,\nu}$ specify the correlations between the parameter space functions and the input space functions, and are the central object of our analysis throughout this work.

\subsubsection{Fourier regression models and relation to QNNs.} \label{subsec:Fourier_models_and_QNN}
We now relate the form of the regression models introduced above to that of QNNs, and give an explicit expression of the basis functions $e_{\mu}(\boldsymbol{x})$ and $\iota_{\nu}(\boldsymbol{\theta})$ in this specific case. We first briefly recap the concept of a QNN. In the context of regression, a QNN is a function measured from a parameterized quantum circuit (PQC) as 
\begin{equation}
    f_{\boldsymbol{\theta}}(\boldsymbol{x})=\bra{0}\hat{U}^{\dagger}_{\boldsymbol{\theta}}(\boldsymbol{x})\hat{M}\,\hat{U}_{\boldsymbol{\theta}}(\boldsymbol{x})\ket{0}\equiv\bra{\psi_{\boldsymbol{\theta}}(\boldsymbol{x})}\hat{M}\ket{\psi_{\boldsymbol{\theta}}(\boldsymbol{x})} \,\,.
    \label{eq:QNN_output}
\end{equation}
Here, $\hat{M}$ is a given observable, and the output state $\ket{\psi_{\boldsymbol{\theta}}(\boldsymbol{x})}=\hat{U}_{\boldsymbol{\theta}}(\boldsymbol{x})\ket{0}$ is obtained from a quantum circuit described by the unitary $\hat{U}_{\boldsymbol{\theta}}(\boldsymbol{x})$ where inputs and parameters are encoded, applied to a reference state $\ket{0}$. Typical implementations of QNNs involve encoding the input data $\boldsymbol{x}$ and trainable parameters $\boldsymbol{\theta}$ as angles of rotation gates in the quantum circuit. As we show in \ref{app:FourierSeriesQNNs}, the QNN output $f_{\boldsymbol{\theta}}(\boldsymbol{x})$ can be written as a Fourier series in both the inputs and parameters as:
\begin{equation}
    f_{\boldsymbol{\theta}}(\boldsymbol{x})=\sum_{\boldsymbol{\omega}}\sum_{\tilde{\boldsymbol{\omega}}}\tilde{\Gamma}_{\boldsymbol{\omega},\tilde{\boldsymbol{\omega}}}\,\mathrm{e}^{\mathrm{i}\boldsymbol{\omega}\cdot\boldsymbol{x}}\,\mathrm{e}^{\mathrm{i}\tilde{\boldsymbol{\omega}}\cdot\boldsymbol{\theta}} \,\,,
    \label{eq:QNN_fourier_expansion_exp}
\end{equation}
where $\boldsymbol{\omega}=(\omega_1,...,\omega_N)$ with $\omega_n\in\Omega_n$ and $\Omega_n$ the set of Fourier frequencies for the $n$-th input component, $\tilde{\boldsymbol{\omega}}=(\tilde{\omega}_1,...,\tilde{\omega}_M)$ with $\tilde{\omega}_m\in\tilde{\Omega}_m$ and $\tilde{\Omega}_m$ the set of Fourier frequencies for the $m$-th parameter, $\tilde{\Gamma}_{\boldsymbol{\omega},\tilde{\boldsymbol{\omega}}}$ complex constants depending only on the quantum gate generators and the measured observable $\hat{M}$, satisfying $\tilde{\Gamma}_{-\boldsymbol{\omega},-\tilde{\boldsymbol{\omega}}}=\tilde{\Gamma}_{\boldsymbol{\omega},\tilde{\boldsymbol{\omega}}}^*$, and $\cdot$ denoting the scalar product of two vectors. The above expression can equivalently be rewritten in the form of Eqs.~\eqref{eq:general_model_def} and \eqref{eq:inputbasis_coeffs_expansion} with $e_{\mu}(\boldsymbol{x})\equiv e_{(\mu_1,...,\mu_N)}(\boldsymbol{x})=\prod_{n=1}^N e_{\mu_n}^{(n)}(x_n)$ and $\iota_{\nu}(\boldsymbol{\theta})\equiv\iota_{(\nu_1,...,\nu_M)}(\boldsymbol{\theta})=\prod_{m=1}^M\iota_{\nu_m}^{(m)}(\theta_m)$, where
\begin{align}
    & e_{\mu_n}^{(n)}(x_n)\in\mathcal{B}_n=\{1,\,\sqrt{2}\cos(\omega_nx_n),\,\sqrt{2}\sin(\omega_nx_n)\}_{\omega_n\in\Omega_n} \,\,, \label{eq:input_basis_funs_Fourier}\\
    & \iota_{\nu_m}^{(m)}(\theta_m)\in\tilde{\mathcal{B}}_m=\{1,\,\sqrt{2}\cos(\tilde{\omega}_m\theta_m),\,\sqrt{2}\sin(\tilde{\omega}_m\theta_m)\}_{\tilde{\omega}_m\in\tilde{\Omega}_m} \label{eq:param_basis_funs_Fourier}\,\,,
\end{align}
normalized in the interval $[-\pi,\pi]$, and with the structure constants $\Gamma_{\mu,\nu}$ given by suitable real linear combinations of the complex coefficients $\tilde{\Gamma}_{\boldsymbol{\omega},\tilde{\boldsymbol{\omega}}}$. We remark that this Fourier series representation is not limited to the noiseless QNN setting, but is valid also in presence of quantum hardware noise and gate errors, as discussed in \ref{app:FourierSeriesQNNs}.

As an explicit example of the sets of frequencies a QNN can have access to, one can consider the common situation where the inputs are encoded multiple times via the re-uploading technique \cite{Schuld2021,PerezSalinas2020}, and both input features and parameters are encoded as angles of single qubit rotations of the form $\mathrm{e}^{-\mathrm{i}\frac{\phi}{2}\boldsymbol{n}\cdot\hat{\boldsymbol{\sigma}}}$ (with $\phi$ being the feature or parameter to be encoded, $\boldsymbol{n}$ an arbitrary rotation axis and $\hat{\boldsymbol{\sigma}}$ the vector of Pauli matrices). In this case, as we show in \ref{app:FourierSeriesQNNs}, the sets $\Omega_n$ and $\tilde{\Omega}_m$ comprise only integer frequencies and read as $\Omega_n=\{1,...,L\}$ and $\tilde{\Omega}_m=\{1\}$, with $L$ being the number of times the input features $x_n$ are uploaded as gate angles in $\hat{U}_{\boldsymbol{\theta}}(\boldsymbol{x})$, and assuming each parameter $\theta_m$ is encoded only once. We refer the reader to the Supplementary Material for a detailed discussion on how the structure of the entangling operations in a QNN influences the basis functions the model has access to.

This latter example exemplifies the not uncommon situation where the number of Fourier modes for every input feature (and every parameter) is the same. For simplicity, we restrict ourselves to this case for the analysis presented in this work. This results in no loss of generality, since our analytical and numerical results can be easily generalized beyond this case. Throughout the rest of this work, we denote with $d\equiv|\mathcal{B}_n|$ the number of `local' basis functions for the input feature space, and with $\tilde{d}\equiv|\tilde{\mathcal{B}}_m|$ the number of `local' basis functions for the parameter space, which results in $D=d^N$ and $K=\tilde{d}^M$.

\subsubsection{Correlations in structure constants.} \label{subsec:correlations_structure_consts}
We now analyze the information that is contained in the structure constants of the model $\Gamma_{\mu,\nu}$. To do so, we view $\Gamma_{\mu,\nu}$ as elements of a $D\times K$ real matrix $\Gamma$ (we assume $K>D$, since typically we consider models with a number of parameters larger than the number of input features), and consider its singular value decomposition (SVD)
\begin{equation}
    \Gamma=USV^{\top} \,\,,
    \label{eq:SVD_gamma}
\end{equation}
where $U$ is a $D\times D$ real orthogonal matrix satisfying with $U^{\top}U=UU^{\top}=I_D$, $S=\mathrm{diag}(s_1,...,s_D)$ is a $D\times D$ diagonal positive semi-definite matrix (with diagonal ordered as $s_1\geq s_2\geq ... \geq s_D$), and $V$ is a $K\times D$ real matrix with orthonormal columns, i.e., $V^{\top}V=I_D$. After the SVD we can rewrite the model output as
\begin{equation}
    \begin{split}
        f_{\boldsymbol{\theta}}(\boldsymbol{x})&=\sum_{\mu=1}^D \sum_{\nu=1}^K\Gamma_{\mu,\nu}\,e_{\mu}(\boldsymbol{x})\,\iota_{\nu}(\boldsymbol{\theta})\\
        &=\sum_{\rho=1}^D \sum_{\mu=1}^D \sum_{\nu=1}^K U_{\mu,\rho}\,s_{\rho}\,[V^{\top}]_{\rho,\nu}\,e_{\mu}(\boldsymbol{x})\,\iota_{\nu}(\boldsymbol{\theta})\\
        &\equiv\sum_{\rho=1}^D s_{\rho}\,e^U_{\rho}(\boldsymbol{x})\,\iota^V_{\rho}(\boldsymbol{\theta}) \,\,,
        \label{eq:regr_model_gen_expr}
    \end{split}
\end{equation}
where $e^U_{\rho}(\boldsymbol{x})\equiv\sum_{\mu=1}^D U_{\mu,\rho}\,e_{\mu}(\boldsymbol{x})$ and $\iota^V_{\rho}(\boldsymbol{\theta})\equiv\sum_{\nu=1}^K [V^{\top}]_{\rho,\nu}\,\iota_{\nu}(\boldsymbol{\theta})$ constitute new sets of orthonormal basis functions in the input and parameters' space. From the above expression one can easily understand how the singular values $s_{\rho}$ control the correlations between the parameter space and the functions in the input space. In the limiting case of $s_1>0$ and $s_{\rho>1}=0$, any change in the parameters $\boldsymbol{\theta}$ induces a change in $f_{\boldsymbol{\theta}}(\boldsymbol{x})$ along only one component, $e^U_1(\boldsymbol{x})$: in this case, the model's space is effectively one-dimensional. Conversely, if all singular values are equal, the model $f_{\boldsymbol{\theta}}(\boldsymbol{x})$ is effectively able to `explore' all the $D$ orthogonal directions $e^U_{\rho}(\boldsymbol{x})$ in the input space. Thus, the distribution of the $s_{\rho}$, in particular how they decay (i.e., their mutual ratios) indicate how changes in the parameters correlate with the independent directions the model can explore. We therefore call the singular values $s_{\rho}$ the \emph{correlation spectrum} of the model. In the next section we discuss how, perhaps unsurprisingly, the correlation spectrum is related to the notions of Fisher information matrix and effective dimension.

\begin{table}
\centering
\begin{tabular}{|l|l|}
\hline 
Symbol & Used for \\ \hline\hline
$N$ & No. of components of input vector (features) \\ 
\hline
$\boldsymbol{x}=(x_1,...,x_N)$ & Input vector \\ 
\hline
$M$ & No. of trainable parameters \\ 
\hline
$\boldsymbol{\theta}=(\theta_1,...,\theta_M)$ & Vector of trainable parameters \\ 
\hline
$\Theta$ & Parameter space \\ 
\hline
$f_{\boldsymbol{\theta}}(\boldsymbol{x})$ & Regression model (Eq.~\eqref{eq:general_model_def}) \\ 
\hline
$D$ & No. of input basis functions \\ 
\hline
$e_{\mu}(\boldsymbol{x})$ & Input basis functions (Eq.~\eqref{eq:ortho_basis_funs_inputs}) \\ 
\hline
$d$ & Dim. of space `local' to each feature ($D=d^N$) \\ 
\hline
$\mu=(\mu_1,...,\mu_N)$ & Index of input basis functions \\ 
\hline
$K$ & No. of basis functions in param. space \\ 
\hline
$\iota_{\nu}(\boldsymbol{\theta})$ & Basis functions in param. space (Eq.~\eqref{eq:ortho_basis_funs_pars}) \\ 
\hline
$\tilde{d}$ & Dim. of space `local' to each param. ($K=\tilde{d}^M$) \\ 
\hline
$\nu=(\nu_1,...,\nu_M)$ & Index of param. basis functions \\ 
\hline
$\iota^{(m)}_{\nu_m}(\theta_m)$ & `Local' basis functions for param. $\theta_m$ \\ 
\hline
$\Gamma$ & Structure constants (Eq.~\eqref{eq:inputbasis_coeffs_expansion}) \\ 
\hline
$U$ & Left singular vectors of $\Gamma$ (Eq.~\eqref{eq:SVD_gamma}) \\ 
\hline
$V$ & Right singular vectors of $\Gamma$ (Eq.~\eqref{eq:SVD_gamma}) \\ 
\hline
$S$ & Singular values of $\Gamma$ (correlation spectrum --- Eq.~\eqref{eq:SVD_gamma}) \\ 
\hline
$F(\boldsymbol{\theta})$ & Fisher information matrix (FIM --- Eq.~\eqref{eq:FIM}) \\ 
\hline
$\hat{F}(\boldsymbol{\theta})$ & Normalized FIM (Eq.~\eqref{eq:norm_FIM}) \\ 
\hline
$\hat{d}_{\mathrm{eff}}$ & Normalized effective dimension (Eq.~\eqref{eq:norm_eff_dim}) \\ 
\hline
$\beta^{(m)}$ & Local derivative tensor for param. $\theta_m$ (Eq.~\eqref{eq:local_deriv_tensor}) \\ 
\hline
\end{tabular}
\caption{List of most used symbols in this work with explanation.}
\label{table:ListSymbols}
\end{table}

\subsection{Fisher information matrix and effective dimension} \label{sec:FIM_and_ED}
In this section we discuss the notion of Fisher information matrix (FIM) and effective dimension (ED), and relate them to the structure constants of the models introduced in the previous section. We derive an analytic expression of the FIM that makes the relation with the correlation spectrum explicit, and we discuss which features of the correlation spectrum affect the effective dimension.

\subsubsection{Definition of FIM and ED.} \label{subsec:FIM_and_ED_def}
The Fisher information matrix (FIM) is a tool of central importance in the field of information geometry, since it provides a local description of how a parameterized model changes when the parameters it depends on are varied. More specifically, the FIM defines a metric in the space (or manifold) of functions a parameterized model can represent \cite{Amari1985,Amari1997,Pascanu2014}. For the regression models studied in this work, the FIM $F$ is a $M\times M$ positive semi-definite matrix with elements (see \cite{Pennington2018,Amari2019,Karakida2020,Hayase2021,Karakida2021} and the Supplementary Material)
\begin{equation}
    F_{j,k}(\boldsymbol{\theta})=\mathbb{E}_{\boldsymbol{x}}\bigg[\frac{\partial f_{\boldsymbol{\theta}}(\boldsymbol{x})}{\partial\theta_j}\frac{\partial f_{\boldsymbol{\theta}}(\boldsymbol{x})}{\partial\theta_k}\bigg] \,\,.
\label{eq:FIM}
\end{equation}
At a given point $\boldsymbol{\theta}$ in the parameter space $\Theta$, the FIM describes which directions in $\Theta$ the model $f_{\boldsymbol{\theta}}(\boldsymbol{x})$ is more (or less) sensitive to. The eigenvalues of the FIM are indeed a measure of the model sensitivity to parameters changes along the direction defined by the corresponding eigenvectors. If all FIM eigenvalues are approximately equal and larger than zero, all parameters are equally contributing to independent model changes. If instead the FIM spectrum contains several eigenvalues close to zero, the corresponding parameter directions are redundant.
Thus, the FIM encodes information about how a model is effectively able to explore its parameter space. A measure of this ability, i.e., the `size' of the region in model space that a model can effectively explore with its parameters, is given by the effective dimension (ED) introduced in \cite{Abbas2021}. The ED, normalized by the number of parameters $M$, is defined as
\begin{equation}
    \hat{d}_{\mathrm{eff}}=\frac{2\,\mathrm{log}\bigg(\frac{1}{V_{\mathrm{\Theta}}}\int_{\mathrm{\Theta}}\sqrt{\mathrm{det}\Big(I_M+c_{\mathfrak{n}}\hat{F}(\boldsymbol{\theta})\Big)}\mathrm{d}\boldsymbol{\theta}\bigg)}{M\,\mathrm{log}\,c_{\mathfrak{n}}} \,\,,
    \label{eq:norm_eff_dim}
\end{equation}
where $c_{\mathfrak{n}}=\frac{\mathfrak{n}}{2\pi\,\mathrm{log}\,\mathfrak{n}}$ with $\mathfrak{n}$ being the number of input data samples, $I_M$ the $M$-dimensional identity matrix, and $\hat{F}(\boldsymbol{\theta})$ being the normalized FIM defined as
\begin{equation}
    \hat{F}(\boldsymbol{\theta})=\frac{M}{\frac{1}{V_{\mathrm{\Theta}}}\int_{\mathrm{\Theta}}\mathrm{tr}\big(F(\boldsymbol{\theta})\big)\mathrm{d}\boldsymbol{\theta}}\,F(\boldsymbol{\theta}) \,\,.
    \label{eq:norm_FIM}
\end{equation}
The normalized ED $\hat{d}_{\mathrm{eff}}$ is bounded in $[0,1]$, and is computed from averages over the parameter space, hence depending solely on the architecture choices and the input distribution. Note that this definition of the ED as global average gives a global view of the ability of a model to explore its parameter space, which is especially relevant in the early stages of training when the parameters are typically sampled randomly. We note in passing that also a local definition of the ED exists \cite{Abbas2021b}, which probes the vicinity of a point and can be therefore monitored during the training to infer the local geometrical properties during optimization. In the next sections, we uncover the relation between the correlation spectrum introduced before and the FIM, which directly affects the ED defined above.

\subsubsection{FIM and correlation spectrum.} \label{subsec:FIM_diagr_expr}
In Section \ref{subsec:correlations_structure_consts}, we discussed how the distribution of the correlation spectrum $\{s_{\rho}\}_{\rho}$ controls how many independent directions (in the space of input functions) the model can explore when the parameters are changed. Since the FIM and the effective dimension also capture a notion of effective degrees of freedom of the model, it is natural to expect a strong relation between these and the properties of the correlation spectrum. In this section, we explicitly uncover this relation and show that the FIM spectral properties are indeed mostly controlled by the correlation spectrum.

To this end, we start by expressing the FIM elements in terms of the structure constants $\Gamma$ and the related correlation spectrum and singular vectors. It is easy to show that the FIM elements can be expressed as
\begin{equation}
     F_{j,k}(\boldsymbol{\theta})=\sum_{\rho}s_{\rho}^2\,\sum_{\nu,\nu'}\iota_{\nu}(\boldsymbol{\theta})\,\iota_{\nu'}(\boldsymbol{\theta})\sum_{\kappa_j,\kappa_k'}\beta^{(j)}_{\kappa_j,\nu_j}[V^{\top}]_{\rho,(\nu_1...\kappa_j...\nu_M)}\beta^{(k)}_{\kappa_k',\nu_k'}[V^{\top}]_{\rho,(\nu_1'...\kappa_k'...\nu_M')} \,\,,
     \label{eq:FIM_analytic_expr}
\end{equation}
where we introduce the (local) derivative tensor $\beta^{(j)}$ for expressing the derivatives of the basis functions $\iota_{\nu}(\boldsymbol{\theta})$ as linear combinations of basis functions, i.e.,
\begin{equation}
    \frac{\partial\iota_{\nu}(\boldsymbol{\theta})}{\partial\theta_j}=\frac{\partial\iota^{(j)}_{\nu_j}(\theta_j)}{\partial\theta_j}\prod_{m\neq j}\iota^{(m)}_{\nu_m}(\theta_m)=\bigg(\sum_{\kappa_j}\beta^{(j)}_{\nu_j,\kappa_j}\,\iota^{(j)}_{\kappa_j}(\theta_j)\bigg)\prod_{m\neq j}\iota^{(m)}_{\nu_m}(\theta_m) \,\,.
    \label{eq:local_deriv_tensor}
\end{equation}
We refer the reader to the Supplementary Material for a derivation. From Eq.~\eqref{eq:FIM_analytic_expr} we can already recognize the dependence on the squared singular values $s_{\rho}^2$. These, in combination with Eq.~\eqref{eq:norm_FIM}, tell us that the normalized FIM $\hat{F}$, hence the ED, is independent of the value of $\mathrm{tr}(S^2)$. Therefore, the ED is sensitive only to the mutual relationship of the values of $s_{\rho}^2$, i.e., the decay properties of $S^2$, and not to the overall magnitude $\mathrm{tr}(S^2)$. In other words, the FIM and ED are a measure of the dimensionality of the space of functions that the model has access to, captured by how many values $s_{\rho}^2$ are effectively different from zero. These observations give us a practical way of designing models with tunable ED, which we will use in our numerical analysis of the training dynamics presented in Section \ref{sec:Training_FIM_Bias}. We now substantiate these statements by stating the following properties of the FIM.\\

\textbf{Property 1.} In the regime $M>D$ (which can be interpreted as an overparameterized regime as described in \cite{Larocca2023}), the rank of the FIM is upper-bounded as follows:
\begin{equation}
    \mathrm{rank}(F(\boldsymbol{\theta}))\leq D \,\,.
    \label{eq:FIM_prop_1}
\end{equation}

\textbf{Property 2.} In expectation over random realizations of $V\in\mathrm{O}(K)$ (with $\mathrm{O}(K)$ the group of $K\times K$ orthogonal matrices) and $\boldsymbol{\theta}\in\Theta$, one has:
\begin{equation}
    \begin{split}
        &\mathbb{E}\big[F_{j,k}\big]\in
        \begin{cases}
        \mathcal{O}(1)\,\mathrm{tr}(S^2)\;,\quad\mathrm{for}\;j=k\\
        \mathcal{O}(K^{-1})\,\mathrm{tr}(S^2)\;,\quad\mathrm{for}\;j\neq k
        \end{cases}\\
        &\mathrm{Var}\big[F_{j,k}\big]\in
        \mathcal{O}(1)\,\mathrm{tr}(S^4) \,\,.
    \end{split}
    \label{eq:FIM_prop_2}
\end{equation}

The derivation of these properties is sketched in \ref{app:FIM_derivations}. From Property 1, together with the observation that the ED is bounded by the maximal rank of the FIM \cite{Abbas2021}, we can conclude that in the regime $M>D$ the ED is upper-bounded by $D$. This establishes a direct connection between the ED and the dimension $D$ of the space of input functions available to the model. Property 2 further strengthens this connection by showing that the decay properties of $S^4$, encoded in the value of $\mathrm{tr}(S^4)$, indeed control the FIM spectrum. We now explain this more in detail.

As observed previously, $\mathrm{tr}(S^2)$ is a normalization factor that drops in the calculation of the ED. The term that has non-trivial effects on the FIM and the ED is $\mathrm{tr}(S^4)$ controlling the variance, which encodes information on the decay properties of the correlation spectrum. Assuming (without loss of generality) a correlation spectrum normalized such that $\mathrm{tr}(S^2)=1$, $\mathrm{tr}(S^4)$ can be interpreted as a \emph{purity}: a completely flat correlation spectrum corresponds to the minimum value of the purity (i.e., $1/D$), whereas $\mathrm{tr}(S^4)=1$ if $s_1=1$ and $s_{\rho>1}=0$. To understand how the value of $\mathrm{tr}(S^4)$ influences the spectral properties of the FIM, we first observe  that the expected value of the FIM is approximately diagonal with diagonal elements having the same values and off-diagonal elements suppressed as $\mathcal{O}(K^{-1})$. Thus, in absence of statistical fluctuations, the spectrum of the FIM would be flat, which would correspond to a high ED. This is approximately the case when the variance is small, i.e., when the correlation spectrum $S$ is flat and $\mathrm{tr}(S^4)$ is small. Conversely, when the correlation spectrum $S$ is not flat and $\mathrm{tr}(S^4)$ is large (i.e., approaching one), $\mathrm{Var}\big[F_{j,k}\big]$ introduces non-negligible statistical fluctuations that make the FIM spectrum deviate from the flat case, which in turn decreases the ED. Thus, with this analysis we identify in the correlation spectrum, and in particular in its decay properties partly captured by $\mathrm{tr}(S^4)$, the key factor controlling the FIM and the ED in regression models. This insight is of central importance in our numerical analysis presented in the next sections. \\

\label{subsubsec:ED_numerics_random_models}
\begin{figure*}
    \centering
    \includegraphics[width=\linewidth]{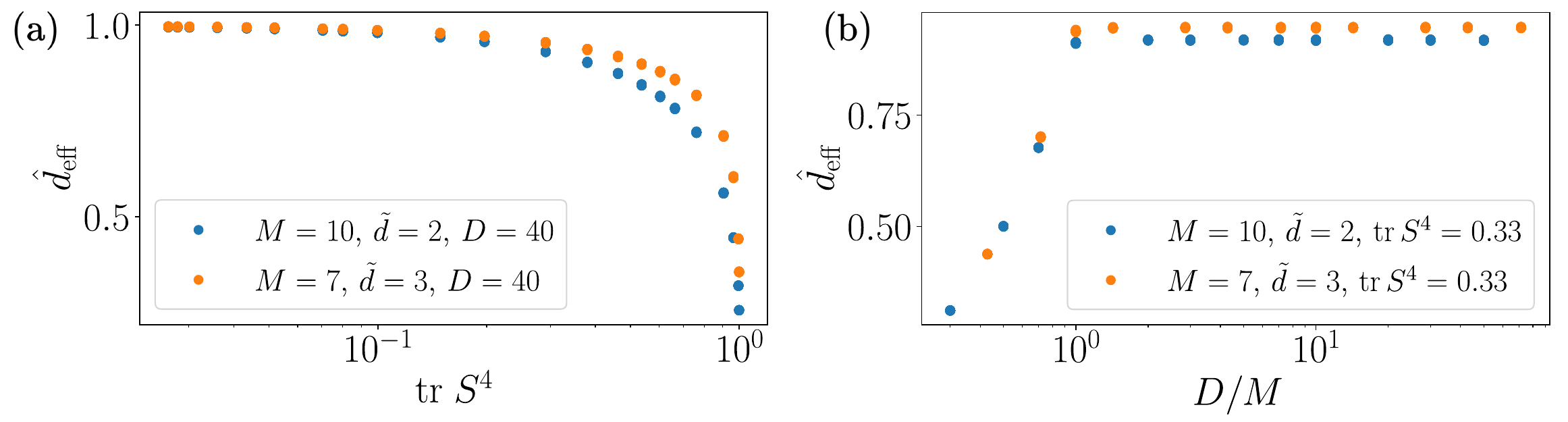}
    \caption{(a) Scaling of normalized ED with the purity $\mathrm{tr}(S^4)$ of the correlation spectrum. (b) Scaling of normalized ED with the ratio $D/M$. Each point corresponds to a random model realization, i.e., a random $\Gamma$ uniformly drawn from $[-1,+1]^{D\times K}$. For every value of $\mathrm{tr}(S^4)$ and $D/M$, $50$ model realizations are drawn (the points are on top of each others). The normalized ED is computed using Eq.~\eqref{eq:norm_eff_dim}, with $150$ parameters samples for estimating the normalized FIM. Here, $\tilde{d}=2$ refers to $\tilde{\mathcal{B}}_m=\{\sqrt{2}\cos\theta_m,\sqrt{2}\sin\theta_m\}$, while $\tilde{d}=3$ refers to $\tilde{\mathcal{B}}_m=\{1,\sqrt{2}\cos\theta_m,\sqrt{2}\sin\theta_m\}$ in Eq.\eqref{eq:param_basis_funs_Fourier}, for all $m=1,...,M$.}
    \label{fig:ED_fullmodels_scalings} 
\end{figure*}
To corroborate our analytical findings, we show numerical results on the dependence of the ED on different model characteristics in Fig.~\ref{fig:ED_fullmodels_scalings}. Specifically, we calculate the normalized ED $\hat{d}_{\mathrm{eff}}$ for several random model realizations, i.e., random realizations of the structure constants $\Gamma$ (uniformly drawn from $[-1,+1]^{D\times K}$), and investigate how $\hat{d}_{\mathrm{eff}}$ changes with $\mathrm{tr}(S^4)$ and the input functions' space dimension $D$. In panel (a) we show the dependence of $\hat{d}_{\mathrm{eff}}$ on the purity of the correlation spectrum $\mathrm{tr}(S^4)$. As expected, $\hat{d}_{\mathrm{eff}}$ decreases with $\mathrm{tr}(S^4)$ increasing towards its maximum value $1$. In order to change $\mathrm{tr}(S^4)$, an exponential decay with variable decay rate is imposed to the singular values, keeping them normalized to $\mathrm{tr}(S^2)=1$ (which has no effect on the ED). In panel (b) we show the dependence of $\hat{d}_{\mathrm{eff}}$ on the ratio $D/M$. For $D<M$, $\hat{d}_{\mathrm{eff}}$ increases with increasing $D$ until it saturates to its maximal value for $D\geq M$. For $D\geq M$, $\hat{d}_{\mathrm{eff}}$ is independent of $D$ and largely controlled by $\mathrm{tr}(S^4)$. The increase of $\hat{d}_{\mathrm{eff}}$ for $D<M$ is due to the fact in this regime, the maximal ED of the model is upper-bounded by $D$, hence the model has more parameters than the input basis functions (i.e., independent directions in model space) it has access to.
Further numerical results are presented in the Supplementary Material and confirm indeed that $\mathrm{tr}(S^4)$ is the key factor controlling the ED.

\subsubsection{FIM and neural tangent kernel.}
\label{subsubsec:FIM_NTK}
To conclude our analysis of the FIM and establish a connection to the training behavior, we note that there is a strong relation between the FIM and the neural tangent kernel (NTK), which is defined as \cite{Jacot2018} $K_{\boldsymbol{\theta}}(\boldsymbol{x},\boldsymbol{x}')=\nabla_{\boldsymbol{\theta}}f_{\boldsymbol{\theta}}(\boldsymbol{x})^\top \nabla_{\boldsymbol{\theta}}f_{\boldsymbol{\theta}}(\boldsymbol{x}')$, where $\nabla_{\boldsymbol{\theta}}f_{\boldsymbol{\theta}}(\boldsymbol{x})$ is the $M$-dimensional gradient of the model. 
The NTK can be used to analyze how gradient-based training changes function values: it provides a local description of which modes (i.e., directions in the input function space) are learned faster or slower under gradient flow \cite{Jacot2018,Fort2020,Loo2022,Amini2022}. In particular, the eigenvalues of the NTK correspond to the rates at which the corresponding modes are learned \cite{Jacot2018}. As shown in the Supplementary Material, the non-zero spectrum of the NTK coincides with that of the FIM $F(\boldsymbol{\theta})$. Furthermore, on average over the parameter space and random realizations of $V\in\mathrm{O}(K)$, eigenvectors and eigenvalues of the NTK coincide with the left singular vectors $U$ and singular values $S^2$ of the model's structure constants. These observations establish the relationship between the NTK eigen-decomposition, the ED (controlled by the FIM), and the bias (defined in the next section, controlled by the model's structure constants). The NTK gives us a way of analytically understanding how different ED and model bias influence the training dynamics, which we elaborate on in the Supplementary Material.

\subsection{Biased and unbiased regression models} \label{sec:Biased_Unbiased}
\begin{figure*}
    \centering
    \includegraphics[width=\linewidth]{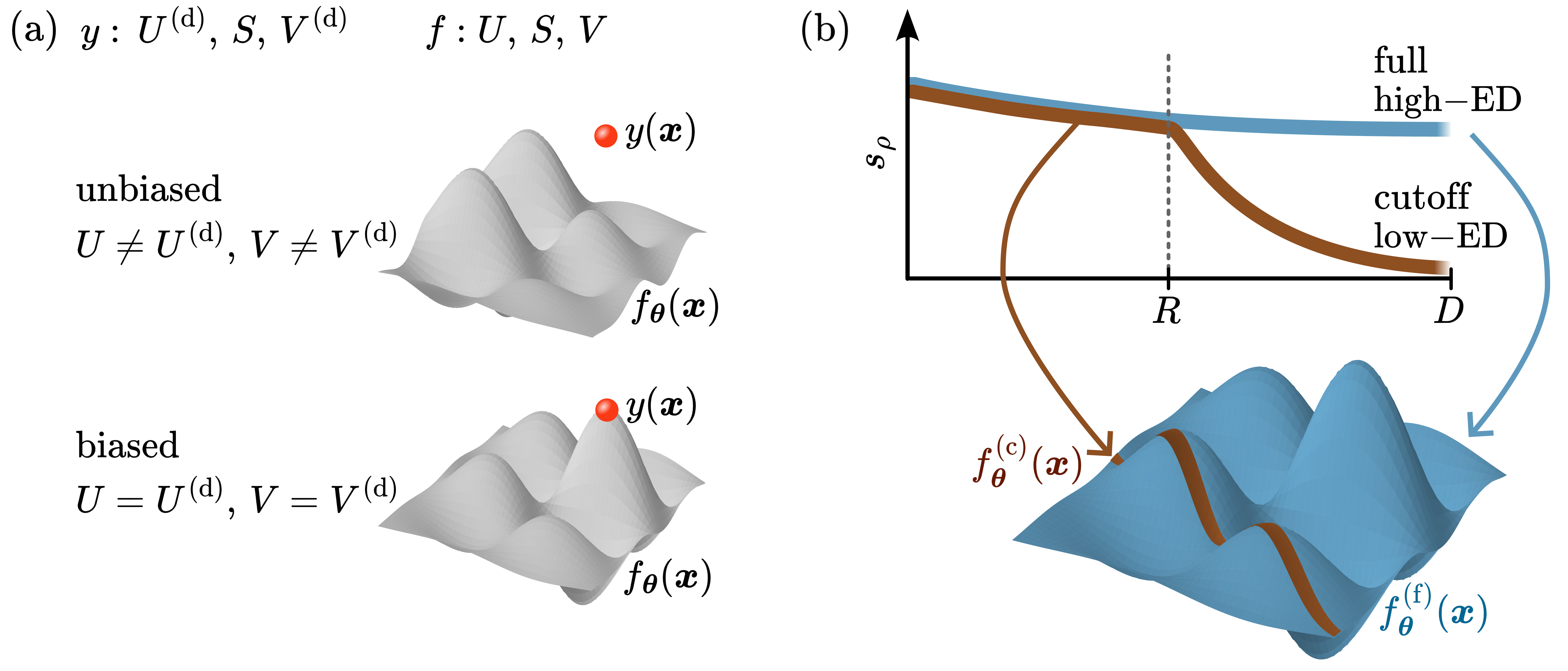}
    \caption{(a) Schematic illustration of the construction of biased and unbiased models. The data-generating function $y$ is specified by matrices $U^{\mathrm{(d)}}$ and $V^{\mathrm{(d)}}$, and the model $f$ by $U$ and $V$, with both $y$ and $f$ having the same correlation spectrum $S$. In the unbiased case, $y$ (represented by the red dot) lies outside the space of functions accessible to $f_{\boldsymbol{\theta}}$ (represented by the gray surface), whereas in the biased case $y$ belongs to that space. (b) Construction of models with tunable ED. Full models $f^{\mathrm{(f)}}_{\boldsymbol{\theta}}(\boldsymbol{x})$ with no imposed decay in the correlation spectrum $s_{\rho}$, as illustrated by the blue line, have high ED and therefore can access a larger functions' space, represented by the blue surface. Cutoff models $f^{\mathrm{(c)}}_{\boldsymbol{\theta}}(\boldsymbol{x})$ with decaying correlation spectrum $s_{\rho}$, as illustrated by the brown line, have low ED and have access to more restricted functions' space, represented by the brown surface.}
    \label{fig:Figure_Bias_Unbias} 
\end{figure*}
In this section we introduce the concept of biased and unbiased regression models used in this work, and we provide a simple recipe for constructing models where the bias and the effective dimension can be tuned at will. The main idea behind our definition of biased and unbiased model is the following:
\begin{itemize}
    \item A model $f_{\boldsymbol{\theta}}(\boldsymbol{x})$ is \emph{biased} towards the data-generating function $y(\boldsymbol{x})$ if there exists a parameter configuration $\boldsymbol{\theta}^*$ for which $f_{\boldsymbol{\theta}^*}(\boldsymbol{x})=y(\boldsymbol{x})$.
    \item If there is no configuration $\boldsymbol{\theta}^*$ for which $f_{\boldsymbol{\theta}^*}(\boldsymbol{x})=y(\boldsymbol{x})$ then the model is \emph{unbiased}.
\end{itemize}
The goal behind the definition of ``model bias" used in this work is to have a controllable way of mimicking the situation where some features of the data-generating function are at least approximately known, and the model is designed such that it can incorporate, or is constrained to represent, these known features. For example, if the task involves predicting some signal coming from a physical process for which the frequency range is known from physical principles or estimations, one could design a model constrained to output signals within the given frequency range: as we remark in the end of this section (and explicitly showcase in the Supplementary Material), this is indeed something that can be achieved in QNNs, via data re-uploading and by tuning their entanglement structure.

For an explicit construction of biased and unbiased models to be used in our numerical experiments, we consider a data-generating function $y(\boldsymbol{x})$ of the form
\begin{equation}
    y(\boldsymbol{x})=\sum_{\mu=1}^{D}\sum_{\nu=1}^{K}e_{\mu}(\boldsymbol{x})\,\iota_{\nu}(\boldsymbol{\theta}^*)\sum_{\rho=1}^{R}s_{\rho}\,U^{\mathrm{(d)}}_{\mu,\rho}\,\big[V^{\mathrm{(d)}\,\top}\big]_{\rho,\nu}\,\,,
    \label{eq:data_gen_function}
\end{equation}
with $\boldsymbol{\theta}^*$ a given parameter configuration, $R<D$ and with $V^{\mathrm{(d)}}$ constructed in order to satisfy the property $\sum_{\nu}V^{\mathrm{(d)}}_{\nu,\rho}\,\iota_{\nu}(\boldsymbol{\theta}^*)=0$ for $\rho=R+1,...,K$.

Any model of the form of Eq.~\eqref{eq:regr_model_gen_expr} with structure constants chosen independently of $U^{\mathrm{(d)}}$ and $V^{\mathrm{(d)}}$, with high probability does not exactly encompass $y(\boldsymbol{x})$ for any choice of the parameters: any such model is agnostic to the form of $y(\boldsymbol{x})$, and is referred to as \emph{unbiased}. Instead, a model $f^{\mathrm{(f)}}_{\boldsymbol{\theta}}(\boldsymbol{x})$ specified by the structure constants
\begin{equation}
    \Gamma^{\mathrm{(f)}}_{\mu,\nu}=\sum_{\rho=1}^{D}U^{\mathrm{(d)}}_{\mu,\rho}\,s_{\rho}\,\big[V^{\mathrm{(d)}\,\top}\big]_{\rho,\nu}
    \label{eq:full_regr_model_gamma}
\end{equation}
satisfies $f^{\mathrm{(f)}}_{\boldsymbol{\theta}^*}(\boldsymbol{x})=y(\boldsymbol{x})$ by construction, and is therefore called \emph{biased}. This is schematically illustrated in Fig.~\ref{fig:Figure_Bias_Unbias}(a). Importantly, one can define several such biased models with different properties of $s_{\rho}$, hence different effective dimensions, by choosing different values for $s_{\rho>R}$, which have no influence on the model prediction for $\boldsymbol{\theta}=\boldsymbol{\theta}^*$. Consider for example the following structure constants
\begin{equation}
    \Gamma^{\mathrm{(c)}}_{\mu,\nu}=\sum_{\rho=1}^{R}U^{\mathrm{(d)}}_{\mu,\rho}\,s_{\rho}\,\big[V^{\mathrm{(d)}\,\top}\big]_{\rho,\nu}+\sum_{\rho=R+1}^{D}U^{\mathrm{(d)}}_{\mu,\rho}\,\mathrm{e}^{-\frac{\rho-R}{\xi}}s_{\rho}\,\big[V^{\mathrm{(d)}\,\top}\big]_{\rho,\nu} \,\,,
    \label{eq:cut_regr_model_gamma}
\end{equation}
with $\xi$ a positive decay rate. Since $\xi$ induces a decay in the correlation spectrum, and hence a higher spectral purity $\mathrm{tr}(S^4)$, a model specified by $\Gamma^{\mathrm{(c)}}$ will have on average a lower ED than a model specified by $\Gamma^{\mathrm{(f)}}$. We refer to models constructed as $\Gamma^{\mathrm{(f)}}$ as \emph{full} models, whereas models constructed as $\Gamma^{\mathrm{(c)}}$, with an imposed decay $\xi>0$ and a lower ED, are referred to as \emph{cutoff} models. This is schematically illustrated in Fig.~\ref{fig:Figure_Bias_Unbias}(b). Tuning $\xi$ gives us a way of tuning the ED, and hence the difference $\hat{d}_{\mathrm{eff}}^{\mathrm{full}}-\hat{d}_{\mathrm{eff}}^{\mathrm{cut}}$, which we use in the numerical experiments presented in the next section. Loosely speaking, one can understand a biased model with high ED as a model incorporating knowledge of the target function, while a biased model with low ED is \emph{constrained} to that knowledge without deviating from it (in a functional sense).

\emph{Partially biased} models, which only approximately encompass $y(\boldsymbol{x})$, can be constructed adding a small perturbation to $V^{\mathrm{(d)}}$ in the data-generating function, i.e., replacing $V^{\mathrm{(d)}}$ in Eq.~\eqref{eq:data_gen_function} with $V^{\mathrm{(d)}}_{\epsilon}$ obtained by adding a suitably chosen random perturbation of strength $\epsilon$ to its elements, as we discuss in \ref{app:biased_unbiased_models}. In this way, $\Gamma^{\mathrm{(f)}}$ and $\Gamma^{\mathrm{(c)}}$ only approximately encompass the newly constructed data-generating function. We quantify the deviation from $y(\boldsymbol{x})$ using the following parameter
\begin{equation}
     \delta_{\mathrm{data}}=\big\Vert S\big(V^{\mathrm{(d)}\,\top}-V_{\epsilon}^{\mathrm{(d)}\,\top}\big)\big|\boldsymbol{\iota}(\boldsymbol{\theta}^*)\big)\big\Vert_1 \,\,,
     \label{eq:partial_bias_2}
\end{equation}
with $\big|\boldsymbol{\iota}(\boldsymbol{\theta}^*)\big)$ being the $K$-dimensional vector with components $\iota_{\nu}(\boldsymbol{\theta}^*)$.

While this definition of bias is based on the structure constants, we remark that in the context of QNNs these depend on the model architecture (i.e., the form of the unitary $\hat{U}_{\boldsymbol{\theta}}(\boldsymbol{x})$), which can be tuned via the structure of the PQC, ultimately providing a way of biasing the model towards certain types of functions. For example, simply adding new rotation gates re-encoding data features changes the accessible input function space adding higher harmonics to the Fourier series. This enables the model to learn such type of functions, and mathematically amounts to a change in the structure constants $\Gamma_{\mu,\nu}$, now with support on larger input or parameter spaces, which can be captured by our definition of bias. We provide further examples of this mechanism, with explicit calculations of the model structure constants for instances of QNNs, in the Supplementary Material: our analysis highlights specific circuit design choices that can be used to bias the QNN outputs, provided one has some knowledge of the data-generating function.

Finally, we comment here on the relation between our definition of bias and the more general concept of inductive bias in ML. Inductive biases typically result from architectural choices (approximately) restricting the model function to have specific symmetries or live on specific manifolds. Our construction defines model bias in a stricter form, i.e., as an Euclidean distance between model and target function space. This definition gives us the advantage of having a tunable parameter for conducting our numerical experiments. Including functional symmetries (such as translations or mirror symmetries) can in principle also be done at the level of the structure constants, bringing us closer to the more general definition of inductive bias. However, in order to showcase the basic functional mechanisms of the ED-bias interplay, we decide to leave this as a possible direction of future work.

\subsection{Tensorized models} \label{sec:tensorized_models}
We now briefly introduce the concept of \emph{tensorized} models, which allow us to circumvent the exponential costs (in $N$ and $M$, since $D=d^N$ and $K=\tilde{d}^M$) of constructing and storing the full $\Gamma$ in our simulations. This enables the numerical study of problem instances with larger number of features $N$ and of parameters $M$. The main idea is to decompose the structure constants $\Gamma$ as a tensor network (TN) \cite{Verstraete2008,Orus2014,Ran2020} with a finite bond dimension $\chi$. A TN decomposition of $\Gamma$ is enabled by the structure of the input and parameter functions' spaces, which are constructed as tensor products of the spaces `local' to each input feature $x_n$ and parameter $\theta_m$, spanned by the local basis functions $e^{(n)}_{\mu_n}(x_n)$ and $\iota^{(m)}_{\nu_m}(\theta_m)$. To construct a TN decomposition of $\Gamma$, we consider the following approximation
\begin{equation}
    \Gamma_{\mu,\nu}\approx\sum_{\rho=1}^D\sum_{\sigma=1}^{\chi}\;\underbrace{U_{\mu,\rho}\,T_{\rho,\sigma}}_{\text{ortho. rotation + isometry}}\,\underbrace{s_{\sigma}}_{\text{\,corr. spectrum\,}}\underbrace{\big[V^{\top}\big]_{\sigma,\nu}}_{\text{map from param. space}} \,\,,
    \label{eq:gamma_tensor_models}
\end{equation}
which differs from Eq.~\eqref{eq:SVD_gamma} by the presence of an isometry $T$, which is a linear mapping from the $D$-dimensional input functions' space to a $\chi$-dimensional reduced space, building an internal TN representation of the input functions' space. 

We decompose the matrices $U$, $T$ and $V$ as products of low-rank tensors as follows. The matrix $V$ containing the right-singular values of $\Gamma$ is expressed as a tensor-train (also known as matrix product state) \cite{Oseledets2011,Schollwoeck2011}, the orthogonal matrix $U$ containing the left-singular values of $\Gamma$ as an orthogonal matrix product operator (MPO) \cite{Pirvu2010,Hubig2017,Styliaris2025}, and the isometry $T$ can be as a tree tensor network (TTN) \cite{Shi2006,Tagliacozzo2009,Murg2010}. We note that the TN decompositions used here are not the only option, and different ones are possible. In \ref{app:tensorized_models} we provide the explicit expressions of these decompositions, together with their diagrammatic representation, the conditions the individual tensors need to fulfill in order to respect the orthogonality of the decomposed matrices, and details on how we practically generate random instances of those. In the Supplementary Material we show how to adapt the procedure described in the previous section to generate biased and unbiased tensorized models. There, we also numerically check that the bond dimension $\chi$ does not have a significant effect on the effective dimension. This means that also for tensorized models the ED is controlled by the decay property of the correlation spectrum, which allows us to use the same procedure as described before for tuning the models' ED.

\section{Results} \label{sec:Training_FIM_Bias}
In this section we present our main results on the effects of the ED on the training of regression models with gradient-based methods, and the interplay with the model's bias towards the regression task at hand. 
The regression tasks investigated here consist in training the parameters $\boldsymbol{\theta}$ of a regression model $f_{\boldsymbol{\theta}}(\boldsymbol{x})$, of the form introduced before, to learn a data-generating function $y(\boldsymbol{x})$. We consider the situation where we have $\mathfrak{n}_{\mathrm{train}}$ input-output pairs $\{\boldsymbol{x}_i,y(\boldsymbol{x}_i)\}_{i=1,...,\mathfrak{n}_{\mathrm{train}}}$ that we use for training the model, using the mean squared error (MSE) as loss function
\begin{equation}
    \mathrm{MSE}=\frac{1}{\mathfrak{n}_{\mathrm{train}}}\sum_{i=1}^{\mathfrak{n}_{\mathrm{train}}}\big(f_{\boldsymbol{\theta}}(\boldsymbol{x}_i)-y(\boldsymbol{x}_i)\big)^2 \,\,.
    \label{eq:MSE}
\end{equation}
The numerical results presented in this section are obtained using Fourier regression models, i.e., with input and parameter basis functions $e^{(n)}_{\mu_n}(x_n)$ and $\iota^{(m)}_{\nu_m}(\theta_m)$ given by Eqs.~\eqref{eq:input_basis_funs_Fourier} and \eqref{eq:param_basis_funs_Fourier}. For simplicity, we restrict to the case where the `local' basis sets $\mathcal{B}_{n}$ and $\tilde{\mathcal{B}}_{m}$, as well as the `local' frequency sets $\Omega_n$ and $\tilde{\Omega}_m$, are independent of the feature index $n$ and the parameter index $m$. The $\mathfrak{n}_{\mathrm{train}}$ input points are chosen uniformly in $[-\pi,\pi]^N$, and the models are then trained with the Adam optimizer \cite{Kingma2014}.

In order to study the interplay of model bias and ED and their effects on training in a statistically sound manner, we perform several training experiments with randomly drawn data-generating function $y(\boldsymbol{x})$ and structure constants $\Gamma$. Specifically, for chosen dimensions $D$ and $K$ (fixed by the choice of $N$, $d$, $M$ and $\tilde{d}$), we draw random instances of $y(\boldsymbol{x})$, and many random instances of models, specified by $\Gamma$, with different degree of bias towards $y(\boldsymbol{x})$. For any given degree of bias, we train several random instances of full models (Eq.~\eqref{eq:full_regr_model_gamma}) and cutoff models (Eq.~\eqref{eq:cut_regr_model_gamma}), in order to compare the training dynamics of models with higher and lower ED, respectively. Furthermore, every model instance is trained several times starting from random parameter configurations: this probes the robustness of the training dynamics against different random starting points, providing an averaged picture of the severity of local minima in the optimization landscape. To visualize the comparison between models with higher and lower ED, for any given degree of bias we consider the minimum MSE attained during training as a proxy for the training quality, denoted with $\mathrm{MSE}_{\mathrm{min}}^{\mathrm{full}}$ and $\mathrm{MSE}_{\mathrm{min}}^{\mathrm{cut}}$ for full and cutoff models, respectively, and study the difference
\begin{equation}
    \Delta_{\mathrm{f-c}}\mathrm{MSE}_{\mathrm{min}}=\mathrm{MSE}_{\mathrm{min}}^{\mathrm{full}}-\mathrm{MSE}_{\mathrm{min}}^{\mathrm{cut}} \,\,,
    \label{eq:DeltaMSE_fullminuscut}
\end{equation}
as a function of the difference in the ED between full and cutoff models, i.e., $\hat{d}_{\mathrm{eff}}^{\mathrm{full}}-\hat{d}_{\mathrm{eff}}^{\mathrm{cut}}$. A positive value of $\Delta_{\mathrm{f-c}}\mathrm{MSE}_{\mathrm{min}}$ implies that the full model (with higher ED) trains to a higher MSE compared to the cutoff one, i.e., the model with lower ED model has a better training performance. Conversely, a negative $\Delta_{\mathrm{f-c}}\mathrm{MSE}_{\mathrm{min}}$ implies a better performance of models with higher ED. Note that this analysis of model training indeed concerns whether the training is successful, i.e., eventually reaches a low value of the loss function, as a proxy of how the exploration of the parameter space was successful in finding the data generating function. While this metric is already sufficient to separate the training behavior of models with high and low ED and bias, as we show in the following, it is important to notice that a more local account of the optimization dynamics, achieved by monitoring the FIM/NTK spectrum or the local ED \cite{Abbas2021b} during training, could provide more information on the properties of the optimization landscape, and is thus a natural extension of the present work.

\subsection{Results with full random structure constants} \label{subsec:numerics_results_full_rnd_models}
\begin{figure*}
    \centering
    \includegraphics[width=\linewidth]{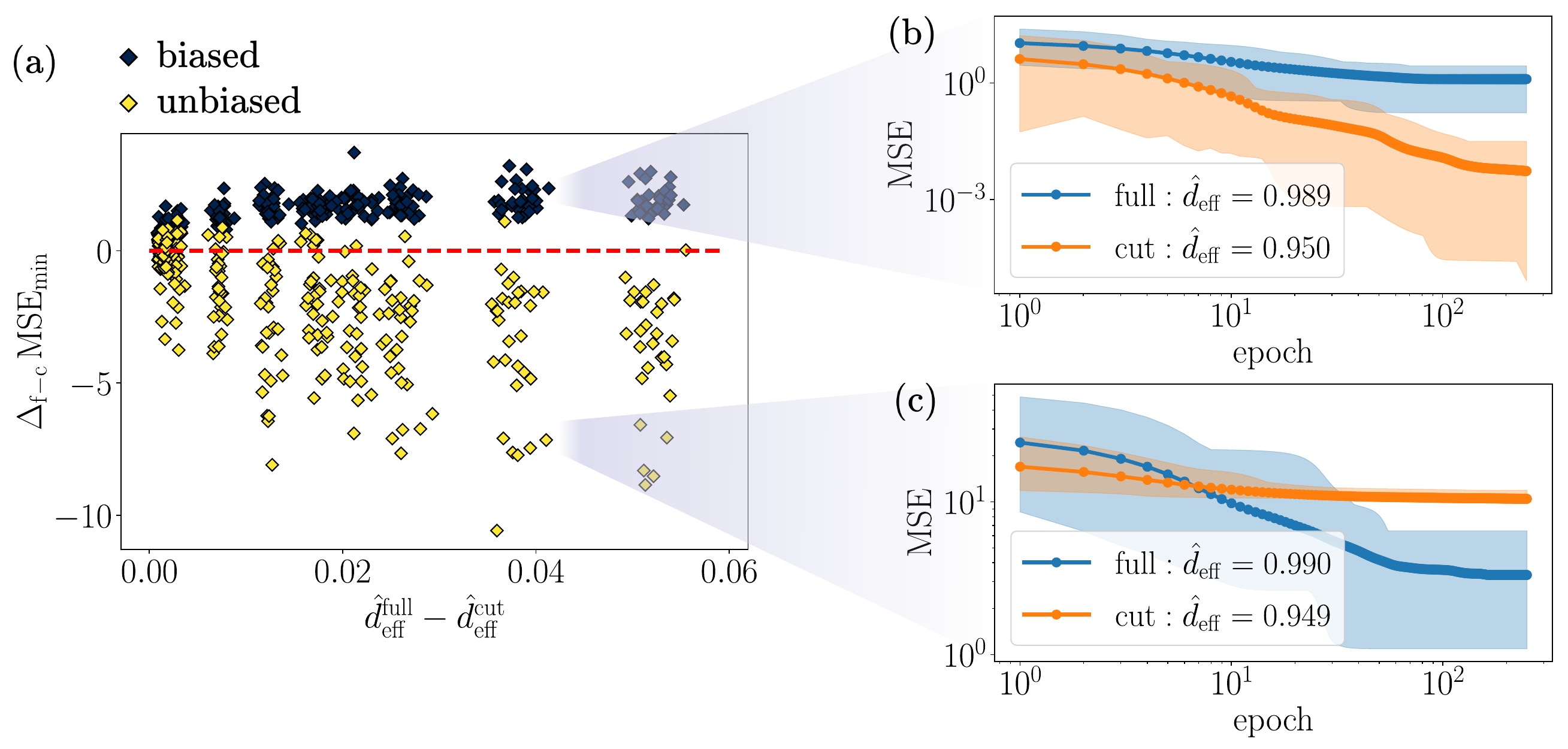}
    \caption{(a) $\Delta_{\mathrm{f-c}}\mathrm{MSE}_{\mathrm{min}}$ for different values of $\hat{d}_{\mathrm{eff}}^{\mathrm{full}}-\hat{d}_{\mathrm{eff}}^{\mathrm{cut}}$, for biased (blue points) and unbiased (yellow points) models. Each point corresponds to $\Delta_{\mathrm{f-c}}\mathrm{MSE}_{\mathrm{min}}$ averaged over $30$ training instances starting from randomly chosen parameters, for a single random model realization, i.e., a random $\Gamma$ uniformly drawn from $[-1,+1]^{D\times K}$. The red line serves as a guide for the eye for zero $\mathrm{MSE}$ difference. The spread over training instances is not shown for clarity, but reported in panels (b) and (c), as representatives of all points in (a). (b) Training curves for a random biased model realization, with full model in blue and cutoff model in orange. (c) Training curves for a random unbiased model realization, with full model in blue and cutoff model in orange. The shading corresponds to the spread over $30$ training instances. 
    For these plots, $N=1$, $\Omega=\{1,...,8\}$ ($d=17$), $\tilde{\Omega}=\{1\}$ ($\tilde{d}=3$), $M=7$, $R=6$, $\mathfrak{n}_{\mathrm{train}}=25$ with a batch size of $5$.}
    \label{fig:MSE_BiasUnbias_wTrain_full_Dloc3} 
\end{figure*}
\begin{figure*}
    \centering
    \includegraphics[width=\linewidth]{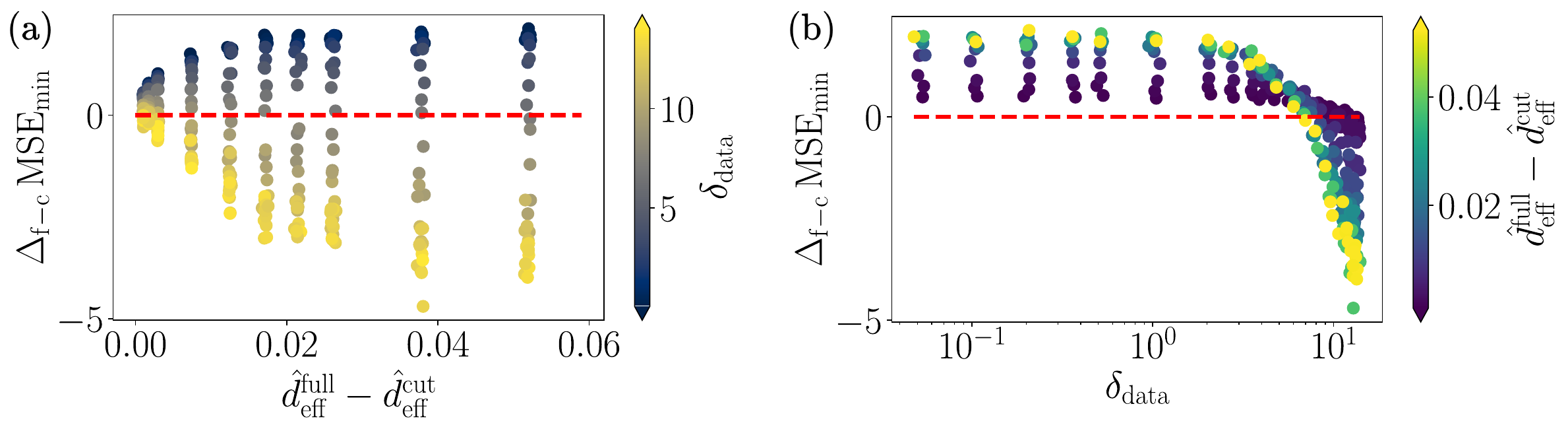}
    \caption{(a) $\Delta_{\mathrm{f-c}}\mathrm{MSE}_{\mathrm{min}}$ for different values of $\hat{d}_{\mathrm{eff}}^{\mathrm{full}}-\hat{d}_{\mathrm{eff}}^{\mathrm{cut}}$, for different values of $\delta_{\mathrm{data}}$ (color scale). Each point corresponds to $\Delta_{\mathrm{f-c}}\mathrm{MSE}_{\mathrm{min}}$ averaged over $30$ training instances for $30$ random model realization. (b) Same as panel (a) but resolved as a function of $\delta_{\mathrm{data}}$. The red line serves as a guide for the eye for zero $\mathrm{MSE}$ difference. 
    For these plots, $N=1$, $\Omega=\{1,...,8\}$ ($d=17$), $\tilde{\Omega}=\{1\}$ ($\tilde{d}=3$), $M=7$, $R=6$, $\mathfrak{n}_{\mathrm{train}}=25$ with a batch size of $5$.}
    \label{fig:MSE_AllBiases_full_Dloc3} 
\end{figure*}
We start by presenting our results obtained by drawing random structure constants $\Gamma$ uniformly drawn in $[-1,+1]^{D\times K}$. As we show in Fig.~\ref{fig:MSE_BiasUnbias_wTrain_full_Dloc3}(a) there is a clear difference in the effect of a higher ED on the training dynamics between biased and unbiased models. Specifically, for biased models (shown in blue) the value of $\Delta_{\mathrm{f-c}}\mathrm{MSE}_{\mathrm{min}}$ is positive and increases with increasing $\hat{d}_{\mathrm{eff}}^{\mathrm{full}}-\hat{d}_{\mathrm{eff}}^{\mathrm{cut}}$. Conversely, for unbiased models (shown in yellow) the value of $\Delta_{\mathrm{f-c}}\mathrm{MSE}_{\mathrm{min}}$ is negative (with high probability) and decreases with increasing $\hat{d}_{\mathrm{eff}}^{\mathrm{full}}-\hat{d}_{\mathrm{eff}}^{\mathrm{cut}}$. That is, in the biased case, models with lower ED train to a lower MSE (as can be seen in Fig.~\ref{fig:MSE_BiasUnbias_wTrain_full_Dloc3}(b) for a specific random model realization), whereas in the unbiased case a higher ED is beneficial for training (as shown in Fig.~\ref{fig:MSE_BiasUnbias_wTrain_full_Dloc3}(c)). These results confirm the following intuitive expectation: a model which is biased towards the data-generating function $y(\boldsymbol{x})$ and which at the same time has a lower ED, can effectively explore a functions' space more constrained around $y(\boldsymbol{x})$, and is therefore easier to train. Conversely, an unbiased model with a higher ED can explore a larger space of function, which makes it probabilistically easier to approximately fit an (in principle) unrelated data-generating function.

To complement these results, we also investigate how a partial bias towards the data-generating function (as defined in Eq.~\eqref{eq:partial_bias_2}) influences the training, for different values of $\hat{d}_{\mathrm{eff}}^{\mathrm{full}}-\hat{d}_{\mathrm{eff}}^{\mathrm{cut}}$. As shown in Fig.~\ref{fig:MSE_AllBiases_full_Dloc3}(a), we again observe that a higher ED is beneficial in the case of unbiased models (where $\Delta_{\mathrm{f-c}}\mathrm{MSE}_{\mathrm{min}}$ is negative), whereas increasing the model's bias (i.e., decreasing $\delta_{\mathrm{data}}$) results in better training performances for models with lower ED (where $\Delta_{\mathrm{f-c}}\mathrm{MSE}_{\mathrm{min}}$ is positive). Furthermore, as shown in Fig.~\ref{fig:MSE_AllBiases_full_Dloc3}(b), there is an extended regime in terms of values of $\delta_{\mathrm{data}}$ where $\Delta_{\mathrm{f-c}}\mathrm{MSE}_{\mathrm{min}}$ is positive, indicating the stability of our results also in the case of a not perfectly biased model (i.e., $\delta_{\mathrm{data}}>0$). This further corroborates the aforementioned intuition that a large ED, as a measure of effectively explorable functions' space, is beneficial for training models with low bias towards the regression task under study, whereas models strongly biased towards the task do benefit from a smaller ED. Further results showcasing the interplay between model bias and ED are provided in the Supplementary Material for different model specifications (i.e., different $N$, $d$, $M$ and $\tilde{d}$), and confirm the findings presented here.

\subsection{Results with tensorized random structure constants} \label{subsec:numerics_train_results_TN_models}
\begin{figure*}
    \centering
    \includegraphics[width=\linewidth]{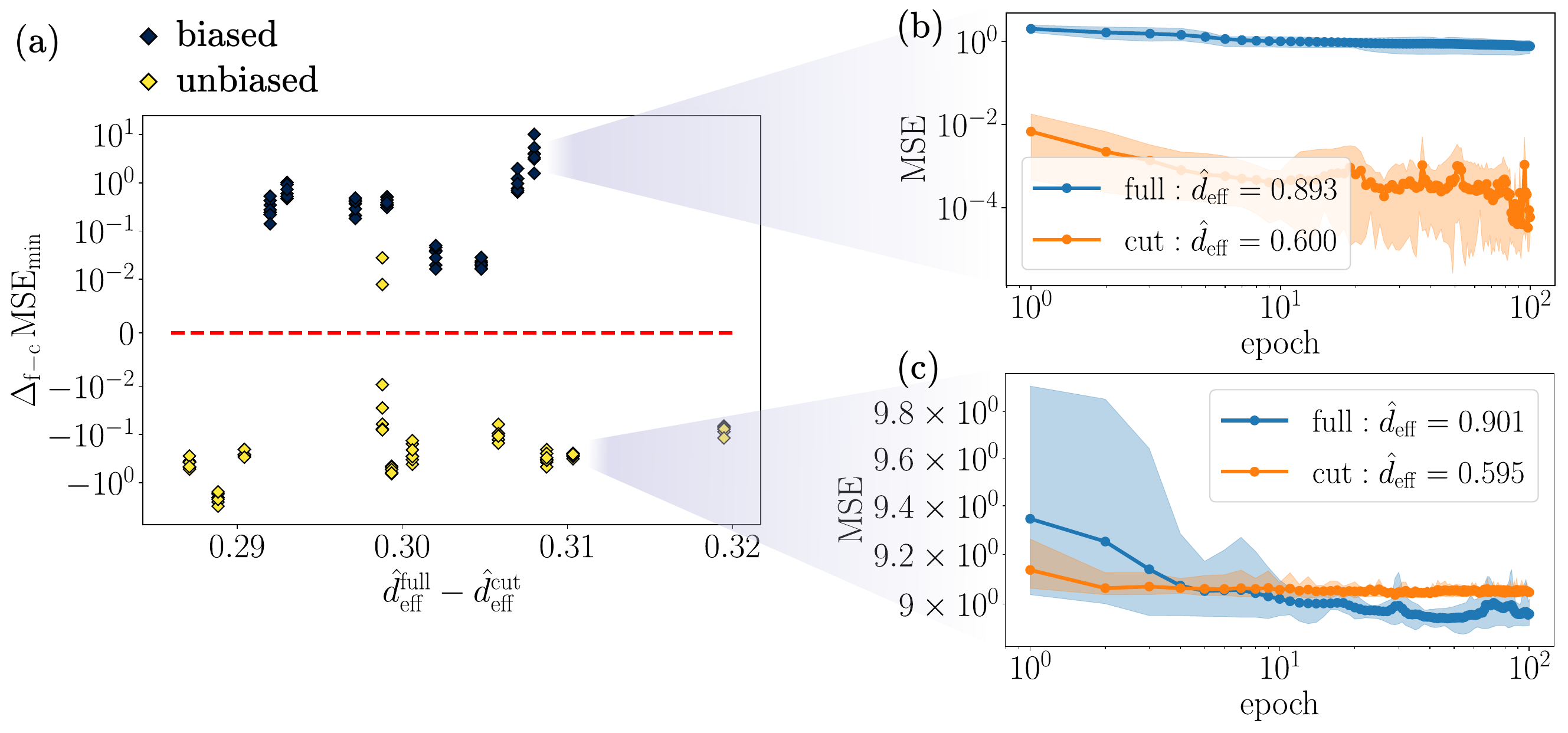}
    \caption{(a) $\Delta_{\mathrm{f-c}}\mathrm{MSE}_{\mathrm{min}}$ for different values of $\hat{d}_{\mathrm{eff}}^{\mathrm{full}}-\hat{d}_{\mathrm{eff}}^{\mathrm{cut}}$, for biased (blue points) and unbiased (yellow points) models. Each point corresponds to $\Delta_{\mathrm{f-c}}\mathrm{MSE}_{\mathrm{min}}$ averaged over $7$ training instances starting from randomly chosen parameters, for a single random model realization, i.e., a random right-normalized tensor train representing $V$, a random MPO representing $U$ and a random TTN representing $T$. The red line serves as a guide for the eye for zero $\mathrm{MSE}$ difference. The spread over training instances is not shown for clarity, but reported in panels (b) and (c), as representatives of all points in (a). (b) Training curves for a random biased model realization, with full model in blue and cutoff model in orange. (c) Training curves for a random unbiased model realization, with full model in blue and cutoff model in orange. The shading corresponds to the spread over $7$ training instances. 
    For these plots, $N=4$, $\Omega=\{1,...,3\}$ ($d=7$), $\tilde{\Omega}=\{1\}$ ($\tilde{d}=3$), $M=24$, $R=2$, $\chi=30$, $\mathfrak{n}_{\mathrm{train}}=6^4$ with a batch size of $12$.}
    \label{fig:MSE_BiasUnbias_wTrain_fullTN_Dloc3} 
\end{figure*}
Here we conduct numerical experiments analogous to those in the previous section, adopting the TN decomposition of the structure constants $\Gamma$ discussed in Section \ref{sec:tensorized_models}, to study how the problem size, i.e., the number of features $N$ and of parameters $M$, affects our results. The training experiments are set up in the same manner as those in in the previous section, with the only difference that the models correspond now to random instances of the TN representing $\Gamma$. As shown in Fig.~\ref{fig:MSE_BiasUnbias_wTrain_fullTN_Dloc3}, the conclusions drawn in Section ~\ref{subsec:numerics_results_full_rnd_models} apply independently of the size of the problem and of the models' details. Specifically, the results shown in Fig.~\ref{fig:MSE_BiasUnbias_wTrain_fullTN_Dloc3}(a) are consistent with those of Fig.~\ref{fig:MSE_BiasUnbias_wTrain_full_Dloc3}(a), showing that in the biased case $\Delta_{\mathrm{f-c}}\mathrm{MSE}_{\mathrm{min}}$ is positive, i.e., models with lower ED train to a lower MSE, whereas in the unbiased case $\Delta_{\mathrm{f-c}}\mathrm{MSE}_{\mathrm{min}}$ is negative, i.e., models with higher ED have better training performance. We refer the reader to \ref{app:tensorized_models} and to the Supplementary Material for further numerical results on training tensorized models, which confirm the findings presented here. The dependence of $\Delta_{\mathrm{f-c}}\mathrm{MSE}_{\mathrm{min}}$ on $M$ and $D$ is analyzed in the next section.

\begin{figure*}
    \centering
    \includegraphics[width=\linewidth]{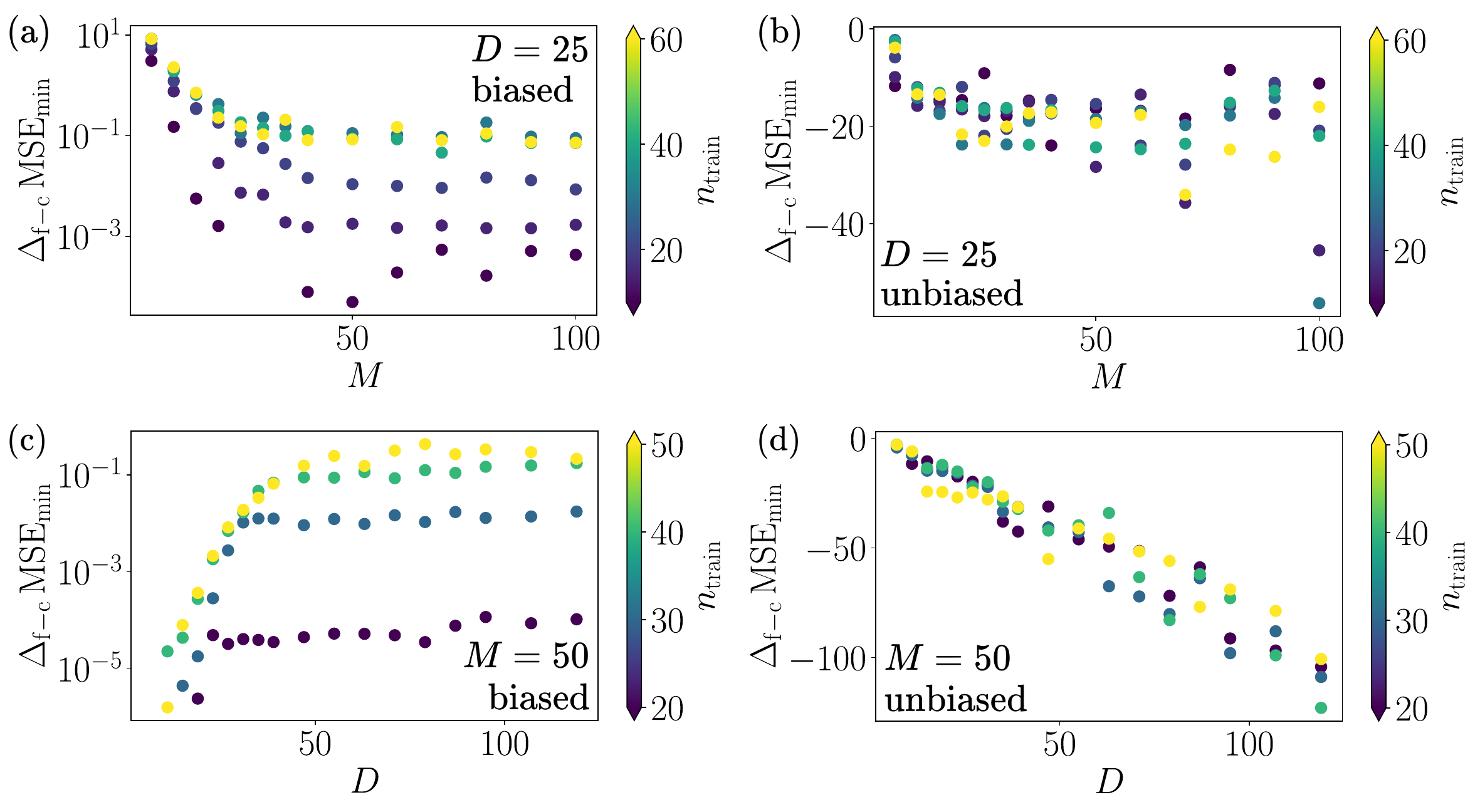}
    \caption{(a) and (b) $\Delta_{\mathrm{f-c}}\mathrm{MSE}_{\mathrm{min}}$ vs.~$M$ for biased and unbiased models, respectively. Here $N=1$, $D=25$, $\tilde{\Omega}=\{1\}$ ($\tilde{d}=3$), $R=3$ and $\chi=50$. (c) and (d) $\Delta_{\mathrm{f-c}}\mathrm{MSE}_{\mathrm{min}}$ vs.~$D$ for biased and unbiased models, respectively. Here $N=1$, $M=50$, $\tilde{\Omega}=\{1\}$ ($\tilde{d}=3$), $R=2$ and $\chi=120$. In all panels, each point corresponds to $\Delta_{\mathrm{f-c}}\mathrm{MSE}_{\mathrm{min}}$ averaged over $30$ training instances and $30$ random model realizations, i.e., random right-normalized tensor trains representing $V$. The color scale refers to the number of training data used $\mathfrak{n}_{\mathrm{train}}$, with a batch size of $5$.}
    \label{fig:Figure_scaling_training} 
\end{figure*}

\subsection{Effect of model under- and overparameterization}
\label{subsec:scaling_with_M}
In this section we investigate how the ED-bias interplay in training changes with the input function space dimension $D$ and the number of parameters $M$. This allows us to assess the behavior across the transition from the underparameterized ($M<D$) to the overparameterized ($M\geq D$) regime \cite{Larocca2023}. The results are shown in Fig.~\ref{fig:Figure_scaling_training}, where we show how $\Delta_{\mathrm{f-c}}\mathrm{MSE}_{\mathrm{min}}$ behaves as a function of $M$ and $D$ for different training set sizes $\mathfrak{n}_{\mathrm{train}}$. Overall, we observe that the sign of $\Delta_{\mathrm{f-c}}\mathrm{MSE}_{\mathrm{min}}$ remains robust across the tested hyperparameters, whereas the magnitude indeed depends on the values of $M$, $D$ and $\mathfrak{n}_{\mathrm{train}}$. The dependence of $\Delta_{\mathrm{f-c}}\mathrm{MSE}_{\mathrm{min}}$ on $M$ and $D$ is mostly visible in the biased case (panels (a) and (c)). In particular, $\Delta_{\mathrm{f-c}}\mathrm{MSE}_{\mathrm{min}}$ rapidly decreases as a function of $M$ reaching an approximately constant (but still positive) value for $M\geq D$ (Fig.~\ref{fig:Figure_scaling_training}(a)). The regime $M\geq D$ corresponds to the overparameterized regime where the number of parameters is larger than the dimension of the space of functions accessible by the model. Hence, several redundant parameters exist, which make the training of a model easier, independently of whether this is has a high or a low ED. As a consequence, in this regime we observe a reduction of $\Delta_{\mathrm{f-c}}\mathrm{MSE}_{\mathrm{min}}$ since the model with high ED can effectively rely on more redundant parameters for finding the global minimum of the loss. Consistently, $\Delta_{\mathrm{f-c}}\mathrm{MSE}_{\mathrm{min}}$ rapidly increases as a function of $D$ reaching an approximately constant value for $D\geq M$ (Fig.~\ref{fig:Figure_scaling_training}(c)). The regime $D\geq M$ corresponds to an underparameterized regime where, in the biased case, the absence of redundant parameters results in a better training performance of models with low ED. This difference between under- and overparameterized regimes in the biased case can equivalently be explained from the perspective of the NTK, and the fact that in the underparameterized regime the NTK has $D-M$ ``frozen" (i.e., with zero eigenvalue) modes. We refer the reader to the Supplementary Material for a qualitative explanation of this phenomenon. There, we also discuss the dependence on $\mathfrak{n}_{\mathrm{train}}$, explaining how the number of training samples affects the NTK and the training dynamics, and why in the biased case a lower $\mathfrak{n}_{\mathrm{train}}$ reduces the gap $\Delta_{\mathrm{f-c}}\mathrm{MSE}_{\mathrm{min}}$ between high and low ED models.

Turning to the unbiased case, Fig.~\ref{fig:Figure_scaling_training}(c) shows a negative $\Delta_{\mathrm{f-c}}\mathrm{MSE}_{\mathrm{min}}$ which slightly decays for increasing $M\leq D$ at fixed $D$. Again, we attribute this to the high ED models being able to explore progressively more directions in the $D$ dimensional input space as $M$ increases, more effectively than low ED ones. Increasing  $M$ above $D$ does not increase the number of different explorable directions, hence does not result in a further separation between high and low ED models. Fig.~\ref{fig:Figure_scaling_training}(d) shows that $\Delta_{\mathrm{f-c}}\mathrm{MSE}_{\mathrm{min}}$ is negative and decays with increasing $D$ for fixed $M$ in both regimes. In this case, models with increased $D$ have a larger base space where a better approximation for the target function can be found, and a higher ED results in an effectively larger portion of this space being available (hence lowering $\Delta_{\mathrm{f-c}}\mathrm{MSE}_{\mathrm{min}}$). In the unbiased case, the effect of changing $\mathfrak{n}_{\mathrm{train}}$ is not as significant, and we provide a tentative explanation for this in the Supplementary Material.

\section{Conclusion and outlook} \label{sec:conclusions}
In this work we investigated how the effective dimension (ED), calculated from the Fisher information matrix (FIM), influences the training of regression models using gradient descent methods. Specifically, we studied the interplay between the ED and the bias that a model has towards the regression task at hand, and were able to draw the following main conclusions.
A high ED, corresponding to a high model capacity of exploring independent directions in model space, is beneficial for training in the low bias regime, i.e., when the model is largely agnostic to the data-generating function to be learned.
Conversely, in the biased regime, i.e., when the model's structure is well suited to the problem's data-generating function, a low ED does result in better training performance.
These results confirm, in a quantitative manner, the intuitive expectation that if the space a model has access to is constrained (i.e., a model with low ED) around the problem's data-generating function (i.e., a biased model), training the model effectively becomes easier.
These findings are corroborated by the analysis of the NTK and its effects on the training dynamics, which we show in the Supplementary Material: by relating model bias with the alignment between NTK eigenfunctions and data, and using the equivalence between FIM and NTK spectra, we can construct a simplified setting qualitatively capturing the phenomena observed here.
Thus, a high ED does not always result in a model's faster training, which we interpret as a further sign of the difficulty of defining a task-independent evaluation metric that can assess a model's performance prior to its training.

We foresee several possible directions for extending the presented results. First, it is important to comment on the fact that our analysis is based on comparing ED, bias and training performance of models with the same underlying structure (i.e., a finite number of chosen basis functions for the inputs' and parameters' functions space). Our choice is motivated by the need of comparing models where bias and ED can be controlled, while eliminating other potential sources of fluctuations coming from different architectural choices. Nevertheless, due to the equivalence between Fourier models and QNNs, our results can already provide guidance on practical QNN design choices that can be used to bias the model outputs towards specific target functions, via data re-uploading and the use of entangling operations or correlated measurements (see Supplementary Material for details). The design of QNNs with tunable ED requires a comprehensive characterization of how the QNN architecture influences the structure constants in QNNs, which at the moment is still lacking.

While the objective of our work is to establish the mechanism for the ED-bias interplay in training, it is important to test this in progressively more real-world-related scenarios, under larger input spaces and different input data distributions. While the ED-bias interplay is a general mechanism, investigating different and more realistic settings might uncover important details that can help the design of models for practical tasks. Analogous experiments can be performed using actual (simulated) QNN models or even classical NNs. While the use of QNNs would still amount to Fourier models with different (yet unknown) distributions in the structure constants, we find the investigation of the present mechanism for classical NNs, and for which conditions and data it may emerge, of high interest. We would find it also interesting to investigate the same mechanism when comparing inherently different models, i.e., models accessing different types of function spaces. For example, designing such experiments comparing quantum and classical NNs could yield more insights on the function classes, and thereby the type of data, most suitable to these two ansatzes.

Going beyond the analysis of the optimization dynamics, an important future direction would be studying how the generalization error changes for models with different ED and bias. While the ED can be used to bound generalization errors \cite{Abbas2021,Abbas2021b}, our results qualitatively suggest the model-task bias as an important factor that may strongly affect the generalization ability (as is the case for kernel methods \cite{Canatar2021}). Obviously, all the analysis and discussion presented here admit natural extensions to classification (see the Supplementary Material for suggestions on a potential generalization) and generative models. Furthermore, it would be interesting to investigate the questions addressed here also in the context of natural gradient descent \cite{Amari1997,Amari2019}, which could help to develop a deeper understanding of the ED-bias interplay in training ML models.

There are also several opportunities for establishing a deeper connection between our findings and the theory of quantum machine learning. One interesting direction could be investigating the connections with (quantum) kernel methods \cite{Schuld2021b}, for which results on the effects of inductive bias (kernel-task alignment) on the training and generalization performance have already been established \cite{Canatar2021,Kuebler2021}. Other important aspects to be addressed in the future concern the connection to existing works on generalization \cite{Caro2021,Banchi2021,Peters2023}, overfitting \cite{Peters2023,Dar2021}, overparameterization \cite{Larocca2023}, and barren plateaus \cite{McClean2018,Thanasilp2023,Larocca2025} in quantum machine learning. Furthermore, since in our numerical analysis we primarily focused on Fourier models, we note that there exist several works investigating the use of Fourier features for \emph{dequantizing}, or building \emph{classical surrogates}, of quantum machine learning models \cite{Landman2022,Schreiber2023,Sweke2023}. In the context of our work, it would be interesting to understand what features of a QNN make it dequantizable via the tensorized model introduced here, generalizing recent works \cite{Shin2024} and enabling further understanding on the function classes encompassed by quantum machine learning models. 

Finally, to connect the present results with quantum hardware implementations, we remark that the Fourier series representation of QNNs remains largely valid also when realistic hardware noise and certain types of gate errors are considered. A more fundamental issue remains that of finite sampling errors, which can hinder the training of a model when gradients are poorly resolved. While we highlight strategies to mitigate these effects \cite{Kreplin2024,Pastori2025}, we also do not exclude the possibility that moderate finite sampling noise can act as further stochastic term helping the model in exploring the accessible function space. Investigating how this influences the ED-bias interplay discussed here is left as interesting future work.

\section*{Acknowledgments}
This project was made possible by the DLR Quantum Computing Initiative and the Federal Ministry for Economic Affairs and Climate Action; qci.dlr.de/projects/klim-qml. V.E.~was funded by the European Research Council (ERC) Synergy Grant “Understanding and Modelling the Earth System with Machine Learning (USMILE)” under the Horizon 2020 research and innovation programme (Grant agreement No.~855187).
V.E.~was additionally supported by the Deutsche Forschungsgemeinschaft (DFG, German Research Foundation) through the Gottfried Wilhelm Leibniz Prize awarded to Veronika Eyring (Reference No.~EY 22/2-1).
This work used resources of the Deutsches Klimarechenzentrum (DKRZ) granted by its Scientific Steering Committee (WLA) under project ID bd1179.

\section*{Data Availability}
The code used for producing the results of this work are available at the following link: \href{https://github.com/EyringMLClimateGroup/pastori26MLST_fim-training-fourier-models}{https://github.com/EyringMLClimateGroup/pastori26MLST\_fim-training-fourier-models}.

\appendix

\section{Fourier series representation of QNN} \label{app:FourierSeriesQNNs}
Here we sketch the derivation of the Fourier series representation for a function $f_{\boldsymbol{\theta}}(\boldsymbol{x})$ obtained as output of a QNN as in Eq.~\eqref{eq:QNN_output}, focusing for simplicity on the case of a one-dimensional input. This can be easily generalized to more input features \cite{Schuld2021,Casas2023} following the same derivation. We consider $L$ layers of data re-uploading  \cite{PerezSalinas2020,Schuld2021} with unitary $\hat{U}_{\boldsymbol{\theta}}(x)$ expressed as
\begin{equation}
    \hat{U}_{\boldsymbol{\theta}}(x)=\prod_{\ell=1}^L\Big[\hat{S}(x)\,\hat{W}^{(\ell)}(\boldsymbol{\theta}^{(\ell)})\Big] \,\,,
\end{equation}
where the input $x$ and trainable parameters $\boldsymbol{\theta}^{(\ell)}$ are encoded in $\hat{S}(x)$ and $\hat{W}^{(\ell)}(\boldsymbol{\theta}^{(\ell)})$ as angles in rotation gates. The vector $\boldsymbol{\theta}$ summarizes the dependence on all $\boldsymbol{\theta}^{(\ell)}$, i.e., $\boldsymbol{\theta}=\{\boldsymbol{\theta}^{(\ell)}\}_{\ell=1,...,L}$. We can write $\hat{W}^{(\ell)}(\boldsymbol{\theta}^{(\ell)})=\prod_{j_{\ell}=1}^{J_{\ell}}\hat{G}^{(\ell)}_{j_{\ell}}(\theta^{(\ell)}_{j_{\ell}})$, and following \cite{Schuld2021,Casas2023} we switch to the diagonal representations of $\hat{S}(x)$ and $\hat{G}^{(\ell)}_{j_{\ell}}(\theta^{(\ell)}_{j_{\ell}})$. These have eigenvalues $\{\mathrm{e}^{-\mathrm{i}\lambda_{\alpha}x}\}_{\alpha=1,...,\mathcal{D}}$ and $\{\mathrm{e}^{-\mathrm{i}\eta^{(\ell,j_{\ell})}_{\beta}\theta^{(\ell)}_{j_{\ell}}}\}_{\beta=1,...,\mathcal{D}}$, respectively, which make the trigonometric dependence of $f_{\boldsymbol{\theta}}(\boldsymbol{x})$ on the inputs and parameters evident. Using these, we can calculate the components of the state $\hat{U}_{\boldsymbol{\theta}}(x)|0\rangle$ expanded in terms of the functions $\mathrm{e}^{-\mathrm{i}\Lambda_{\boldsymbol{\alpha}}x}$ and $\mathrm{e}^{-\mathrm{i}\boldsymbol{\eta}_{\boldsymbol{\beta}}\cdot\boldsymbol{\theta}}$, with the shorthand notation $\boldsymbol{\alpha}=(\alpha_1,...,\alpha_L)$, $\boldsymbol{\beta}=(\beta_1,...,\beta_M)$, $\Lambda_{\boldsymbol{\alpha}}=\sum_{\ell}\lambda_{\alpha_{\ell}}$ and $\boldsymbol{\eta}_{\boldsymbol{\beta}}=(\eta^{(1)}_{\beta_1},...,\eta^{(M)}_{\beta_M})$. We refer to the Supplementary Material for the details of the calculation, and we report here the result
\begin{equation}
    f_{\boldsymbol{\theta}}(x)=\sum_{\boldsymbol{\alpha},\boldsymbol{\alpha}'}\sum_{\boldsymbol{\beta},\boldsymbol{\beta}'}\tilde{\Gamma}_{(\boldsymbol{\alpha};\boldsymbol{\alpha}'),(\boldsymbol{\beta};\boldsymbol{\beta}')}\,\mathrm{e}^{\mathrm{i}(\Lambda_{\boldsymbol{\alpha}}-\Lambda_{\boldsymbol{\alpha}'})x}\,\mathrm{e}^{\mathrm{i}(\boldsymbol{\eta}_{\boldsymbol{\beta}}-\boldsymbol{\eta}_{\boldsymbol{\beta}'})\cdot\boldsymbol{\theta}} \,\,,
    \label{eqapp:QNN_output_fourier}
\end{equation}
which has the same form as Eq.~\eqref{eq:QNN_fourier_expansion_exp} in the main text. As an explicit example of the sets of frequencies accessible to the model, we consider the case where input features and variational parameters are encoded as angles of single qubit rotations of the form $\mathrm{e}^{-\mathrm{i}\frac{\phi}{2}\boldsymbol{n}\cdot\hat{\boldsymbol{\sigma}}}$ (with $\phi$ being the feature/parameter to be encoded, $\boldsymbol{n}$ an arbitrary rotation axis and $\hat{\boldsymbol{\sigma}}$ the vector of Pauli matrices). In this case, the eigenvalues of the generators of $\hat{S}(x)$ and $\hat{G}^{(\ell)}_{j_{\ell}}(\theta^{(\ell)}_{j_{\ell}})$ are $\lambda_{\alpha_{\ell}}\in\Big\{-\frac{1}{2},+\frac{1}{2}\Big\}$ and $\eta^{(m)}_{\beta_{m}}\in\Big\{-\frac{1}{2},+\frac{1}{2}\Big\}$, respectively. For the dependence on the inputs $x$ we therefore have $\lambda_{\alpha_{\ell}}-\lambda_{\alpha_{\ell}'}\in\Big\{-1,0,+1\Big\}$, hence $\Lambda_{\boldsymbol{\alpha}}-\Lambda_{\boldsymbol{\alpha}'}\in\Big\{-L,-L+1,...,L-1,L\Big\}$, thus yielding the local basis set $\mathcal{B}=\{1,\,\sqrt{2}\cos(x),...,\,\sqrt{2}\cos(Lx),\,\sqrt{2}\sin(x),...,\,\sqrt{2}\sin(Lx)\}$. Similarly, for the dependence on the parameters $\theta_m$ we have $\eta^{(m)}_{\beta_{m}}-\eta^{(m)}_{\beta_{m}'}\in\Big\{-1,0,+1\Big\}$, which yields the local basis set for the parameters $\tilde{\mathcal{B}}_m=\{1,\,\sqrt{2}\cos\theta_m,\,\sqrt{2}\sin\theta_m\}$.

The Fourier representation is not restricted to the noiseless, pure-state setting. In the presence of hardware noise, the circuit evolution can be described at the level of density matrices by a composition of quantum channels
\begin{equation}
    \hat{\rho}_{\boldsymbol{\theta}}(x)=\bigcirc_{\ell=1}^L\Big[\mathcal{E}_{\ell}\circ\mathcal{S}(x)\circ\mathcal{W}^{(\ell)}(\boldsymbol{\theta}^{(\ell)})\Big](\hat{\rho}_0) \,\,,
\end{equation}
with $\mathcal{E}_{\ell}$ being the noise channels, and $\mathcal{S}(x)$ and $\mathcal{W}^{(\ell)}(\boldsymbol{\theta}^{(\ell)})$ the unitary channels corresponding to $\hat{S}(x)$ and $\hat{W}^{(\ell)}(\boldsymbol{\theta}^{(\ell)})$, respectively, acting as, e.g., $\mathcal{S}(x)(\cdot)=\hat{S}(x)\cdot\hat{S}^{\dagger}(x)$. The QNN output is given by the measurement output $\hat{\rho}_{\boldsymbol{\theta}}(x)=\mathrm{tr}\big(\hat{M}\hat{\rho}_{\boldsymbol{\theta}}(x)\big)$. Since both quantum channels and the final measurement map are linear, and angle-encoding gates still introduce only trigonometric dependence on $x$ and $\boldsymbol{\theta}$, the output retains the same Fourier-model form. This means that hardware noise does not invalidate the Fourier-series description, but instead modifies the structure constants $\Gamma_{\mu,\nu}$. Similarly, also random gate rotation errors can effectively be described by dephasing channels with strength depending only on the variance of the fluctuations \cite{Crow2014}, and therefore they also do not affect the Fourier series representation but only the structure constants. From the trainability perspective, noise can qualitatively alter the optimization landscape and may induce noise-induced barren plateaus \cite{Wang2021}, leading to exponentially vanishing gradients in the number of qubits and circuit depth.

\section{Properties of FIM for regression models} \label{app:FIM_derivations}
We provide here a sketch of the derivation of the properties of the FIM discussed in Section \ref{subsec:FIM_diagr_expr}. The details of the calculations are given in the Supplementary Material. Our starting point is Eq.~\eqref{eq:FIM_analytic_expr}, which can be rewritten as $ F_{j,k}(\boldsymbol{\theta})=\big(\boldsymbol{\iota}(\boldsymbol{\theta})\big|\mathcal{F}^{(j,k)}\big|\boldsymbol{\iota}(\boldsymbol{\theta})\big)$, with $\big|\boldsymbol{\iota}(\boldsymbol{\theta})\big)$ the $K$-dimensional vector with components $\iota_{\nu}(\boldsymbol{\theta})$ and 
\begin{equation}
    \mathcal{F}^{(j,k)}=B_j^{\top}\,V\,S^2\,V^{\top}B_k = \sum_{\rho=1}^D s_{\rho}^2\;B_j^{\top}\,P_{\rho}\,B_k \,\,,
\end{equation}
where $B_j=I^{(1)}_{\tilde{d}}\otimes I^{(2)}_{\tilde{d}}\otimes\,... \otimes\,\beta^{(j)}\,\otimes\,...\otimes I^{(M)}_{\tilde{d}}$ (with $I^{(k)}_{\tilde{d}}$ being the $\tilde{d}$-dimensional identity matrix acting on the function space `local' to the $k$-th parameter), and $P_{\rho}$ being the projection on the subspace spanned by the vector $V_{\cdot,\rho}$. For showing Eq.~\eqref{eq:FIM_prop_1}, it is sufficient to note that thanks to the presence of the projection $P_{\rho}$, the FIM $F(\boldsymbol{\theta})$ can be expressed as a sum of at most $D$ linearly independent $M\times M$ projections, which implies that its rank can be at most $D$ (if $M>D$). For showing Eq.~\eqref{eq:FIM_prop_2}, we use results from random matrix theory \cite{Brouwer1996,Collins2006,Collins2009} to derive the expectation value and the variance of the FIM elements over random realizations of $V\in\mathrm{O}(K)$. In particular, we show that
\begin{equation}
    \mathbb{E}_{V\in \mathrm{O}(K)}\big[F_{j,k}(\boldsymbol{\theta})\big]=\frac{\mathrm{tr}(S^2)}{K}\,\big(\boldsymbol{\iota}(\boldsymbol{\theta})\big|B_j^{\top}B_k\big|\boldsymbol{\iota}(\boldsymbol{\theta})\big) \,\,,
\end{equation}
and
\begin{equation}
    \begin{split}
        \mathrm{Var}_{V\in \mathrm{O}(K)}\big[F_{j,k}(\boldsymbol{\theta})\big]&=\frac{\mathrm{tr}(S^4)}{K^2}\,\Big[\big(\boldsymbol{\iota}(\boldsymbol{\theta})\big|B_j^{\top}B_k\big|\boldsymbol{\iota}(\boldsymbol{\theta})\big)^2+\\
        &\quad\quad\quad\quad\quad\big(\boldsymbol{\iota}(\boldsymbol{\theta})\big|B_j^{\top}B_j\big|\boldsymbol{\iota}(\boldsymbol{\theta})\big)\big(\boldsymbol{\iota}(\boldsymbol{\theta})\big|B_k^{\top}B_k\big|\boldsymbol{\iota}(\boldsymbol{\theta})\big)\Big]+\\
        &\quad\quad\quad\quad\quad\mathcal{O}(K^{-3}) \,\,.
    \end{split}
\end{equation}
Then, using the orthonormality of the basis functions $\iota^{(m)}_{\nu_m}(\theta_m)$, one has that $(\boldsymbol{\theta})\big|B_j^{\top}B_k\big|\boldsymbol{\iota}(\boldsymbol{\theta})\big)\in\mathcal{O}(K)$ if $j=k$, being suppressed otherwise, which results in Eq.~\eqref{eq:FIM_prop_2}.

\section{Construction of biased and unbiased models} \label{app:biased_unbiased_models}
In this appendix we provide more details on the construction of biased and unbiased models. Further details can be found in the Supplementary Material. We start from the expression of the data-generating function $y(\boldsymbol{x})$ provided in Eq.~\eqref{eq:data_gen_function}. In order to construct a $K\times D$ matrix $V^{\mathrm{(d)}}$ with orthonormal columns and satisfying the property $\sum_{\nu}V^{\mathrm{(d)}}_{\nu,\rho}\,\iota_{\nu}(\boldsymbol{\theta}^*)=0$ for $\rho=R+1,...,K$, it is sufficient to construct is as $V^{\mathrm{(d)}}=\big[V\quad W\big]$, i.e., by horizontally stacking a $K\times R$ matrix $V$ and a $K\times(D-R)$ matrix $W$ both with orthonormal columns, satisfying $W^{\top}V=0$ and with $\sum_{\nu}W_{\nu,\sigma}\,\iota_{\nu}(\boldsymbol{\theta}^*)=0$. Once $V$ has been constructed (e.g., randomly drawn), the two conditions on $W$ can be implemented using Gram-Schmidt orthogonalization from a set of $(D-R)$ randomly chosen vectors.
For constructing partially biased models, we replace $V^{\mathrm{(d)}}$ in Eq.~\eqref{eq:data_gen_function} with $V^{\mathrm{(d)}}_{\epsilon}$ obtained by adding a random perturbation of strength $\epsilon$ to its elements. This is done by setting $V^{\mathrm{(d)}}_{\epsilon}=\mathrm{ortho}(V^{\mathrm{(d)}}+\epsilon\,G)$, where $G$ is a $K\times D$ matrix with standard Gaussian entries and $\mathrm{ortho}(\cdot)$ refers to the process of Gram-Schmidt orthogonalization of the columns of the argument.

\section{Tensorized models and additional results} \label{app:tensorized_models}
\begin{figure*}
    \centering
    \includegraphics[width=\linewidth]{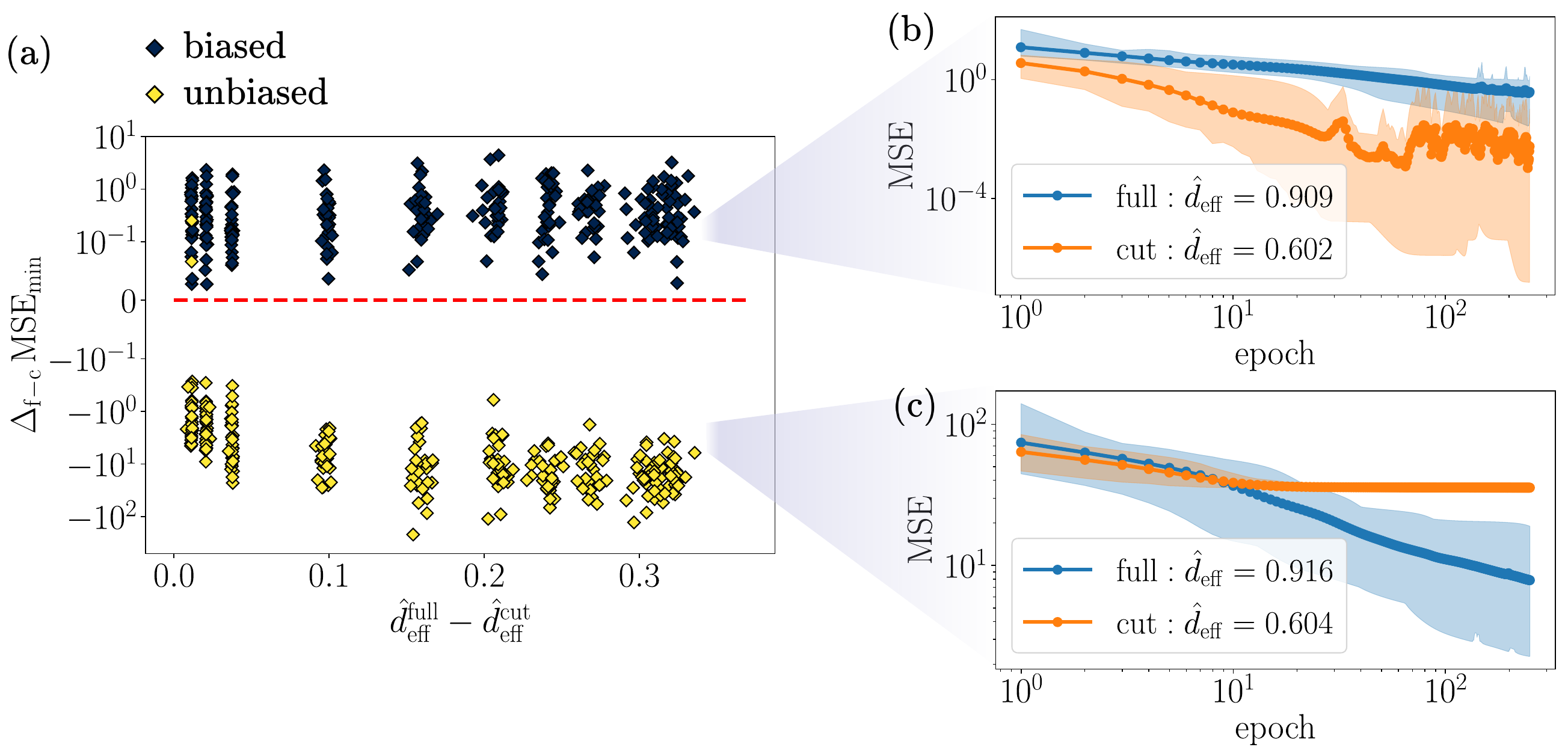}
    \caption{(a) $\Delta_{\mathrm{f-c}}\mathrm{MSE}_{\mathrm{min}}$ for different values of $\hat{d}_{\mathrm{eff}}^{\mathrm{full}}-\hat{d}_{\mathrm{eff}}^{\mathrm{cut}}$, for biased (blue points) and unbiased (yellow points) models. Each point corresponds to $\Delta_{\mathrm{f-c}}\mathrm{MSE}_{\mathrm{min}}$ averaged over $30$ training instances starting from randomly chosen parameters, for a single random model realization, i.e., a random right-normalized tensor train representing $V$. The red line serves as a guide for the eye for zero $\mathrm{MSE}$ difference. The spread over training instances is not shown for clarity, but reported in panels (b) and (c), as representatives of all points in (a). (b) Training curves for a random biased model realization, with full model in blue and cutoff model in orange. (c) Training curves for a random unbiased model realization, with full model in blue and cutoff model in orange. The shading corresponds to the spread over $30$ training instances. 
    For these plots, $N=1$, $\Omega=\{1,...,17\}$ ($d=35$), $\tilde{\Omega}=\{1\}$ ($\tilde{d}=3$), $M=32$, $R=7$, $\chi=60$, $\mathfrak{n}_{\mathrm{train}}=30$ with a batch size of $5$.}
    \label{fig:MSE_BiasUnbias_wTrain_TN_Dloc3} 
\end{figure*}
\begin{figure*}
    \centering
    \includegraphics[width=\linewidth]{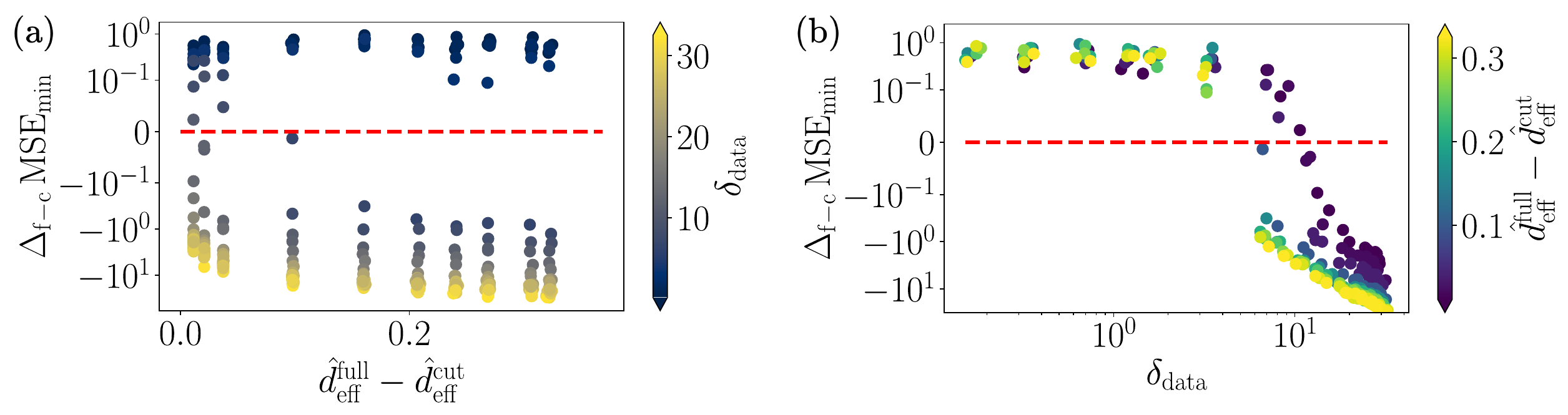}
    \caption{(a) $\Delta_{\mathrm{f-c}}\mathrm{MSE}_{\mathrm{min}}$ for different values of $\hat{d}_{\mathrm{eff}}^{\mathrm{full}}-\hat{d}_{\mathrm{eff}}^{\mathrm{cut}}$, for different values of $\delta_{\mathrm{data}}$ (color scale). Each point corresponds to $\Delta_{\mathrm{f-c}}\mathrm{MSE}_{\mathrm{min}}$ averaged over $30$ training instances for $30$ random model realization. (b) Same as panel (a) but resolved as a function of $\delta_{\mathrm{data}}$. The red line serves as a guide for the eye for zero $\mathrm{MSE}$ difference. 
    For these plots, $N=1$, $\Omega=\{1,...,17\}$ ($d=35$), $\tilde{\Omega}=\{1\}$ ($\tilde{d}=3$), $M=32$, $R=7$, $\chi=60$, $\mathfrak{n}_{\mathrm{train}}=30$ with a batch size of $5$.}
    \label{fig:MSE_AllBiases_TN_Dloc3} 
\end{figure*}
Here we provide details on the construction of tensorized models together with additional numerical results on the effects of bias and ED on their training dynamics. More details on their construction and further numerical results can be found in the Supplementary Material. The tensor-train representation of $V$ has the following form
\begin{equation}
    V_{\nu,\sigma}\approx\sum_{a_1,...,a_{M-1}=1}^{\chi}\mathcal{V}^{[1]\,\nu_1}_{\sigma,a_1}\,\mathcal{V}^{[2]\,\nu_2}_{a_1,a_2}\,...\mathcal{V}^{[M]\,\nu_M}_{a_{M-1},1} \,\,,
\end{equation}
where $\mathcal{V}^{[m]}$ are rank-3 tensors satisfying the right-normalization condition, in order for $V$ to have orthonormal columns. This decomposition of $V$ admits the following graphical representation
\begin{align}
    V_{\nu,\sigma}=\;
    \begin{tikzpicture} [baseline={([yshift=+0ex]current bounding box.center)}]
        \node[] (rho) at (0,-1) {$\sigma$};
        \node[rectangle,minimum width=10em,draw] (V) at (2.5,-1) {$V$};
        \node[] (nu1) at (1,0) {$\nu_1$};
        \node[] (nu2) at (2,0) {$\nu_2$};
        \node[] (dots) at (3,0) {$\cdots$};
        \node[] (nuM) at (4,0) {$\nu_M$};
        \draw [thick] (rho) -- (V);
        \draw [thick] (nu1) -- (nu1 |- V.north);
        \draw [thick] (nu2) -- (nu2 |- V.north);
        \draw [thick] (nuM) -- (nuM |- V.north);
    \end{tikzpicture}
    \;\approx\;
    \begin{tikzpicture} [baseline={([yshift=+0ex]current bounding box.center)}]
        \node[] (rho) at (0,-1) {$\sigma$};
        \node[rectangle, draw] (V1) at (1,-1) {$\mathcal{V}^{[1]}$};
        \node[rectangle, draw] (V2) at (2.1,-1) {$\mathcal{V}^{[2]}$};
        \node[] (Vdots) at (3.2,-1) {$\cdots$};
        \node[rectangle, draw] (VM) at (4.3,-1) {$\mathcal{V}^{[M]}$};
        \node[] (nu1) at (1,0) {$\nu_1$};
        \node[] (nu2) at (2.1,0) {$\nu_2$};
        \node[] (dots) at (3.2,0) {$\cdots$};
        \node[] (nuM) at (4.3,0) {$\nu_M$};
        \draw [thick] (rho) -- (V1) -- (V2) -- (Vdots) -- (VM);
        \draw [thick] (nu1) -- (nu1 |- V1.north);
        \draw [thick] (nu2) -- (nu2 |- V2.north);
        \draw [thick] (nuM) -- (nuM |- VM.north);
    \end{tikzpicture}
\end{align}
The orthogonal matrix $U$ is decomposed as an orthogonal matrix product operator (MPO) 
\begin{equation}
    U_{\mu,\rho}\approx\sum_{a_1,...,a_{N-1}=1}^{\chi}\mathcal{U}^{[1]\,\mu_1,\rho_1}_{1,a_1}\,\mathcal{U}^{[2]\,\mu_2,\rho_2}_{a_1,a_2}\,...\,\mathcal{U}^{[M]\,\mu_N,\rho_N}_{a_{N-1},1} \,\,,
\end{equation}
with $\mathcal{U}^{[n]}$ being rank-4 tensors constrained to yield an orthogonal $U$. The TN decomposition of $U$ has the following diagrammatic representation
\begin{align}
    U_{\mu,\rho}=\;
    \begin{tikzpicture} [baseline={([yshift=+0ex]current bounding box.center)}]
        \node[] (mu1) at (0,2) {$\mu_1$};
        \node[] (mu2) at (0,1) {$\mu_2$};
        \node[] (mudots) at (0,0) {$\vdots$};
        \node[] (muN) at (0,-1) {$\mu_N$};
        \node[rectangle,minimum height=10em,draw] (U) at (1,0.5) {$U$};
        \node[] (rho1) at (2,2) {$\rho_1$};
        \node[] (rho2) at (2,1) {$\rho_2$};
        \node[] (rhodots) at (2,0) {$\vdots$};
        \node[] (rhoN) at (2,-1) {$\rho_N$};
        \draw [thick] (mu1) -- (mu1 -| U.west);
        \draw [thick] (mu2) -- (mu2 -| U.west);
        \draw [thick] (muN) -- (muN -| U.west);
        \draw [thick] (rho1) -- (rho1 -| U.east);
        \draw [thick] (rho2) -- (rho2 -| U.east);
        \draw [thick] (rhoN) -- (rhoN -| U.east);
    \end{tikzpicture}
    \;\approx\;
    \begin{tikzpicture} [baseline={([yshift=+0ex]current bounding box.center)}]
        \node[] (mu1) at (0,2) {$\mu_1$};
        \node[] (mu2) at (0,1) {$\mu_2$};
        \node[] (mudots) at (0,0) {$\vdots$};
        \node[] (muN) at (0,-1) {$\mu_N$};
        \node[rectangle, draw] (U1) at (1,2) {$\mathcal{U}^{[1]}$};
        \node[rectangle, draw] (U2) at (1,1) {$\mathcal{U}^{[2]}$};
        \node[] (Udots) at (1,0) {$\vdots$};
        \node[rectangle, draw] (UN) at (1,-1) {$\mathcal{U}^{[N]}$};
        \node[] (rho1) at (2,2) {$\rho_1$};
        \node[] (rho2) at (2,1) {$\rho_2$};
        \node[] (rhodots) at (2,0) {$\vdots$};
        \node[] (rhoN) at (2,-1) {$\rho_N$};
        \draw [thick] (U1) -- (U2) -- (Udots) -- (UN);
        \draw [thick] (mu1) -- (mu1 -| U1.west);
        \draw [thick] (mu2) -- (mu2 -| U2.west);
        \draw [thick] (muN) -- (muN -| UN.west);
        \draw [thick] (rho1) -- (rho1 -| U1.east);
        \draw [thick] (rho2) -- (rho2 -| U2.east);
        \draw [thick] (rhoN) -- (rhoN -| UN.east);
    \end{tikzpicture}
\end{align}
Finally, the isometry $T$ can be decomposed as a tree tensor network (TTN) with the following diagrammatic representation
\begin{align}
    T_{\rho,\sigma}\approx\;
    \begin{tikzpicture} [baseline={([yshift=+0ex]current bounding box.center)}]
        \node[] (rho1) at (0,3) {$\rho_1$};
        \node[] (rho2) at (0,2) {$\rho_2$};
        \node[] (rho3) at (0,1) {$\rho_3$};
        \node[] (rho4) at (0,0) {$\rho_4$};
        \node[] (rho5) at (0,-1) {$\rho_5$};
        \node[] (rho6) at (0,-2) {$\rho_6$};
        \node[] (rho7) at (0,-3) {$\rho_7$};
        \node[] (rho8) at (0,-4) {$\rho_8$};
        \node[trapezium, minimum height=8mm, shape border rotate=270, draw](T11) at (1,2.5) {$\mathcal{T}^{[1]}_{[1]}$};
        \node[trapezium, minimum height=8mm, shape border rotate=270, draw](T12) at (1,0.5) {$\mathcal{T}^{[2]}_{[1]}$};
        \node[trapezium, minimum height=8mm, shape border rotate=270, draw](T13) at (1,-1.5) {$\mathcal{T}^{[3]}_{[1]}$};
        \node[trapezium, minimum height=8mm, shape border rotate=270, draw](T14) at (1,-3.5) {$\mathcal{T}^{[4]}_{[1]}$};
        \node[trapezium, minimum height=8mm, shape border rotate=270, draw](T21) at (2.5,1.0) {$\mathcal{T}^{[1]}_{[2]}$};
        \node[trapezium, minimum height=8mm, shape border rotate=270, draw](T22) at (2.5,-2.0) {$\mathcal{T}^{[2]}_{[2]}$};
        \node[trapezium, minimum height=8mm, shape border rotate=270, draw](T31) at (4,-0.5) {$\mathcal{T}^{[1]}_{[3]}$};
        \node[] (sigma) at (5,-0.5) {$\sigma$};
        \draw [thick] (rho1) -- (rho1 -| T11.west);
        \draw [thick] (rho2) -- (rho2 -| T11.west);
        \draw [thick] (rho3) -- (rho3 -| T12.west);
        \draw [thick] (rho4) -- (rho4 -| T12.west);
        \draw [thick] (rho5) -- (rho5 -| T13.west);
        \draw [thick] (rho6) -- (rho6 -| T13.west);
        \draw [thick] (rho7) -- (rho7 -| T14.west);
        \draw [thick] (rho8) -- (rho8 -| T14.west);
        \draw [thick] (sigma) -- (sigma -| T31.east);
        \draw [thick] (T11.east) -- (T21);
        \draw [thick] (T12) -- (T12 -| T21.west);
        \draw [thick] (T13) -- (T13 -| T22.west);
        \draw [thick] (T14.east) -- (T22);
        \draw [thick] (T21.east) -- (T31);
        \draw [thick] (T22.east) -- (T31);
    \end{tikzpicture}
\end{align}
where the tensors $\mathcal{T}^{[\tau]}_{[\ell]}$ are isometric rank-3 tensors. Further details on the explicit construction of $U$, $V$ and $T$ can be found in the Supplementary Material.

Further numerical results on the effects of ED and bias on training are shown in Figs.~\ref{fig:MSE_BiasUnbias_wTrain_TN_Dloc3} and Figs.~\ref{fig:MSE_AllBiases_TN_Dloc3}, for models where only the matrix $V$ is decomposed as a tensor train. The numerical experiments presented here are conducted in the same way as discussed in Section \ref{sec:Training_FIM_Bias} in the main text, and confirm the conclusions drawn there: a lower ED is beneficial during training ($\Delta_{\mathrm{f-c}}\mathrm{MSE}_{\mathrm{min}}$ is positive) for models that are biased towards the data-generating function to be learned, whereas in the unbiased case a higher ED leads to better training performance.

\vspace{1cm}

\bibliography{biblio.bib}

\clearpage
\renewcommand{\thesection}{S\arabic{section}} % e.g., "S1, S2, S3" for Supplement
\setcounter{section}{0}                       % reset to 0 so first one is S1
\renewcommand{\thefigure}{S\arabic{figure}}   % optional: S1, S2 for figures
\setcounter{figure}{0}
\renewcommand{\thetable}{S\arabic{table}}     % optional: S1, S2 for tables
\setcounter{table}{0}

\section*{Supplementary Material}
\addcontentsline{toc}{section}{Supplementary Material}
%\tableofcontents

\section{Preliminaries: regression models and structure constants}
We quickly recap our definition of structure constants for regression models that we use in the main text as well as in the derivations presented in this document. We consider regression models taking as input a vector $\boldsymbol{x}\in\mathbb{R}^N$, with $N$ the number of input components, and parameterized by $M$ trainable parameters $\boldsymbol{\theta}\in\mathbb{R}^M$. We focus on a single real output, in the case where it can be expanded in a finite number $D$ and $K$ of inputs' and parameters' basis functions $e_{\mu}(\boldsymbol{x})$ and $\iota_{\nu}(\boldsymbol{\theta})$, respectively, as
\begin{equation}
    f_{\boldsymbol{\theta}}(\boldsymbol{x})=\sum_{\mu=1}^D\sum_{\nu=1}^K \Gamma_{\mu,\nu}\,e_{\mu}(\boldsymbol{x})\,\iota_{\nu}(\boldsymbol{\theta}) \,\,.
    \label{eq:general_model_def_S}
\end{equation}
The coefficients $\Gamma\in\mathbb{R}^{D\times K}$ are called \emph{structure constants} of the model, while the basis functions $e_{\mu}(\boldsymbol{x})$ are taken to form orthonormal bases in the $D$ and $K$-dimensional spaces of input and parameters functions, respectively, i.e.,
\begin{equation}
    \mathbb{E}_{\boldsymbol{x}\sim p}\big[e_{\mu}(\boldsymbol{x})e_{\mu'}(\boldsymbol{x})\big]=\int e_{\mu}(\boldsymbol{x})e_{\mu'}(\boldsymbol{x})\,p(\boldsymbol{x})\,\mathrm{d}\boldsymbol{x}=\delta_{\mu,\mu'} \,\,,
    \label{eq:ortho_basis_funs_inputs_S}
\end{equation}
with $p(\boldsymbol{x})$ the probability density function for the inputs, and
\begin{equation}
    \frac{1}{V_{\Theta}}\int_{\Theta} \iota_{\nu}(\boldsymbol{\theta})\iota_{\nu'}(\boldsymbol{\theta})\,\mathrm{d}\boldsymbol{\theta}=\delta_{\nu,\nu'} \,\,,
    \label{eq:ortho_basis_funs_pars_S}
\end{equation}
with $\Theta$ denoting the parameter space and $V_{\Theta}$ its volume. As discussed in the main text and later in this document, for a broad class of quantum neural networks (QNNs) we have $e_{\mu}(\boldsymbol{x})\equiv e_{(\mu_1,...,\mu_N)}(\boldsymbol{x})=\prod_{n=1}^N e_{\mu_n}^{(n)}(x_n)$ and $\iota_{\nu}(\boldsymbol{\theta})\equiv\iota_{(\nu_1,...,\nu_M)}(\boldsymbol{\theta})=\prod_{m=1}^M\iota_{\nu_m}^{(m)}(\theta_m)$, where
\begin{align}
    & e_{\mu_n}^{(n)}(x_n)\in\mathcal{B}_n=\{1,\,\sqrt{2}\cos(\omega_nx_n),\,\sqrt{2}\sin(\omega_nx_n)\}_{\omega_n\in\Omega_n} \,\,, \label{eq:input_basis_funs_Fourier_S}\\
    & \iota_{\nu_m}^{(m)}(\theta_m)\in\tilde{\mathcal{B}}_m=\{1,\,\sqrt{2}\cos(\tilde{\omega}_m\theta_m),\,\sqrt{2}\sin(\tilde{\omega}_m\theta_m)\}_{\tilde{\omega}_m\in\tilde{\Omega}_m} \label{eq:param_basis_funs_Fourier_S}\,\,,
\end{align}
normalized in the interval $[-\pi,\pi]$, with $\Omega_n$ and $\tilde{\Omega}_m$ finite sets of frequencies the QNN has access to. In the common situation where the inputs are encoded multiple times via the re-uploading technique \cite{Schuld2021,PerezSalinas2020}, and both input features and parameters are encoded as angles of single qubit rotations, the sets $\Omega_n$ and $\tilde{\Omega}_m$ comprise only integer frequencies and read as $\Omega_n=\{1,...,L\}$ and $\tilde{\Omega}_m=\{1\}$, with $L$ being the number of times the input features $x_n$ are uploaded.

The structure constants $\Gamma_{\mu,\nu}$ can be viewed as elements of a $D\times K$ real matrix $\Gamma$ which admits the following singular value decomposition (SVD)
\begin{equation}
    \Gamma=USV^{\top} \,\,,
    \label{eq:SVD_gamma_S}
\end{equation}
where $U$ is a $D\times D$ real orthogonal matrix satisfying with $U^{\top}U=UU^{\top}=I_D$, $S=\mathrm{diag}(s_1,...,s_D)$ is a $D\times D$ diagonal positive semi-definite matrix (with diagonal ordered as $s_1\geq s_2\geq ... \geq s_D$), and $V$ is a $K\times D$ real matrix with orthonormal columns, i.e., $V^{\top}V=I_D$. The singular values $s_{\rho}$ control the correlations between the parameter space and the functions in the input space, and the set $\{s_{\rho}\}_{\rho}$ is therefore referred to as \emph{correlation spectrum}.

\section{Basis functions and structure constants of QNNs}
\begin{figure*}
    \centering
    \includegraphics[width=0.6\linewidth]{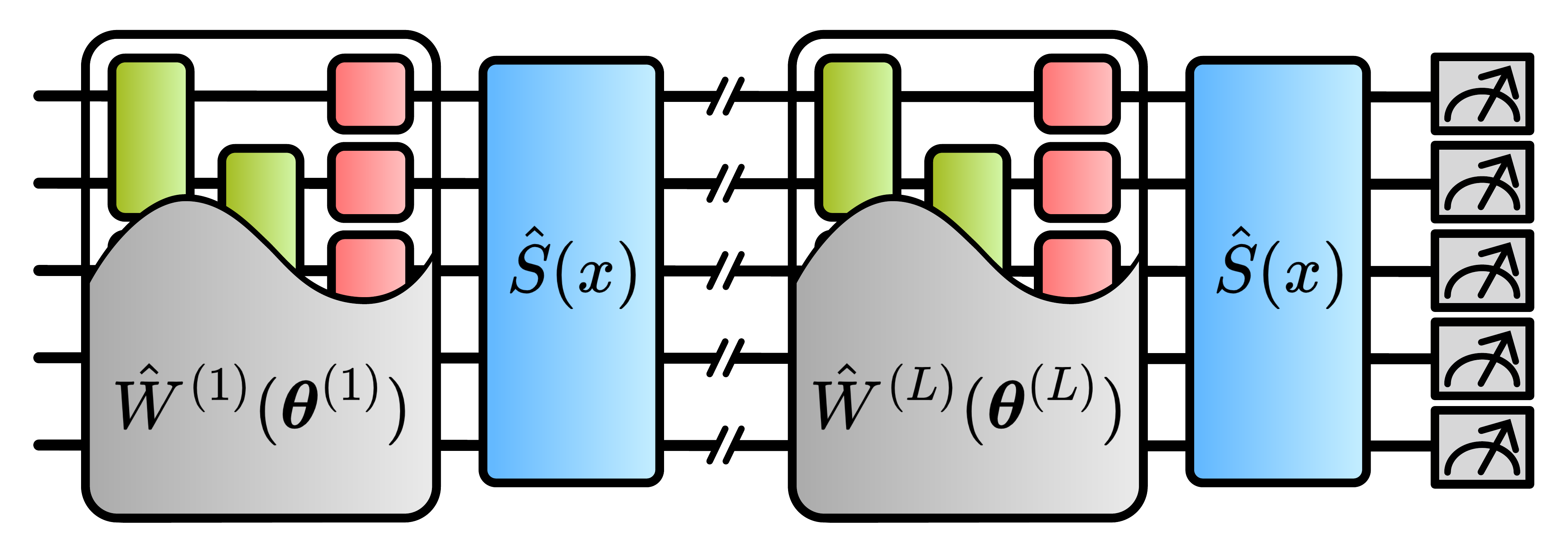}
    \caption{Schematics of the parameterized quantum circuit used as a QNN.}
    \label{figapp:Figure_QNN} 
\end{figure*}
We now provide the explicit connection between the form of regression models introduced before and the outputs of quantum neural networks (QNNs). We define a QNN as function $f_{\boldsymbol{\theta}}(\boldsymbol{x})$ defined by a parameterized quantum circuit (PQC) as \cite{Farhi2018,Schuld2021}
\begin{equation}
    f_{\boldsymbol{\theta}}(\boldsymbol{x})=\bra{0}\hat{U}^{\dagger}_{\boldsymbol{\theta}}(\boldsymbol{x})\hat{M}\,\hat{U}_{\boldsymbol{\theta}}(\boldsymbol{x})\ket{0}\equiv\bra{\psi_{\boldsymbol{\theta}}(\boldsymbol{x})}\hat{M}\ket{\psi_{\boldsymbol{\theta}}(\boldsymbol{x})} \,\,,
    \label{eq:QNN_out}
\end{equation}
where $\hat{U}_{\boldsymbol{\theta}}(\boldsymbol{x})$ is the unitary operator implemented by the PQC, acting on a $\mathcal{D}$-dimensional Hilbert space, and $\hat{M}$ is an observable whose expectation value over the parameterized state $\ket{\psi_{\boldsymbol{\theta}}(\boldsymbol{x})}$ corresponds to the QNN output. We consider a general PQC implementing $L$ layers of data re-uploading  \cite{PerezSalinas2020,Schuld2021} with the following unitary (see Fig.~\ref{figapp:Figure_QNN} for a visualization)
\begin{equation}
    \hat{U}_{\boldsymbol{\theta}}(\boldsymbol{x})=\prod_{\ell=1}^L\Big[\hat{S}(\boldsymbol{x})\,\hat{W}^{(\ell)}(\boldsymbol{\theta}^{(\ell)})\Big] \,\,.
\end{equation}
The $\hat{S}(\boldsymbol{x})$ is a unitary operator where the components of the input datum $\boldsymbol{x}$ are encoded as angles in rotation gates. The $\hat{W}^{(\ell)}(\boldsymbol{\theta}^{(\ell)})$ are variational blocks, where the trainable parameters $\boldsymbol{\theta}^{(\ell)}$ are also encoded as angles in rotation gates. The vector $\boldsymbol{\theta}$ summarizes the dependence on all $\boldsymbol{\theta}^{(\ell)}$, i.e., $\boldsymbol{\theta}=\{\boldsymbol{\theta}^{(\ell)}\}_{\ell=1,...,L}$.

\subsection{Fourier series representation of QNN} \label{app:FourierSeriesQNNs_S}
We now show that QNN outputs as Eq.~\eqref{eq:QNN_out} can be expressed as Eq.~\eqref{eq:general_model_def_S} with $e_{\mu}(\boldsymbol{x})$ and $\iota_{\nu}(\boldsymbol{\theta})$ trigonometric basis functions given by Eqs.~\eqref{eq:input_basis_funs_Fourier_S} and \eqref{eq:param_basis_funs_Fourier_S}. We show here for simplicity the case of a one-dimensional input $x$, while this can be easily generalized to more input features \cite{Schuld2021,Casas2023} following the same derivation. Without loss of generality, we can write the variational blocks as
\begin{equation}
    \hat{W}^{(\ell)}(\boldsymbol{\theta}^{(\ell)})=\prod_{j_{\ell}=1}^{J_{\ell}}\hat{G}^{(\ell)}_{j_{\ell}}(\theta^{(\ell)}_{j_{\ell}}) \,\,,
\end{equation}
where the dependence on the single parameters $\theta^{(\ell)}_{j_{\ell}}$, the components of $\boldsymbol{\theta}^{(\ell)}$, has been split into individual operators $\hat{G}^{(\ell)}_{j_{\ell}}$, as it is typically the case for QNNs where each parameter controls one rotation gate. We then switch to the diagonal representation of $\hat{S}(x)$ and $\hat{G}^{(\ell)}_{j_{\ell}}(\theta^{(\ell)}_{j_{\ell}})$
\begin{equation}
    \hat{S}(x)=\hat{V}\,\hat{\Sigma}(x)\,\hat{V}^{\dagger} \,\,;\quad \hat{G}^{(\ell)}_{j_{\ell}}(\theta^{(\ell)}_{j_{\ell}})=\hat{Q}^{(\ell)}_{j_{\ell}}\,\hat{\Gamma}^{(\ell)}_{j_{\ell}}(\theta^{(\ell)}_{j_{\ell}})\,\hat{Q}^{(\ell)\,\dagger}_{j_{\ell}} \,\,,
\end{equation}
with $\hat{\Sigma}(x)=\mathrm{diag}(\{\mathrm{e}^{-\mathrm{i}\lambda_{\alpha}x}\}_{\alpha=1,...,\mathcal{D}})$ and $\hat{\Gamma}^{(\ell)}_{j_{\ell}}(\theta^{(\ell)}_{j_{\ell}})=\mathrm{diag}(\{\mathrm{e}^{-\mathrm{i}\eta^{(\ell,j_{\ell})}_{\beta}\theta^{(\ell)}_{j_{\ell}}}\}_{\beta=1,...,\mathcal{D}})$. To ease the notation, we count the parameters with the index $m$ replacing the labels $(\ell)$ and $j_{\ell}$, i.e., $\theta^{(\ell)}_{j_{\ell}}\to\theta_m$, $\hat{Q}^{(\ell)}_{j_{\ell}}\to \hat{Q}_m$ and $\hat{\Gamma}^{(\ell)}_{j_{\ell}}\to \hat{\Gamma}_m$. Then,
\begin{equation}
    \begin{split}
        \hat{U}_{\boldsymbol{\theta}}(x)=&\,\hat{V}\,\hat{\Sigma}(x)\,\hat{V}^{\dagger}\,\prod_{m=J_1+...+J_{L-1}+1}^{J_L}\hat{Q}_m\,\hat{\Gamma}_m(\theta_m)\,\hat{Q}_m^{\dagger}\,\times\\
        &\hat{V}\,\hat{\Sigma}(x)\,\hat{V}^{\dagger}\,\prod_{m=J_1+...+J_{L-2}+1}^{J_{L-1}}\hat{Q}_m\,\hat{\Gamma}_m(\theta_m)\,\hat{Q}_m^{\dagger}\,\times\\
        &...\,\times\\
        &\hat{V}\,\hat{\Sigma}(x)\,\hat{V}^{\dagger}\,\prod_{m=1}^{J_{1}}\hat{Q}_m\,\hat{\Gamma}_m(\theta_m)\,\hat{Q}_m^{\dagger} \,\,.
    \end{split}
\end{equation}
The components $\psi_i(x;\boldsymbol{\theta})$ of the state $\ket{\psi_{\boldsymbol{\theta}}(x)}=\hat{U}_{\boldsymbol{\theta}}(x)\ket{0}$ correspond to the elements $\big[\hat{U}_{\boldsymbol{\theta}}(x)\big]_{i,1}$ of the unitary $\hat{U}_{\boldsymbol{\theta}}(x)$. In order to arrive at the Fourier series expansion of $\psi_i(x;\boldsymbol{\theta})$ we start by considering the Fourier expansion of the following sequence of unitaries:
\begin{equation}
    \begin{split}
        \big[\hat{Q}_2\,&\hat{\Gamma}_2(\theta_2)\,\hat{Q}_2^{\dagger}\,\hat{Q}_1\,\hat{\Gamma}_1(\theta_1)\,\hat{Q}_1^{\dagger}\big]_{k,1}=\sum_{\beta_1}\big[\hat{Q}_2\,\hat{\Gamma}_2(\theta_2)\,\hat{Q}_2^{\dagger}\,\hat{Q}_1\,\hat{\Gamma}_1(\theta_1)\big]_{k,\beta_1}\big[\hat{Q}_1^{\dagger}\big]_{\beta_1,1}\\
        &=\sum_{\beta_1}\big[\hat{Q}_2\,\hat{\Gamma}_2(\theta_2)\,\hat{Q}_2^{\dagger}\,\hat{Q}_1\big]_{k,\beta_1}\big[\hat{Q}_1^{\dagger}\big]_{\beta_1,1}\mathrm{e}^{-\mathrm{i}\eta^{(1)}_{\beta_1}\theta_1}\\
        &=\sum_{\beta_1}\sum_{\gamma_1}\big[\hat{Q}_2\,\hat{\Gamma}_2(\theta_2)\,\hat{Q}_2^{\dagger}\big]_{k,\gamma_1}\big[\hat{Q}_1\big]_{\gamma_1,\beta_1}\big[\hat{Q}_1^{\dagger}\big]_{\beta_1,1}\mathrm{e}^{-\mathrm{i}\eta^{(1)}_{\beta_1}\theta_1}\\
        &=\sum_{\beta_1,\beta_2}\sum_{\gamma_1}\big[\hat{Q}_2\,\hat{\Gamma}_2(\theta_2)\big]_{k,\beta_2}\big[\hat{Q}_2^{\dagger}\big]_{\beta_2,\gamma_1}\big[\hat{Q}_1\big]_{\gamma_1,\beta_1}\big[\hat{Q}_1^{\dagger}\big]_{\beta_1,1}\mathrm{e}^{-\mathrm{i}\eta^{(1)}_{\beta_1}\theta_1}\\
        &=\sum_{\beta_1,\beta_2}\sum_{\gamma_1}\big[\hat{Q}_2\big]_{k,\beta_2}\big[\hat{Q}_2^{\dagger}\big]_{\beta_2,\gamma_1}\big[\hat{Q}_1\big]_{\gamma_1,\beta_1}\big[\hat{Q}_1^{\dagger}\big]_{\beta_1,1}\mathrm{e}^{-\mathrm{i}\eta^{(1)}_{\beta_1}\theta_1}\mathrm{e}^{-\mathrm{i}\eta^{(2)}_{\beta_2}\theta_2}\\
        &\equiv\sum_{\beta_1,\beta_2}q_{k,1}^{(\beta_1,\beta_2)}\mathrm{e}^{-\mathrm{i}\eta^{(1)}_{\beta_1}\theta_1}\mathrm{e}^{-\mathrm{i}\eta^{(2)}_{\beta_2}\theta_2} \,\,,
    \end{split}
\end{equation}
with $q_{k,1}^{(\beta_1,\beta_2)}\equiv\sum_{\gamma_1}\big[\hat{Q}_2\big]_{k,\beta_2}\big[\hat{Q}_2^{\dagger}\big]_{\beta_2,\gamma_1}\big[\hat{Q}_1\big]_{\gamma_1,\beta_1}\big[\hat{Q}_1^{\dagger}\big]_{\beta_1,1}$. This can be iterated over longer sequences, e.g.,
\begin{equation}
    \begin{split}
        &\big[\hat{Q}_3\,\hat{\Gamma}_3(\theta_3)\,\hat{Q}_3^{\dagger}\,\hat{Q}_2\,\hat{\Gamma}_2(\theta_2)\,\hat{Q}_2^{\dagger}\,\hat{Q}_1\,\hat{\Gamma}_1(\theta_1)\,\hat{Q}_1^{\dagger}\big]_{k,1}\\
        &=\sum_{\beta_1,\beta_2}\sum_{\gamma_2}\big[\hat{Q}_3\,\hat{\Gamma}_3(\theta_3)\,\hat{Q}_3^{\dagger}\big]_{k,\gamma_2}q_{\gamma_2,1}^{(\beta_1,\beta_2)}\mathrm{e}^{-\mathrm{i}\eta^{(1)}_{\beta_1}\theta_1}\mathrm{e}^{-\mathrm{i}\eta^{(2)}_{\beta_2}\theta_2}\\
        &=\sum_{\beta_1,\beta_2,\beta_3}\sum_{\gamma_2}\big[\hat{Q}_3\,\hat{\Gamma}_3(\theta_3)\big]_{k,\beta_3}\big[\hat{Q}_3^{\dagger}\big]_{\beta_3,\gamma_2}q_{\gamma_2,1}^{(\beta_1,\beta_2)}\mathrm{e}^{-\mathrm{i}\eta^{(1)}_{\beta_1}\theta_1}\mathrm{e}^{-\mathrm{i}\eta^{(2)}_{\beta_2}\theta_2}\\
        &=\sum_{\beta_1,\beta_2,\beta_3}\sum_{\gamma_2}\big[\hat{Q}_3\big]_{k,\beta_3}\big[\hat{Q}_3^{\dagger}\big]_{\beta_3,\gamma_2}q_{\gamma_2,1}^{(\beta_1,\beta_2)}\mathrm{e}^{-\mathrm{i}\eta^{(1)}_{\beta_1}\theta_1}\mathrm{e}^{-\mathrm{i}\eta^{(2)}_{\beta_2}\theta_2}\mathrm{e}^{-\mathrm{i}\eta^{(3)}_{\beta_3}\theta_3}\\
        &\equiv\sum_{\beta_1,\beta_2,\beta_3}q_{k,1}^{(\beta_1,\beta_2,\beta_3)}\mathrm{e}^{-\mathrm{i}\eta^{(1)}_{\beta_1}\theta_1}\mathrm{e}^{-\mathrm{i}\eta^{(2)}_{\beta_2}\theta_2}\mathrm{e}^{-\mathrm{i}\eta^{(3)}_{\beta_3}\theta_3} \,\,.
    \end{split}
\end{equation}
Thus, iterating this calculation over all operators in the expression for $\psi_i(x;\boldsymbol{\theta})=\big[\hat{U}_{\boldsymbol{\theta}}(x)\big]_{i,1}$ (including also the $\hat{V}$ and $\hat{\Sigma}(x)$ operators from the encoding), we arrive at
\begin{equation}
    \psi_i(x;\boldsymbol{\theta})=\sum_{\boldsymbol{\alpha}}\sum_{\boldsymbol{\beta}}q_{i,1}^{(\boldsymbol{\alpha};\boldsymbol{\beta})}\,\mathrm{e}^{-\mathrm{i}\Lambda_{\boldsymbol{\alpha}}x}\,\mathrm{e}^{-\mathrm{i}\boldsymbol{\eta}_{\boldsymbol{\beta}}\cdot\boldsymbol{\theta}} \,\,,
\end{equation}
where we adopt the shorthand notation $\boldsymbol{\alpha}=(\alpha_1,...,\alpha_L)$, $\boldsymbol{\beta}=(\beta_1,...,\beta_M)$, $\Lambda_{\boldsymbol{\alpha}}=\sum_{\ell}\lambda_{\alpha_{\ell}}$ and $\boldsymbol{\eta}_{\boldsymbol{\beta}}=(\eta^{(1)}_{\beta_1},...,\eta^{(M)}_{\beta_M})$. The QNN output can be therefore expanded as
\begin{equation}
    \begin{split}
        f_{\boldsymbol{\theta}}(x)&=\bra{\psi_{\boldsymbol{\theta}}(x)}\hat{M}\ket{\psi_{\boldsymbol{\theta}}(x)}=\sum_{i,j}M_{i,j}\,\psi^*_i(x;\boldsymbol{\theta})\psi_j(x;\boldsymbol{\theta})\\
        &=\sum_{\boldsymbol{\alpha},\boldsymbol{\alpha}'}\sum_{\boldsymbol{\beta},\boldsymbol{\beta}'}\mathrm{e}^{\mathrm{i}(\Lambda_{\boldsymbol{\alpha}}-\Lambda_{\boldsymbol{\alpha}'})x}\,\mathrm{e}^{\mathrm{i}(\boldsymbol{\eta}_{\boldsymbol{\beta}}-\boldsymbol{\eta}_{\boldsymbol{\beta}'})\cdot\boldsymbol{\theta}}\sum_{i,j}M_{i,j}\,q_{i,1}^{(\boldsymbol{\alpha};\boldsymbol{\beta})*}q_{j,1}^{(\boldsymbol{\alpha}';\boldsymbol{\beta}')}\\
        &\equiv\sum_{\boldsymbol{\alpha},\boldsymbol{\alpha}'}\sum_{\boldsymbol{\beta},\boldsymbol{\beta}'}\tilde{\Gamma}_{(\boldsymbol{\alpha};\boldsymbol{\alpha}'),(\boldsymbol{\beta};\boldsymbol{\beta}')}\,\mathrm{e}^{\mathrm{i}(\Lambda_{\boldsymbol{\alpha}}-\Lambda_{\boldsymbol{\alpha}'})x}\,\mathrm{e}^{\mathrm{i}(\boldsymbol{\eta}_{\boldsymbol{\beta}}-\boldsymbol{\eta}_{\boldsymbol{\beta}'})\cdot\boldsymbol{\theta}} \,\,,
    \end{split}
    \label{eq:QNN_output_fourier_S}
\end{equation}
with $\tilde{\Gamma}_{(\boldsymbol{\alpha};\boldsymbol{\alpha}'),(\boldsymbol{\beta};\boldsymbol{\beta}')}\equiv\sum_{i,j}M_{i,j}\,q_{i,1}^{(\boldsymbol{\alpha};\boldsymbol{\beta})*}q_{j,1}^{(\boldsymbol{\alpha}';\boldsymbol{\beta}')}$. We now denote the sets of frequencies generated by the encoding and variational gates as
\begin{equation}
    \Omega=\{\Lambda_{\boldsymbol{\alpha}}-\Lambda_{\boldsymbol{\alpha}'},\,\boldsymbol{\alpha},\boldsymbol{\alpha}'\in[\mathcal{D}]^L\} \,\,,
\end{equation}
and
\begin{equation}
    \tilde{\Omega}_m=\{\eta^{(m)}_{\beta_m}-\eta^{(m)}_{\beta_m'},\,\beta_m,\beta_m'=1,...,\mathcal{D}\} \,\,.
\end{equation}
We can therefore write
\begin{equation}
    f_{\boldsymbol{\theta}}(x)=\sum_{\omega\in\Omega}\sum_{\tilde{\omega}_1\in\tilde{\Omega}_1}...\sum_{\tilde{\omega}_M\in\tilde{\Omega}_M}\tilde{\Gamma}_{\omega,(\tilde{\omega}_1,...,\tilde{\omega}_M)}\,\mathrm{e}^{\mathrm{i}\omega x}\,\prod_{m=1}^M\mathrm{e}^{\mathrm{i}\tilde{\omega}_m\theta_m}
\end{equation}
and, after expressing the Fourier components in terms of real trigonometric basis functions and normalizing them in the chosen input and parameter space, we arrive at the desired form of Eqs.~\eqref{eq:general_model_def_S}, \eqref{eq:input_basis_funs_Fourier_S} and \eqref{eq:param_basis_funs_Fourier_S}.

\subsection{Dependence of structure constants on QNN entangling layers} \label{app:basis_QNN_entanglinglayers}
\begin{figure*}
    \centering
    \includegraphics[width=\linewidth]{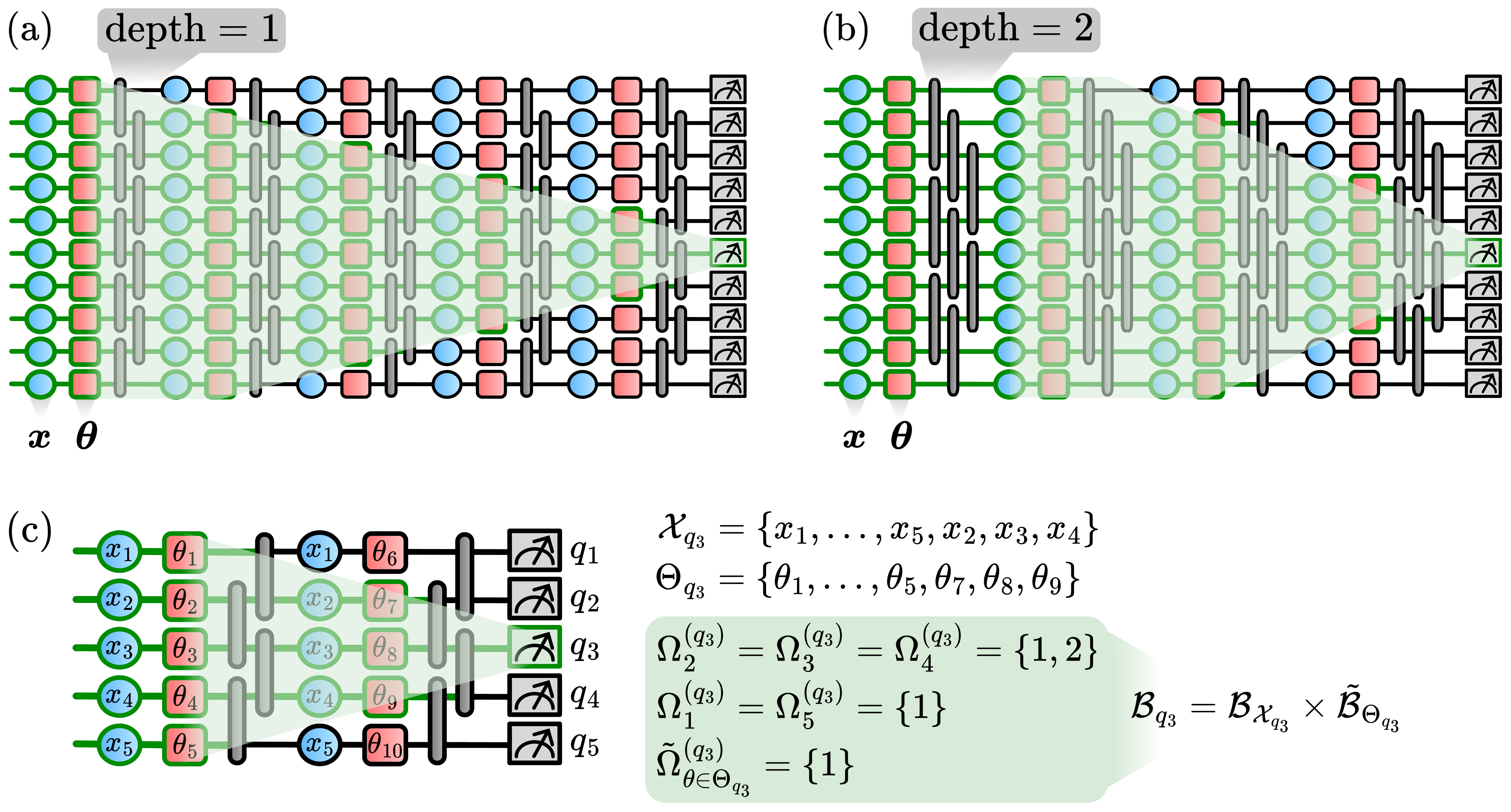}
\caption{Schematics of the construction of the backward light-cones (BWLCs) for calculating the admissible QNN basis states. The blue and red single-qubit gates are the gates where the input features $x_n$ and the parameters $\theta_m$ are encoded, respectively. The gray gates are multi-qubit entangling gates. The green shaded areas denote the extension of the BWLC from a given measured qubit. The gates belonging to the BWLC are framed in green. (a) For entangling layers coupling only nearest-neighbor qubits on a chain (depth one), the BWLC increases by two qubit lines for every layer of entangling gates. (b) For entangling layers coupling up to next-nearest-neighbor qubits on a chain (depth two), the BWLC spreads faster, i.e., by four qubit lines for every layer of entangling gates. (c) Explicit example of construction of BWLC for the third qubit of a five-qubits QNN. $\mathcal{X}_{q_3}$ and $\Theta_{q_3}$ are the sets of input features and parameters contained in the BWLC, respectively, and include how many times a given feature or parameter appears. The set of Fourier frequencies $\Omega_n^{(q_3)}$ capture this multiplicity (i.e., $\Omega_n^{(q_3)}=\{1,...,L\}$ if $x_n$ appears $L$ times in $\mathcal{X}_{q_3}$). The sets of frequencies $\Omega_n^{(q_3)}$ and $\tilde{\Omega}_{\theta}^{(q_3)}$ define the basis functions $\mathcal{B}_{\mathcal{X}_{q_3}}$ and $\tilde{\mathcal{B}}_{\Theta_{q_3}}$. The functions obtained by measuring $q_3$ belong to the product space $\mathrm{span}(\mathcal{B}_{\mathcal{X}_{q_3}})\otimes\mathrm{span}(\tilde{\mathcal{B}}_{\Theta_{q_3}})$.}
    \label{figapp:Figure_QNN_BWLC} 
\end{figure*}
Here we discuss how the structure of the entangling layers in a QNN influences the space of functions the model has access to. For simplicity, we consider the situation where input features and variational parameters are encoded as angles of single qubit rotations of the form $\mathrm{e}^{-\mathrm{i}\frac{\phi}{2}\boldsymbol{n}\cdot\hat{\boldsymbol{\sigma}}}$ (with $\phi$ being the feature/parameter to be encoded, $\boldsymbol{n}$ an arbitrary rotation axis and $\hat{\boldsymbol{\sigma}}$ the vector of Pauli matrices). We define an \emph{encoding layer} to be the sequence of single-qubit gates encoding the input features on all qubits as
\begin{equation}
    \hat{S}(\boldsymbol{x})=\prod_{n=1}^{N}\mathrm{e}^{-\mathrm{i}\frac{x_n}{2}\boldsymbol{n}\cdot\hat{\boldsymbol{\sigma}}_n} \,\,,
\end{equation}
where we assume for simplicity that the number of qubits equals the number of features $N$. Similarly, we define a \emph{variational layer} to be the sequence of single-qubit gates encoding $N$ variational parameters $\boldsymbol{\theta}^{(\ell)}$ on all qubits as
\begin{equation}
    \hat{R}(\boldsymbol{\theta}^{(\ell)})=\prod_{n=1}^{N}\mathrm{e}^{-\mathrm{i}\frac{\theta^{(\ell)}_n}{2}\boldsymbol{n}\cdot\hat{\boldsymbol{\sigma}}_n} \,\,.
\end{equation}
When using the data re-uploading technique, encoding and variational blocks are repeated multiple times and interleaved with \emph{entangling layers} $\hat{V}$, i.e., sequences of multi-qubit gates. We assume these gates to be fixed, i.e., not parameterized by any variational parameter. At the end of the sequence, a (optional) measurement unitary $\hat{W}_M$ is performed, accounting for the rotation to the measurement basis (assuming for simplicity only one measurement basis is used). The resulting unitary is then
\begin{equation}
    \hat{U}_{\boldsymbol{\theta}}(\boldsymbol{x})=\prod_{\ell=1}^L\Big[\hat{S}(\boldsymbol{x})\,\hat{R}(\boldsymbol{\theta}^{(\ell)})\,\hat{V}\Big]\hat{W}_M \,\,,
\end{equation}
which corresponds to the structure discussed previously setting $\hat{W}^{(\ell)}(\boldsymbol{\theta}^{(\ell)})=\hat{R}(\boldsymbol{\theta}^{(\ell)})\,\hat{V}$.
We consider the situation where (after the unitary $\hat{W}_M$), all qubits are measured in a fixed basis, e.g., the $z$ basis, and the outputs summed together to form
\begin{equation}
    f_{\boldsymbol{\theta}}(\boldsymbol{x})=\sum_{n=1}^{N}\bra{0}\hat{U}^{\dagger}_{\boldsymbol{\theta}}(\boldsymbol{x})\,\hat{\sigma}^z_n\,\hat{U}_{\boldsymbol{\theta}}(\boldsymbol{x})\ket{0}\,\,,
\end{equation}
which corresponds to the structure discussed previously setting $\hat{M}=\sum_{n=1}^{N}\hat{\sigma}^z_n$. Our goal here is to investigate how the structure of the entangling operations $\hat{V}$ and potentially $\hat{W}_M$ influences the structure constants $\Gamma$ of the model. More specifically, we define the \emph{depth} of the entangling (and measurement) layers as the number of qubits that are entangled to a given one after applying the layer to a product state (i.e., on a linear chain, depth one corresponds to the layer consisting of nearest-neighbor two-qubit gates, depth two to next-nearest-neighbor three-qubit gates, and so on), and we study how the depth affects the basis functions $e_{\mu}(\boldsymbol{x})$ and $\iota_{\nu}(\boldsymbol{\theta})$ accessible to the model.

To do this, we introduce the concept of \emph{backward light-cone} (BWLC) of a given measured qubit (see Fig.~\ref{figapp:Figure_QNN_BWLC} for a schematic representation). The BWLC of a measured qubit $q_n$ is the set of input features $\mathcal{X}_{q_n}$ and the set of parameters $\Theta_{q_n}$ the one-qubit reduced density matrix for $q_n$ depends on, defined as
\begin{equation}
    \hat{\varrho}_{q_n}(\boldsymbol{x},\boldsymbol{\theta})=\mathrm{tr}_{\overline{q_n}}\big[\hat{U}_{\boldsymbol{\theta}}(\boldsymbol{x})\ket{0}\bra{0}\hat{U}^{\dagger}_{\boldsymbol{\theta}}(\boldsymbol{x})\big] \,\,.
\end{equation}
Input features and parameters encoded via gates outside the BWLC cannot influence $\hat{\varrho}_{q_n}(\boldsymbol{x},\boldsymbol{\theta})$, since their effect has not had `time' to be propagated to $q_n$ via the entangling gates. Importantly, $\mathcal{X}_{q_n}$ and $\Theta_{q_n}$ are defined to also account for the multiplicity a given feature or parameter appears in the BWLC. These therefore allow to define the sets of Fourier frequencies $\Omega_{x}^{(q_n)}$ (for $x\in\mathcal{X}_{q_n}$) and $\tilde{\Omega}_{\theta}^{(q_n)}$ (for $\theta\in\Theta_{q_n}$) for the Fourier series expansion of the elements of $\hat{\varrho}_{q_n}(\boldsymbol{x},\boldsymbol{\theta})$, and therefore of the expectation value $\bra{0}\hat{U}^{\dagger}_{\boldsymbol{\theta}}(\boldsymbol{x})\,\hat{\sigma}^z_n\,\hat{U}_{\boldsymbol{\theta}}(\boldsymbol{x})\ket{0}$. For instance, if a given input feature $x\in\mathcal{X}_{q_n}$ is contained $\ell$ times in $\mathcal{X}_{q_n}$, then $\Omega_{x}^{(q_n)}=\{1,...,\ell\}$. The sets of frequencies $\Omega_{x}^{(q_n)}$ and $\tilde{\Omega}_{\theta}^{(q_n)}$ define the basis functions $\mathcal{B}_{x}^{(q_n)}$ and $\tilde{\mathcal{B}}_{\theta}^{(q_n)}$
\begin{align}
    & \mathcal{B}_{x}^{(q_n)}=\{1,\,\sqrt{2}\cos(\omega x),\,\sqrt{2}\sin(\omega x)\}_{\omega\in\Omega_{x}^{(q_n)}} \,\,, \\
    & \tilde{\mathcal{B}}_{\theta}^{(q_n)}=\{1,\,\sqrt{2}\cos(\tilde{\omega}\theta),\,\sqrt{2}\sin(\tilde{\omega}\theta)\}_{\tilde{\omega}\in\tilde{\Omega}_{\theta}^{(q_n)}} \,\,,
\end{align}
which can then be used to define inputs' and parameters' basis functions $\mathcal{B}_{\mathcal{X}_{q_n}}$ and $\tilde{\mathcal{B}}_{\Theta_{q_n}}$ for $\hat{\varrho}_{q_n}(\boldsymbol{x},\boldsymbol{\theta})$ as
\begin{equation}
    \mathcal{B}_{\mathcal{X}_{q_n}}=\bigtimes_{x\in\mathcal{X}_{q_n}}\mathcal{B}_{x}^{(q_n)} \quad\mathrm{and}\quad\tilde{\mathcal{B}}_{\Theta_{q_n}}=\bigtimes_{\theta\in\Theta_{q_n}}\tilde{\mathcal{B}}_{\theta}^{(q_n)} \,\,,
\end{equation}
i.e., as the product sets from all single-input (single-parameter) basis functions in the BWLC. The elements of $\hat{\varrho}_{q_n}(\boldsymbol{x},\boldsymbol{\theta})$, as well as the functions obtained by measuring $q_n$, belong to the product space $\mathrm{span}(\mathcal{B}_{\mathcal{X}_{q_n}})\otimes\mathrm{span}(\tilde{\mathcal{B}}_{\Theta_{q_n}})$, with basis set given by
\begin{equation}
    \mathcal{B}_{q_n}=\mathcal{B}_{\mathcal{X}_{q_n}}\times\tilde{\mathcal{B}}_{\Theta_{q_n}} \,\,.
\end{equation}
Importantly, the number of basis elements $|\mathcal{B}_{\mathcal{X}_{q_n}}|\leq(2L+1)^N$, and similarly $|\mathcal{B}_{\mathcal{X}_{q_n}}|\leq 3^M$, since in general the BWLC from $q_n$ does not contain all encoding and variational gates. More specifically, the smaller the depth of the entangling layers (and of the measurement operator), the `slower' the BWLC spreads, i.e., the smaller the sizes of $\mathcal{X}_{q_n}$ and $\Theta_{q_n}$, and therefore $|\mathcal{B}_{\mathcal{X}_{q_n}}|$ and $|\mathcal{B}_{\mathcal{X}_{q_n}}|$, become. This is schematically shown in panels (a) and (b) of Fig.~\ref{figapp:Figure_QNN_BWLC}.

Since the QNN output is given by the sum of all $\bra{0}\hat{U}^{\dagger}_{\boldsymbol{\theta}}(\boldsymbol{x})\,\hat{\sigma}^z_n\,\hat{U}_{\boldsymbol{\theta}}(\boldsymbol{x})\ket{0}$, the basis functions it has access to are given by
\begin{equation}
    \mathcal{B}=\bigcup_{n=1}^N\mathcal{B}_{q_n} \,\,,
\end{equation}
which, depending on the depth of the entangling and measurement layers, may have less elements than the maximum number $(2L+1)^N\times 3^M$.

\begin{figure*}
    \centering
    \includegraphics[width=\linewidth]{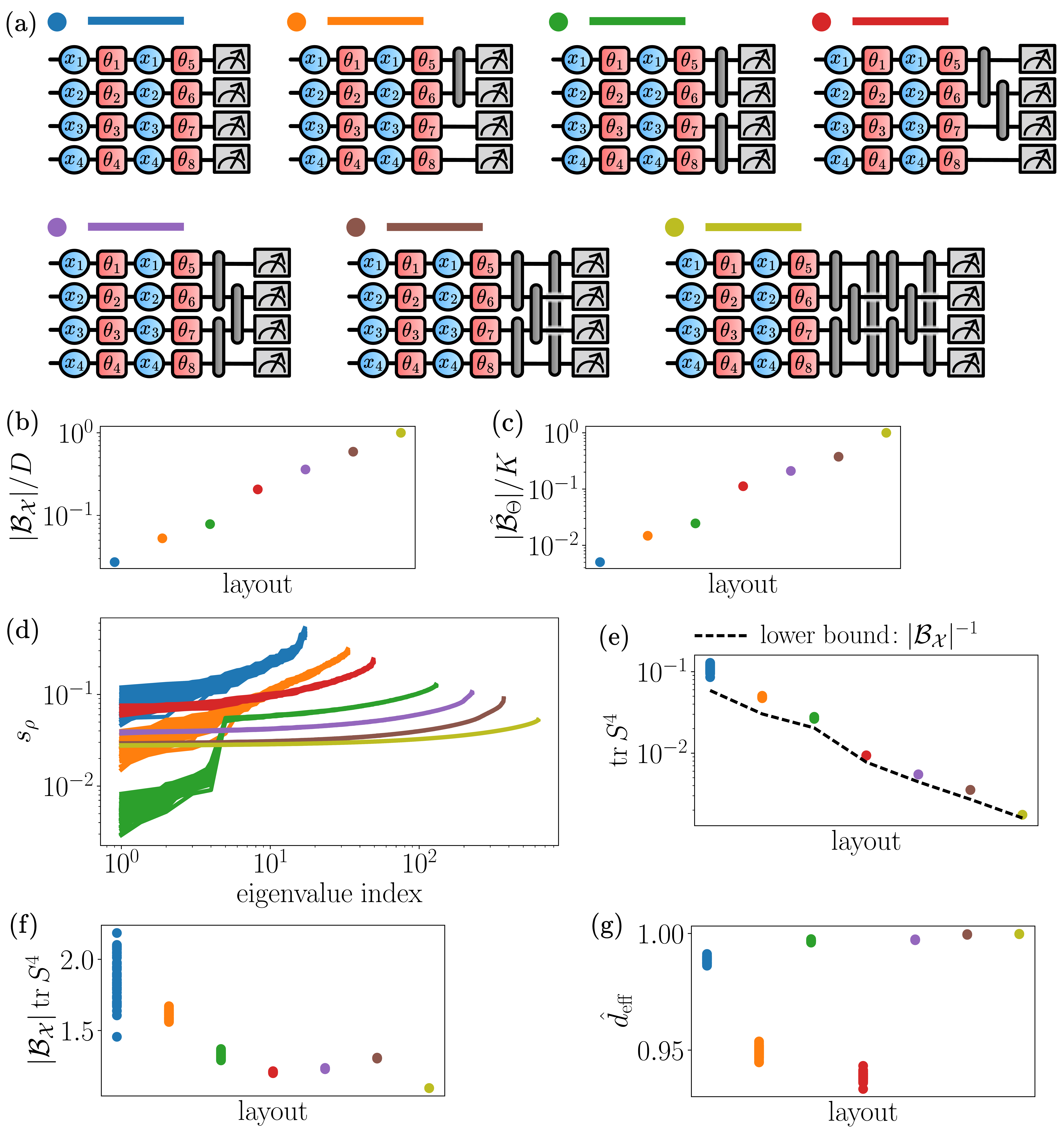}
    \caption{(a) QNN layouts used for the numerical analysis in this figure. Each color is related to a different structure of the two-qubits entangling gates in the measurement layer. (b) and (c) Scaling of $|\mathcal{B}_{\mathcal{X}}|$ and $|\tilde{\mathcal{B}}_{\Theta}|$, respectively, for the different layouts (with $\mathcal{B}_{\mathcal{X}}=\bigtimes_{n=1}^N\mathcal{B}_{\mathcal{X}_{q_n}}$ and $\tilde{\mathcal{B}}_{\Theta}=\bigtimes_{n=1}^N\tilde{\mathcal{B}}_{\Theta_{q_n}}$). Maximum values ($(2L+1)^N$ and $3^M$, with $L=2$, $N=4$ and $M=8$) are obtained for entangling layers with depth scaling with $N$. (d) Correlation spectra for the different layouts ($50$ independent model draws per layout are shown). (e)-(f) Purity of correlation spectra for the different layouts, with lower bound $1/|\mathcal{B}_{\mathcal{X}}|$ shown by the black dashed line in (e), and normalized by the lower bound in (f). $|\mathcal{B}_{\mathcal{X}}|\,\mathrm{tr}(S^4)$ does not strongly depend on the layout. (g) Normalized ED for the different layouts: also the normalized ED does not strongly depend on the layout, mirroring the weak dependence of $|\mathcal{B}_{\mathcal{X}}|\,\mathrm{tr}(S^4)$ on the layout.}
    \label{figapp:Figure_QNN_layouts} 
\end{figure*}

In Fig.~\ref{figapp:Figure_QNN_layouts} we show how the structure (depth) of the measurement layer in a simple QNN influences the number of basis functions, the correlation spectrum and the ED of the model. The different QNN layouts compared are shown in panel (a). For each model layout, for we consider random structure constants $\Gamma$ uniformly drawn in $[-1,+1]^{D\times K}$ with the elements corresponding to basis functions outside the allowed basis set $\mathcal{B}$ set to zero.
In panels (b) and (c) we show how the sizes of $\mathcal{B}_{\mathcal{X}}=\bigtimes_{n=1}^N\mathcal{B}_{\mathcal{X}_{q_n}}$ and $\tilde{\mathcal{B}}_{\Theta}=\bigtimes_{n=1}^N\tilde{\mathcal{B}}_{\Theta_{q_n}}$ scale for the different layouts, where we can observe that indeed the maximum size of the basis sets ($(2L+1)^N$ and $3^M$) is achieved when the depth is $\lfloor N/2\rfloor$. For random model realizations corresponding to the different layouts, the correlation spectrum (shown in (d)) has a purity $\mathrm{tr}(S^4)$ whose value is close to the lower bound set by $|\mathcal{B}_{\mathcal{X}}|^{-1}$ as shown in panels (e) and (f). This in turn results in the normalized ED not having a strong dependence on the QNN entangling layer structure. Therefore, a more in depth investigation should include information on the actual distribution of the values of the structure constants (here taken to be uniform), which depends on the details of the individual gates in the QNN circuit.

\subsection{Example: circuit-induced bias through entangling layers}
\label{app:qnn-structure-constants-example}
In the main text, biased Fourier regression models are constructed by choosing structure constants with prescribed alignment to a target data-generating function. Here we show that an analogous mechanism can arise in QNNs: the structure constants are not arbitrary, but are determined by the encoding gates, the variational gates, the entangling layers, and the measured observable. Consequently, changing these circuit ingredients can generate or remove specific input and parameter basis functions in the Fourier expansion, thereby biasing the model toward different classes of target functions. This showcases the existence of QNN design choices that can modify the form of the QNN output in a controlled way.

To exemplify those basic mechanisms, we consider a four-qubit example with initial state $\ket{0}$ and output observable $\hat{\sigma}^x_2$. Let
\begin{equation}
\hat{R}^x_{n}(x_n)=\exp\left(-\mathrm{i}\,\hat{\sigma}^x_n \,\frac{x_n}{2}\right)\,\,,
\qquad
\hat{R}^y_{n}(\theta_n)=\exp\left(-\mathrm{i}\,\hat{\sigma}^y_n \,\frac{\theta_n}{2}\right)\,\,,
\end{equation}
be the single-qubit rotation gates and
\begin{equation}
\hat{E}_{zz}=\prod_{n=1}^{3}\exp\left(-\mathrm{i}\,\hat{\sigma}^z_n \hat{\sigma}^z_{n+1}\frac{\pi}{4}\right)\,\,,
\qquad
\hat{E}_{yy}=\prod_{n=1}^{3}\exp\left(-\mathrm{i}\,\hat{\sigma}^y_n \hat{\sigma}^y_{n+1}\frac{\pi}{4}\right)\,\,,
\end{equation}
the non-parameterized entangling layers. We define the ansatzes
\begin{align}
&\hat{U}_{0}
=
\left[\prod_{n=1}^{4}\hat{R}^y_{n}(\theta_{n+4})\right]
\hat{E}_{zz}
\left[\prod_{n=1}^{4}\hat{R}^y_{n}(\theta_n)\right]
\left[\prod_{n=1}^{4}\hat{R}^x_{n}(x_n)\right]\,\,,
\\
&\hat{U}_{ZZ}
=
\hat{E}_{zz}\,\hat{U}_{0}\,\,,
\\
&\hat{U}_{YY+ZZ}
=
\hat{E}_{yy}\,U_{ZZ}\,\,.
\end{align}
The corresponding model outputs are
\begin{align}
&f^{0}_{\boldsymbol{\theta}}(\boldsymbol{x})=
\bra{0}\hat{U}_{0}^\dagger\,\hat{\sigma}^x_2\,\hat{U}_{0}\ket{0} \,\,,\\
&f^{ZZ}_{\boldsymbol{\theta}}(\boldsymbol{x})=
\bra{0}\hat{U}_{ZZ}^\dagger\,\hat{\sigma}^x_2\,\hat{U}_{ZZ}\ket{0} \,\,,\\
&f^{YY+ZZ}_{\boldsymbol{\theta}}(\boldsymbol{x})=
\bra{0}\hat{U}_{YY+ZZ}^\dagger\,\hat{\sigma}^x_2\,\hat{U}_{YY+ZZ}\ket{0} \,\,.
\end{align}
Equivalently, adding $\hat{E}_{zz}$ and $\hat{E}_{yy}$ can be viewed as leaving the original ansatz $\hat{U}_{0}$ unchanged but measuring the correlated observables
\begin{align}
&\hat{M}_{ZZ}=\hat{E}_{zz}^\dagger\,\hat{\sigma}^x_2\,\hat{E}_{zz}=-\hat{\sigma}^z_1\,\hat{\sigma}^x_2\,\hat{\sigma}^z_3\,\,,\\
&\hat{M}_{YY+ZZ}=\hat{E}_{zz}^\dagger\hat{E}_{yy}^\dagger\,\hat{\sigma}^x_2\,\hat{E}_{yy}\hat{E}_{zz}=-\hat{\sigma}^y_1\,\hat{\sigma}^x_2\,\hat{\sigma}^x_3\,\hat{\sigma}^z_4\,\,.
\end{align}
Thus, entangling layers placed before measurement and correlated measurement operators are two equivalent ways of modifying the accessible Fourier components.
We expand the output as
\begin{align}
&f_{\boldsymbol{\theta}}(\boldsymbol{x})
=
\sum_{\mu,\nu}
\Gamma_{\mu,\nu}
\prod_{j=1}^{4} b_{\mu}(x_j)
\prod_{k=1}^{8} b_{\nu}(\theta_k)\,\,,\\
&b_0(z)=1\,\,,\quad b_c(z)=\cos z\,\,,\quad b_s(z)=\sin z\,\,,
\end{align}
where $\mu\in\{0,c,s\}^4$ and $\nu\in\{0,c,s\}^8$. For compactness, the following tables list the nonzero coefficients $\widetilde\Gamma_{\mu,\nu}$ in this unnormalized trigonometric basis. In the orthonormal convention used in the rest of the manuscript, $\phi_0=1$, $\phi_c=\sqrt2\cos$, $\phi_s=\sqrt2\sin$, each entry is converted as
\begin{equation}
\Gamma_{\mu,\nu}
=
2^{-r(\mu,\nu)/2}\,
\widetilde\Gamma_{\mu,\nu}\,\,,
\end{equation}
where $r(\mu,\nu)$ is the total number of nonzero symbols, $c$ or $s$, appearing in the multi-indices $\mu$ and $\nu$.
For the $\hat{U}_0$ ansatz, the nonzero coefficients are
\begin{center}
\small
\begin{tabular}{c|c|c|c}
$\mu$ & $\nu$ & $\widetilde\Gamma_{\mu,\nu}$ & $r(\mu,\nu)$\\
\hline
$(0,c,0,0)$ & $(0,c,0,0,0,s,0,0)$ & $+1$ & $3$\\
$(c,c,c,0)$ & $(c,s,c,0,0,c,0,0)$ & $-1$ & $7$
\end{tabular}
\end{center}
All other entries vanish. For the $ZZ$ ansatz, the nonzero coefficients are
\begin{center}
\small
\begin{tabular}{c|c|c|c}
$\mu$ & $\nu$ & $\widetilde\Gamma_{\mu,\nu}$ & $r(\mu,\nu)$\\
\hline
$(0,c,0,0)$ & $(0,s,0,0,c,c,c,0)$ & $+1$ & $5$\\
$(0,s,s,c)$ & $(0,0,0,c,c,c,s,0)$ & $+1$ & $7$\\
$(c,0,c,c)$ & $(c,0,s,c,c,s,s,0)$ & $-1$ & $9$\\
$(c,c,c,0)$ & $(c,c,c,0,c,s,c,0)$ & $-1$ & $9$\\
$(c,c,s,c)$ & $(s,s,0,c,s,c,s,0)$ & $-1$ & $10$\\
$(c,s,0,0)$ & $(s,0,0,0,s,c,c,0)$ & $+1$ & $6$\\
$(s,0,c,0)$ & $(0,0,c,0,s,s,c,0)$ & $+1$ & $6$\\
$(s,c,c,c)$ & $(0,c,s,c,s,s,s,0)$ & $+1$ & $10$
\end{tabular}
\end{center}
After adding the $YY$ entangling layer, the nonzero coefficients become
\begin{center}
\small
\begin{tabular}{c|c|c|c}
$\mu$ & $\nu$ & $\widetilde\Gamma_{\mu,\nu}$ & $r(\mu,\nu)$\\
\hline
$(c,0,0,s)$ & $(s,0,0,0,0,s,s,s)$ & $+1$ & $6$\\
$(c,0,c,c)$ & $(s,0,c,c,0,s,s,c)$ & $-1$ & $9$\\
$(c,c,c,0)$ & $(s,c,s,0,0,s,c,c)$ & $+1$ & $9$\\
$(c,c,s,c)$ & $(s,c,0,s,0,s,c,s)$ & $+1$ & $10$\\
$(s,c,c,c)$ & $(0,s,s,s,0,c,c,s)$ & $-1$ & $10$\\
$(s,c,s,0)$ & $(0,s,0,0,0,c,c,c)$ & $+1$ & $7$\\
$(s,s,0,c)$ & $(0,0,0,c,0,c,s,c)$ & $+1$ & $7$\\
$(s,s,c,s)$ & $(0,0,c,0,0,c,s,s)$ & $-1$ & $8$
\end{tabular}
\end{center}
This example gives a practical demonstration of how entangling operations can add non-linear correlations among the input feature functions, and illustrates a possible way of generating model bias at the circuit level. 
In $\hat{U}_0$, one layer of entangling operations $\hat{E}_{zz}$ is not sufficient to correlate all four input features, hence no products of functions of all four $x_n$'s appear in the output input space (the index $\mu$ of the structure constants). Furthermore, when measuring only $\hat{\sigma}^x_2$ in $\hat{U}_0$, the BWLC cannot spread to the qubit encoding $x_4$, hence the output cannot depend in this feature.
Adding the entangling layer $\hat{E}_{zz}$ in $\hat{U}_{ZZ}$ enlarges the BWLC and provides an output that includes all four input features and has a higher degree of correlations (i.e., more products of input basis functions) among them. In this example, it is possible to see that the parameter $\theta_8$ has no effect on the output, which is again a consequence of the BWLC not including the rotation gate involving $\theta_8$. The dependence on $\theta_8$ appears in $\hat{U}_{YY+ZZ}$, where the added entangling layer $\hat{E}_{yy}$ enlarges the BWLC. (Here, the dependence on $\theta_5$ disappears since the measurement operator $\hat{M}_{YY+ZZ}$ commutes with $\hat{R}^y_1(\theta_5)$, hence $\theta_5$ has no effect on the output.)
The two circuits $\hat{U}_{ZZ}$ and $\hat{U}_{YY+ZZ}$ differ only by a fixed entangling layer, yet the support of the structure constants changes. For example, the $ZZ$ circuit contains the term
\begin{equation}
\cos x_2\,
\sin\theta_2\cos\theta_5\cos\theta_6\cos\theta_7\,\,,
\end{equation}
corresponding to
\begin{equation}
\widetilde\Gamma_{(0,c,0,0),(0,s,0,0,c,c,c,0)}=1\,\,,
\end{equation}
whereas this coefficient vanishes after adding $\hat{E}_{yy}$. Conversely, the $YY+ZZ$ circuit generates, for instance,
\begin{equation}
\cos x_1\sin x_4\,
\sin\theta_1\sin\theta_6\sin\theta_7\sin\theta_8\,\,,
\end{equation}
corresponding to
\begin{equation}
\widetilde\Gamma_{(c,0,0,s),(s,0,0,0,0,s,s,s)}=1\,\,,
\end{equation}
which is absent in the $ZZ$ circuit.
Therefore, practical QNN design choices impose structured zeros and nonzero patterns in $\Gamma$. These patterns determine which input basis functions can be reached and how their coefficients depend on trainable parameter functions. In this sense, encodings, entangling layers, and measurement operators naturally bias QNNs toward some target functions and away from others, rather than bias being only an artificial property imposed by manually setting structure constants.

\section{Definition of Fisher information matrix}
We provide here further details on the definition of the Fisher information matrix (FIM), and how to potentially extend our results to the case of models used for classification, thus going beyond regression models.

\subsection{FIM for regression models} \label{app:FIMregression}
We derive the formula used for calculating the FIM in the case of regression with mean squared error (MSE) loss function. For a statistical model $p(\boldsymbol{x},y;\boldsymbol{\theta})$ the elements of the FIM are defined as \cite{Amari1985,Amari1997,Pascanu2014}
\begin{equation}
F_{j,k}(\boldsymbol{\theta})=\mathbb{E}_{(\boldsymbol{x},y)\sim p}\bigg[\frac{\partial\,\mathrm{log}\,p(\boldsymbol{x},y;\boldsymbol{\theta})}{\partial\theta_j}\frac{\partial\,\mathrm{log}\,p(\boldsymbol{x},y;\boldsymbol{\theta})}{\partial\theta_k}\bigg] \,\,,
\end{equation}
which can be rewritten as 
\begin{equation}
F_{j,k}(\boldsymbol{\theta})=\mathbb{E}_{(\boldsymbol{x},y)\sim p}\bigg[\frac{\partial\,\mathrm{log}\,p_{\boldsymbol{\theta}}(y|\boldsymbol{x})}{\partial\theta_j}\frac{\partial\,\mathrm{log}\,p_{\boldsymbol{\theta}}(y|\boldsymbol{x})}{\partial\theta_k}\bigg] \,\,,
\end{equation}
by noting that $p(\boldsymbol{x},y;\boldsymbol{\theta})=p(\boldsymbol{x})p_{\boldsymbol{\theta}}(y|\boldsymbol{x})$, with $p_{\boldsymbol{\theta}}(y|\boldsymbol{x})$ being the model output probability conditioned on the input $\boldsymbol{x}$. 
The quantity $-\log p_{\boldsymbol{\theta}}(y|\boldsymbol{x})\equiv\ell(y,\boldsymbol{x};\boldsymbol{\theta})$ corresponds to the negative log-likelihood, typically used as loss function for training statistical models. 
A statistical model corresponding to the MSE loss function can be constructed by setting $p_{\boldsymbol{\theta}}(y|\boldsymbol{x})=\mathcal{N}_{f_{\boldsymbol{\theta}}(\boldsymbol{x}),\sigma^2}(y)$ with
\begin{equation}
    \mathcal{N}_{f_{\boldsymbol{\theta}}(\boldsymbol{x}),\sigma^2}(y)=\frac{1}{\sigma\sqrt{2\pi}}\mathrm{exp}\bigg[-\frac{(y-f_{\boldsymbol{\theta}}(\boldsymbol{x}))^2}{2\sigma^2}\bigg]
\end{equation}
for a fictitious $\sigma$ \cite{Pennington2018,Amari2019,Karakida2020,Hayase2021,Karakida2021}, with $f_{\boldsymbol{\theta}}(\boldsymbol{x})$ being the deterministic regression value output by our model. 
Using $\partial_{\theta_j}p_{\boldsymbol{\theta}}(y|\boldsymbol{x})=\sigma^{-2}(f_{\boldsymbol{\theta}}(\boldsymbol{x})-y)\,\partial_{\theta_j}f_{\boldsymbol{\theta}}(\boldsymbol{x})$, we can rewrite the FIM as
\begin{equation}
F_{j,k}(\boldsymbol{\theta})=\mathbb{E}_{\boldsymbol{x}}\bigg[\int\mathcal{N}_{f_{\boldsymbol{\theta}}(\boldsymbol{x}),\sigma^2}(y)\,(f_{\boldsymbol{\theta}}(\boldsymbol{x})-y)^2\,\sigma^{-4}\,\frac{\partial f_{\boldsymbol{\theta}}(\boldsymbol{x})}{\partial\theta_j}\frac{\partial f_{\boldsymbol{\theta}}(\boldsymbol{x})}{\partial\theta_k}\,\mathrm{d}y\bigg] \,\,.
\end{equation}
Performing the Gaussian integration over $y$, we finally arrive at
\begin{equation}
    F_{j,k}(\boldsymbol{\theta})=\sigma^{-2}\,\mathbb{E}_{\boldsymbol{x}}\bigg[\frac{\partial f_{\boldsymbol{\theta}}(\boldsymbol{x})}{\partial\theta_j}\frac{\partial f_{\boldsymbol{\theta}}(\boldsymbol{x})}{\partial\theta_k}\bigg] \,\,,
\end{equation}
which corresponds to the definition used when setting $\sigma=1$. This definition makes it clear that the FIM can be interpreted as a metric tensor that determines the response of the output of a model to a local change in the parameters, averaged over the input space, as
\begin{equation}
    \begin{split}
        \mathbb{E}_{\boldsymbol{x}}\big[(f_{\boldsymbol{\theta}+\mathrm{d}\boldsymbol{\theta}}(\boldsymbol{x})-f_{\boldsymbol{\theta}}(\boldsymbol{x}))^2\big]&=\mathbb{E}_{\boldsymbol{x}}\bigg[\bigg(\sum_{j=1}^M \frac{\partial f_{\boldsymbol{\theta}}(\boldsymbol{x})}{\partial\theta_j}\,\mathrm{d}\theta_j\bigg)^2\,\bigg]+\mathcal{O}(\Vert\mathrm{d}\boldsymbol{\theta}\Vert^2)\\
        &=\sum_{j,k} \mathbb{E}_{\boldsymbol{x}}\bigg[\frac{\partial f_{\boldsymbol{\theta}}(\boldsymbol{x})}{\partial\theta_j}\frac{\partial f_{\boldsymbol{\theta}}(\boldsymbol{x})}{\partial\theta_k}\bigg] \mathrm{d}\theta_j\mathrm{d}\theta_k+\mathcal{O}(\Vert\mathrm{d}\boldsymbol{\theta}\Vert^2)\\
        &\approx \mathrm{d}\boldsymbol{\theta}^{\top} F(\boldsymbol{\theta})\,\mathrm{d}\boldsymbol{\theta} \,\,.
    \end{split}
\end{equation}

\subsection{From regression to probabilistic models} \label{app:FIM_regr_to_prob}
Our FIM analysis and arguments could be extended to the case of probabilistic models by making the following observation. Let us for simplicity consider the case of classification of discrete labels. For a given input $\boldsymbol{x}$ and parameters $\boldsymbol{\theta}$, the model outputs the probability $p_{\ell}(\boldsymbol{x};\boldsymbol{\theta})$ for a given class $\ell$, with $\sum_{\ell}p_{\ell}(\boldsymbol{x};\boldsymbol{\theta})=1$. The FIM elements in this case are
\begin{equation}
\begin{split}
F_{j,k}(\boldsymbol{\theta})&=\mathbb{E}_{\boldsymbol{x},\ell}\bigg[\frac{\partial\,\mathrm{log}\,p_{\ell}(\boldsymbol{x};\boldsymbol{\theta})}{\partial\theta_j}\frac{\partial\,\mathrm{log}\,p_{\ell}(\boldsymbol{x};\boldsymbol{\theta})}{\partial\theta_k}\bigg] \\
&=\mathbb{E}_{\boldsymbol{x}}\bigg[\sum_{\ell}p_{\ell}(\boldsymbol{x};\boldsymbol{\theta})\,\frac{\partial\,\mathrm{log}\,p_{\ell}(\boldsymbol{x};\boldsymbol{\theta})}{\partial\theta_j}\frac{\partial\,\mathrm{log}\,p_{\ell}(\boldsymbol{x};\boldsymbol{\theta})}{\partial\theta_k}\bigg] \\
&=\mathbb{E}_{\boldsymbol{x}}\bigg[\sum_{\ell}\frac{1}{p_{\ell}(\boldsymbol{x};\boldsymbol{\theta})}\,\frac{\partial\,p_{\ell}(\boldsymbol{x};\boldsymbol{\theta})}{\partial\theta_j}\frac{\partial\,p_{\ell}(\boldsymbol{x};\boldsymbol{\theta})}{\partial\theta_k}\bigg] \\
&=4\,\mathbb{E}_{\boldsymbol{x}}\bigg[\sum_{\ell}\frac{\partial\sqrt{p_{\ell}(\boldsymbol{x};\boldsymbol{\theta})}}{\partial\theta_j}\frac{\partial\sqrt{p_{\ell}(\boldsymbol{x};\boldsymbol{\theta})}}{\partial\theta_k}\bigg] \\
&\equiv4\,\mathbb{E}_{\boldsymbol{x}}\bigg[\sum_{\ell}\frac{\partial\,q_{\ell}(\boldsymbol{x};\boldsymbol{\theta})}{\partial\theta_j}\frac{\partial\,q_{\ell}(\boldsymbol{x};\boldsymbol{\theta})}{\partial\theta_k}\bigg] \,\,,
\end{split}
\end{equation}
where we set $q_{\ell}(\boldsymbol{x};\boldsymbol{\theta})\equiv\sqrt{p_{\ell}(\boldsymbol{x};\boldsymbol{\theta})}$. The above expression has now the same form of that for regression models, for which the analytical arguments presented in our work apply.

\section{Analysis of FIM for regression models}
In this section we establish the connection between the FIM and the properties of the structure constants $\Gamma$, with particular focus on how the correlation spectrum $S$ influences the spectral properties of the FIM.

\subsection{Derivation of diagrammatic FIM expression} \label{app:FIM_diagramm_expr_deriv}
Here we derive the analytical and diagrammatic expression of the FIM elements in terms of the structure constants $\Gamma$ and the related correlation spectrum and singular vectors. We recall the definition of the (local) derivative tensor $\beta^{(j)}$, introduced as
\begin{equation}
    \frac{\partial\iota_{\nu}(\boldsymbol{\theta})}{\partial\theta_j}=\frac{\partial\iota^{(j)}_{\nu_j}(\theta_j)}{\partial\theta_j}\prod_{m\neq j}\iota^{(m)}_{\nu_m}(\theta_m)=\bigg(\sum_{\kappa_j}\beta^{(j)}_{\nu_j,\kappa_j}\,\iota^{(j)}_{\kappa_j}(\theta_j)\bigg)\prod_{m\neq j}\iota^{(m)}_{\nu_m}(\theta_m) \,\,.
\end{equation}
For example, if
\begin{equation}
    \iota^{(j)}_{\nu_j}(\theta_j)=\{1,\,\sqrt{2}\cos(\theta_j),...,\,\sqrt{2}\cos(L\theta_j),\,\sqrt{2}\sin(\theta_j),...,\,\sqrt{2}\sin(L\theta_j)\} \,\,,
\end{equation}
for some integer $L$, the derivative tensor $\beta^{(j)}$ is a $(2L+1)\times(2L+1)$ matrix of the form
\begin{equation}
    \beta^{(j)}=
    \begin{pmatrix}
        \begin{matrix}
            0 
        \end{matrix}
        & \rvline & 
        \begin{matrix}
            0 & 0 & \cdots & 0
        \end{matrix}
        & \rvline & 
        \begin{matrix}
            0 & 0 & \cdots & 0
        \end{matrix} \\
        \hline
        \begin{matrix}
            0 \\ 0 \\ \vdots \\ 0
        \end{matrix} & \rvline & \bigzero & \rvline & 
        \begin{matrix}
            1 & & & \\
             & 2 & & \\
             & & \ddots & \\
             & & & L 
        \end{matrix} \\
        \hline
        \begin{matrix}
            0 \\ 0 \\ \vdots \\ 0
        \end{matrix} & \rvline & \begin{matrix}
            -1 & & & \\
             & -2 & & \\
             & & \ddots & \\
             & & & -L 
        \end{matrix} & \rvline & \bigzero
    \end{pmatrix} \,\,,
\end{equation}
The derivative of the model output $f_{\boldsymbol{\theta}}(\boldsymbol{x})$ takes then the form
\begin{equation}
    \begin{split}
         \frac{\partial f_{\boldsymbol{\theta}}(\boldsymbol{x})}{\partial\theta_j}&=\sum_{\mu=1}^D e_{\mu}(\boldsymbol{x})\sum_{\nu=1}^K\Gamma_{\mu,\nu}\,\frac{\partial\iota_{\nu}(\boldsymbol{\theta})}{\partial\theta_j}\\
         &=\sum_{\mu=1}^D e_{\mu}(\boldsymbol{x})\sum_{\nu=(\nu_1...\nu_M)}\Gamma_{\mu,(\nu_1...\nu_M)}\,\sum_{\kappa_j}\beta^{(j)}_{\nu_j,\kappa_j}\,\iota^{(j)}_{\kappa_j}(\theta_j)\prod_{m\neq j}\iota^{(m)}_{\nu_m}(\theta_m)\\
         &=\sum_{\mu=1}^D e_{\mu}(\boldsymbol{x})\sum_{\kappa_j}\sum_{\nu=(\nu_1...\nu_M)}\beta^{(j)}_{\kappa_j,\nu_j}\,\Gamma_{\mu,(\nu_1...\kappa_j...\nu_M)}\,\iota_{\nu}(\boldsymbol{\theta})\\
         &=\sum_{\mu=1}^D e_{\mu}(\boldsymbol{x}) \sum_{\nu=(\nu_1...\nu_M)} \iota_{\nu}(\boldsymbol{\theta}) \sum_{\rho=1}^D U_{\mu,\rho}\,s_{\rho} \sum_{\kappa_j}\beta^{(j)}_{\kappa_j,\nu_j}\,[V^{\top}]_{\rho,(\nu_1...\kappa_j...\nu_M)}\\
         &\equiv\sum_{\mu=1}^D\sum_{\nu=1}^K\Lambda^{(j)}_{\mu,\nu}\,e_{\mu}(\boldsymbol{x})\,\iota_{\nu}(\boldsymbol{\theta}) \,\,,
    \end{split}
\end{equation}
with $\Lambda^{(j)}_{\mu,\nu}\equiv\sum_{\rho=1}^D U_{\mu,\rho}\,s_{\rho} \sum_{\kappa_j}\beta^{(j)}_{\kappa_j,\nu_j}\,[V^{\top}]_{\rho,(\nu_1...\kappa_j...\nu_M)}$. The element $F_{j,k}$ of the FIM then reads as
\begin{equation}
    \begin{split}
         F_{j,k}(\boldsymbol{\theta})&=\mathbb{E}_{\boldsymbol{x}}\bigg[\frac{\partial f_{\boldsymbol{\theta}}(\boldsymbol{x})}{\partial\theta_j}\frac{\partial f_{\boldsymbol{\theta}}(\boldsymbol{x})}{\partial\theta_k}\bigg]\\
         &=\sum_{\mu,\mu'}\sum_{\nu,\nu'}\Lambda^{(j)}_{\mu,\nu}\,\Lambda^{(k)}_{\mu',\nu'}\,\iota_{\nu}(\boldsymbol{\theta})\,\iota_{\nu'}(\boldsymbol{\theta})\,\mathbb{E}_{\boldsymbol{x}}\big[e_{\mu}(\boldsymbol{x})\,e_{\mu'}(\boldsymbol{x})\big]\\
         &=\sum_{\mu}\sum_{\nu,\nu'}\Lambda^{(j)}_{\mu,\nu}\,\Lambda^{(k)}_{\mu,\nu'}\,\iota_{\nu}(\boldsymbol{\theta})\,\iota_{\nu'}(\boldsymbol{\theta})\\
         &=\sum_{\nu,\nu'}\iota_{\nu}(\boldsymbol{\theta})\,\iota_{\nu'}(\boldsymbol{\theta})\sum_{\rho,\rho'}s_{\rho}\,s_{\rho'}\sum_{\mu}U_{\mu,\rho}\,U_{\mu,\rho'}\times\\
         &\quad\sum_{\kappa_j,\kappa_k'}\beta^{(j)}_{\kappa_j,\nu_j}[V^{\top}]_{\rho,(\nu_1...\kappa_j...\nu_M)}\beta^{(k)}_{\kappa_k',\nu_k'}[V^{\top}]_{\rho',(\nu_1'...\kappa_k'...\nu_M')}\\
         &=\sum_{\nu,\nu'}\iota_{\nu}(\boldsymbol{\theta})\,\iota_{\nu'}(\boldsymbol{\theta})\sum_{\rho}s_{\rho}^2\sum_{\kappa_j,\kappa_k'}\beta^{(j)}_{\kappa_j,\nu_j}[V^{\top}]_{\rho,(\nu_1...\kappa_j...\nu_M)}\times\\
         &\quad\beta^{(k)}_{\kappa_k',\nu_k'}[V^{\top}]_{\rho,(\nu_1'...\kappa_k'...\nu_M')}\\
         &\equiv\sum_{\nu,\nu'}\mathcal{F}^{(j,k)}_{\nu,\nu'}\,\iota_{\nu}(\boldsymbol{\theta})\,\iota_{\nu'}(\boldsymbol{\theta}) \\
         &\equiv \big(\boldsymbol{\iota}(\boldsymbol{\theta})\big|\mathcal{F}^{(j,k)}\big|\boldsymbol{\iota}(\boldsymbol{\theta})\big) \,\,,
         \label{eqapp:FIM_analytic_expr}
    \end{split}
\end{equation}
where in the second line we used the fact that $\mathbb{E}_{\boldsymbol{x}}\big[e_{\mu}(\boldsymbol{x})\,e_{\mu'}(\boldsymbol{x})\big]=\delta_{\mu,\mu'}$, in the fourth line $\sum_{\mu}U_{\mu,\rho}\,U_{\mu,\rho'}=\delta_{\rho,\rho'}$, and in the last line $\big|\boldsymbol{\iota}(\boldsymbol{\theta})\big)$ the $K$-dimensional vector with components $\iota_{\nu}(\boldsymbol{\theta})$.
In the above equation, we define the $\mathcal{F}^{(j,k)}$ tensor as
\begin{equation}
    \begin{split}
    \mathcal{F}^{(j,k)}_{\nu,\nu'}&=\sum_{\rho}s_{\rho}^2\sum_{\kappa_j,\kappa_k'}\beta^{(j)}_{\kappa_j,\nu_j}[V^{\top}]_{\rho,(\nu_1...\kappa_j...\nu_M)}\beta^{(k)}_{\kappa_k',\nu_k'}[V^{\top}]_{\rho,(\nu_1'...\kappa_k'...\nu_M')}\\
    &\equiv \big[(V^{\top}B_j)^{\top}S^2\,(V^{\top}B_k)\big]_{\nu,\nu'}\,\,.
    \end{split}
    \label{eqapp:F_tensor_analytic_expr}
\end{equation}
where $B_j=I^{(1)}_{\tilde{d}}\otimes I^{(2)}_{\tilde{d}}\otimes\,... \otimes\,\beta^{(j)}\,\otimes\,...\otimes I^{(M)}_{\tilde{d}}$ (with $I^{(k)}_{\tilde{d}}$ being the $\tilde{d}$-dimensional identity matrix acting on the function space `local' to the $k$-th parameter).

The $\mathcal{F}^{(j,k)}$ tensor introduced above and the FIM elements can be naturally represented as tensor networks. This is useful for evaluating the FIM for tensorized models, for numerically studying models with larger number of parameters and input features. The $\Lambda^{(j)}_{\mu,\nu}$ tensor 
\begin{align}
    \Lambda^{(j)}_{\mu,\nu}=\;
    \begin{tikzpicture} [baseline={([yshift=+0ex]current bounding box.center)}]
        \node[draw, shape=rectangle] (U) at (1,-1) {$U$};
        \node[draw, shape=diamond] (S) at (2,-1) {$S$};
        \node[rectangle,minimum width=8em,draw] (V) at (4.2,-1) {$V$};
        \node[draw, shape=rectangle] (bj) at (4.2,0) {$\beta^{(j)}$};
        \node[] (mu) at (0,-1) {$\mu$};
        \node[] (nu1) at (3,1) {$\nu_1$};
        \node[] (nuj) at (4.2,1) {$\nu_j$};
        \node[] (nuM) at (5.4,1) {$\nu_M$};
        \draw [thick] (mu) -- (U) -- (S) -- (V);
        \draw [thick] (nu1) -- (nu1 |- V.north);
        \draw [thick] (nuj) -- (nuj |- bj.north)
                      (bj) -- (bj |- V.north);
        \draw [thick] (nuM) -- (nuM |- V.north);
    \end{tikzpicture}
\end{align}
The $\mathcal{F}^{(j,k)}_{\nu,\nu'}$ tensor 
\begin{equation}
    \mathcal{F}^{(j,k)}_{\nu,\nu'}=\;
    \begin{tikzpicture} [baseline={([yshift=+0ex]current bounding box.center)}]
        \node[] (nu1) at (0,2) {$\nu_1$};
        \node[] (nuj) at (0,1) {$\nu_j$};
        \node[] (nuM) at (0,0) {$\nu_M$};
        \node[draw, shape=rectangle] (bj) at (1,1) {$\beta^{(j)}$};
        \node[rectangle, minimum height=8em, draw] (V1) at (2,1) {$V$};
        \node[draw, shape=diamond] (S) at (3,-1) {$S^2$};
        \node[rectangle, minimum height=8em, draw] (V2) at (4,1) {$V$};
        \node[draw, shape=rectangle] (bk) at (5,1) {$\beta^{(k)}$};
        \node[] (nup1) at (6,2) {$\nu_1'$};
        \node[] (nupk) at (6,1) {$\nu_k'$};
        \node[] (nupM) at (6,0) {$\nu_M'$};
        \draw [thick] (nu1) -- (nu1 -| V1.west);
        \draw [thick] (nuj) -- (bj) -- (V1);
        \draw [thick] (nuM) -- (nuM -| V1.west);
        \draw [thick] (nup1) -- (nup1 -| V2.east);
        \draw [thick] (V2) -- (bk) -- (nupk);
        \draw [thick] (nupM) -- (nupM -| V2.east);
        \draw [thick] (V1) -- (V1 |- S.west);
        \draw [thick] (S) -- (S -| V1.south);
        \draw [thick] (V2) -- (V2 |- S.east);
        \draw [thick] (S) -- (S -| V2.south);
    \end{tikzpicture}
\end{equation}
which then leads to the tensor network expression of the FIM elements
\begin{equation}
    F_{j,k}(\boldsymbol{\theta})=\;
    \begin{tikzpicture} [baseline={([yshift=+0ex]current bounding box.center)}]
        \node[rectangle, draw] (i1) at (0,2) {$\boldsymbol{\iota}^{(1)}(\theta_1)$};
        \node[rectangle, draw] (ij) at (0,1) {$\boldsymbol{\iota}^{(j)}(\theta_j)$};
        \node[rectangle, draw] (iM) at (0,0) {$\boldsymbol{\iota}^{(M)}(\theta_M)$};
        \node[draw, shape=rectangle] (bj) at (1.5,1) {$\beta^{(j)}$};
        \node[rectangle, minimum height=8em, draw] (V1) at (2.5,1) {$V$};
        \node[draw, shape=diamond] (S) at (3.5,-1) {$S^2$};
        \node[rectangle, minimum height=8em, draw] (V2) at (4.5,1) {$V$};
        \node[draw, shape=rectangle] (bk) at (5.5,1) {$\beta^{(k)}$};
        \node[rectangle, draw] (ip1) at (7,2) {$\boldsymbol{\iota}^{(1)}(\theta_1)$};
        \node[rectangle, draw] (ipk) at (7,1) {$\boldsymbol{\iota}^{(k)}(\theta_k)$};
        \node[rectangle, draw] (ipM) at (7,0) {$\boldsymbol{\iota}^{(M)}(\theta_M)$};
        \draw [thick] (i1) -- (i1 -| V1.west);
        \draw [thick] (ij) -- (bj) -- (V1);
        \draw [thick] (iM) -- (iM -| V1.west);
        \draw [thick] (ip1) -- (ip1 -| V2.east);
        \draw [thick] (V2) -- (bk) -- (ipk);
        \draw [thick] (ipM) -- (ipM -| V2.east);
        \draw [thick] (V1) -- (V1 |- S.west);
        \draw [thick] (S) -- (S -| V1.south);
        \draw [thick] (V2) -- (V2 |- S.east);
        \draw [thick] (S) -- (S -| V2.south);
    \end{tikzpicture}
\end{equation}
where $\boldsymbol{\iota}^{(j)}(\theta_j)$ is a rank-1 tensor (vector) whose $\tilde{d}$ components are the local basis functions evaluated at $\theta_j$, i.e., $\boldsymbol{\iota}^{(j)}(\theta_j)=\big(\iota^{(j)}_1(\theta_j),...,\iota^{(j)}_{\tilde{d}}(\theta_j)\big)$.

\subsection{Correlation bounds on FIM and effective dimension} \label{app:ED_bound}
We show here that the effective dimension (ED) is upper-bounded by the rank of the correlation spectrum $D$ (in case $D<M$). To do this, it is sufficient to rewrite Eqs.~\eqref{eqapp:FIM_analytic_expr} and \eqref{eqapp:F_tensor_analytic_expr} as
\begin{equation}
\begin{split}
    F_{j,k}(\boldsymbol{\theta})&=\sum_{\rho=1}^Ds_{\rho}^2\,\big(\boldsymbol{\iota}(\boldsymbol{\theta})\big|B_j^{\top}P_{\rho}\,B_k\big|\boldsymbol{\iota}(\boldsymbol{\theta})\big)=\sum_{\rho=1}^Ds_{\rho}^2\,\big(\boldsymbol{\iota}(\boldsymbol{\theta})\big|B_j^{\top}P_{\rho}^{\top}P_{\rho}\,B_k\big|\boldsymbol{\iota}(\boldsymbol{\theta})\big)\\
    &=\sum_{\rho=1}^Ds_{\rho}^2\,\alpha_j^{\rho}(\boldsymbol{\theta})\,\alpha_k^{\rho}(\boldsymbol{\theta})\,\big(\boldsymbol{\mathrm{v}}_{\rho}\big|\,\boldsymbol{\mathrm{v}}_{\rho}\big)=\sum_{\rho=1}^Ds_{\rho}^2\,\alpha_j^{\rho}(\boldsymbol{\theta})\,\alpha_k^{\rho}(\boldsymbol{\theta})\,\,,
\end{split}
\end{equation}
with $P_{\rho}$ being the projection on the subspace spanned by the vector $V_{\cdot,\rho}$, which satisfies $P_{\rho}^{\top}P_{\rho}=P_{\rho}$ used in the first line, and $P_{\rho}\,B_k\big|\boldsymbol{\iota}(\boldsymbol{\theta})\big)=\alpha_k^{\rho}(\boldsymbol{\theta})\big|\boldsymbol{\mathrm{v}}_{\rho}\big)$ with $\big|\boldsymbol{\mathrm{v}}_{\rho}\big)=V_{\cdot,\rho}$. This allows us to write
\begin{equation}
     F(\boldsymbol{\theta})=\sum_{\rho=1}^Ds_{\rho}^2\,P_{\boldsymbol{\alpha}^{\rho}(\boldsymbol{\theta})} \,\,,
\end{equation}
with $P_{\boldsymbol{\alpha}^{\rho}(\boldsymbol{\theta})}=\boldsymbol{\alpha}^{\rho}(\boldsymbol{\theta})\,\boldsymbol{\alpha}^{\rho}(\boldsymbol{\theta})^{\top}$, which is proportional to the projection onto the $M$-components vector $\boldsymbol{\alpha}^{\rho}(\boldsymbol{\theta})=\big(\alpha_1^{\rho}(\boldsymbol{\theta}),...,\alpha_M^{\rho}(\boldsymbol{\theta})\big)^{\top}$. Since $F(\boldsymbol{\theta})$ is a sum of at most $D$ linearly independent projections, we conclude that $\mathrm{rank}\,F(\boldsymbol{\theta})\leq D$. By the results of \cite{Abbas2021}, the ED is upper-bounded by the maximal rank of $F(\boldsymbol{\theta})$, and hence by $D$ in the regime $D<M$.

\subsection{Random matrix theory analysis of the FIM} \label{app:RMT_FIM_derivs}
We derive here the expected value and variance of the FIM elements over random realizations of the orthogonal matrix $V\in \mathrm{O}(K)$, with $\mathrm{O}(K)$ the group of $K\times K$ orthogonal matrices, using results from random matrix theory \cite{Brouwer1996,Collins2006,Collins2009}. We recall the expression of the FIM elements derived in Section \ref{app:FIM_diagramm_expr_deriv}
\begin{equation}
    F_{j,k}(\boldsymbol{\theta})=\sum_{\nu,\nu'}\mathcal{F}^{(j,k)}_{\nu,\nu'}\,\iota_{\nu}(\boldsymbol{\theta})\,\iota_{\nu'}(\boldsymbol{\theta})\equiv\big(\boldsymbol{\iota}(\boldsymbol{\theta})\big|\mathcal{F}^{(j,k)}\big|\boldsymbol{\iota}(\boldsymbol{\theta})\big) \,\,,
\end{equation}
with $\big|\boldsymbol{\iota}(\boldsymbol{\theta})\big)$ the $K$-dimensional vector with components $\iota_{\nu}(\boldsymbol{\theta})$ and
\begin{equation}
    \begin{split}
        \mathcal{F}^{(j,k)}_{\nu,\nu'}&=\sum_{\rho}s_{\rho}^2\sum_{\kappa_j,\kappa_k'}\beta^{(j)}_{\kappa_j,\nu_j}[V^{\top}]_{\rho,(\nu_1...\kappa_j...\nu_M)}\beta^{(k)}_{\kappa_k',\nu_k'}[V^{\top}]_{\rho,(\nu_1'...\kappa_k'...\nu_M')}\\
        &\equiv\big[(V^{\top}B_j)^{\top}S^2\,(V^{\top}B_k)\big]_{\nu,\nu'} \,\,.
    \end{split}
\end{equation}
We start by computing the expectation value of $F_{j,k}(\boldsymbol{\theta})$ over $\mathrm{O}(K)$.
\begin{equation}
    \begin{split}
        \mathbb{E}_{V\in \mathrm{O}(K)}\big[F_{j,k}(\boldsymbol{\theta})\big]&=\mathbb{E}_{V\in \mathrm{O}(K)}\big[\big(\boldsymbol{\iota}(\boldsymbol{\theta})\big|(V^{\top}B_j)^{\top}S^2\,(V^{\top}B_k)\big|\boldsymbol{\iota}(\boldsymbol{\theta})\big)\big]\\
        &=\big(\boldsymbol{\iota}(\boldsymbol{\theta})\big|B_j^{\top}\,\mathbb{E}_{V\in \mathrm{O}(K)}\big[V\,S^2\,V^{\top}\big]B_k\big|\boldsymbol{\iota}(\boldsymbol{\theta})\big)\\
        &=\sum_{\nu,\nu'}\iota_{\nu}(\boldsymbol{\theta})\,\iota_{\nu'}(\boldsymbol{\theta})\sum_{\kappa,\kappa'}[B_j^{\top}]_{\nu,\kappa}
        [B_k]_{\kappa',\nu'}\sum_{\rho}s_{\rho}^2\,\mathbb{E}_{V\in \mathrm{O}(K)}\big[V_{\kappa,\rho}V_{\kappa',\rho}\big]\\
        &=\frac{1}{K}\sum_{\nu,\nu'}\iota_{\nu}(\boldsymbol{\theta})\,\iota_{\nu'}(\boldsymbol{\theta})\sum_{\kappa}[B_j^{\top}]_{\nu,\kappa}
        [B_k]_{\kappa,\nu'}\sum_{\rho}s_{\rho}^2\\
        &=\frac{\mathrm{tr}(S^2)}{K}\,\big(\boldsymbol{\iota}(\boldsymbol{\theta})\big|B_j^{\top}B_k\big|\boldsymbol{\iota}(\boldsymbol{\theta})\big) \,\,,
    \end{split}
\end{equation}
where in the third line we used \cite{Brouwer1996,Collins2006,Collins2009}
\begin{equation}
    \mathbb{E}_{V\in \mathrm{O}(K)}\big[V_{\kappa,\rho}V_{\kappa',\rho}\big]=\frac{\delta_{\kappa,\kappa'}}{K} \,\,.
\end{equation}
In order to calculate the variance $\mathrm{Var}_{V\in \mathrm{O}(K)}\big[F_{j,k}(\boldsymbol{\theta})\big]$ we need the following identity \cite{Collins2006,Collins2009}
\begin{equation}
    \begin{split}
        \mathbb{E}_{V\in \mathrm{O}(K)}&\big[V_{\alpha,\rho}V_{\beta,\rho}V_{\alpha',\rho'}V_{\beta',\rho'}\big]=\\
        &=\frac{K+1}{D^3_K}\,\delta_{\alpha,\beta}\,\delta_{\alpha',\beta'}-\frac{1}{D^3_K}\,\Big(\delta_{\alpha,\alpha'}\,\delta_{\beta,\beta'}+\delta_{\alpha,\beta'}\,\delta_{\alpha',\beta}\Big)+\\
        &\quad\;\,\delta_{\rho,\rho'}\bigg[\frac{K+1}{D^3_K}\,\delta_{\alpha,\alpha'}\,\delta_{\beta,\beta'}-\frac{1}{D^3_K}\,\Big(\delta_{\alpha,\beta}\,\delta_{\alpha',\beta'}+\delta_{\alpha,\beta'}\,\delta_{\alpha',\beta}\Big)+\\
        &\quad\quad\quad\;\;\,\frac{K+1}{D^3_K}\,\delta_{\alpha,\beta'}\,\delta_{\alpha',\beta}-\frac{1}{D^3_K}\,\Big(\delta_{\alpha,\beta}\,\delta_{\alpha',\beta'}+\delta_{\alpha,\alpha'}\,\delta_{\beta,\beta'}\Big)\bigg]\\
        &=\frac{\delta_{\alpha,\beta}\,\delta_{\alpha',\beta'}}{K^2}+\frac{\delta_{\rho,\rho'}}{K^2}\Big(\delta_{\alpha,\alpha'}\,\delta_{\beta,\beta'}+\delta_{\alpha,\beta'}\,\delta_{\alpha',\beta}\Big)+\mathcal{O}(K^{-3}) \,\,,
    \end{split}
\end{equation}
with $D^3_K\equiv K(K-1)(K+2)$. The variance $\mathrm{Var}_{V\in \mathrm{O}(K)}\big[F_{j,k}(\boldsymbol{\theta})\big]$ is then
\begin{equation}
    \begin{split}
        \mathrm{Var}_{V\in \mathrm{O}(K)}\big[F_{j,k}(\boldsymbol{\theta})\big]&=\mathbb{E}_{V\in \mathrm{O}(K)}\big[F_{j,k}(\boldsymbol{\theta})^2\big]-\mathbb{E}_{V\in \mathrm{O}(K)}\big[F_{j,k}(\boldsymbol{\theta})\big]^2\\
        &=\sum_{\nu,n}\iota_{\nu}(\boldsymbol{\theta})\,\iota_{n}(\boldsymbol{\theta})\sum_{\nu',n'}\iota_{\nu'}(\boldsymbol{\theta})\,\iota_{n'}(\boldsymbol{\theta})\sum_{\rho,\rho'}s^2_{\rho}\,s^2_{\rho'}\,\times\\
        &\quad\;\,\sum_{\alpha,\beta,\alpha',\beta'}[B_j^{\top}]_{\nu,\alpha}
        [B_k]_{\beta,n}[B_j^{\top}]_{\nu',\alpha'}
        [B_k]_{\beta',n'}\,\times\\
        &\quad\;\Big[\mathbb{E}_{V\in \mathrm{O}(K)}\big[V_{\alpha,\rho}V_{\beta,\rho}V_{\alpha',\rho'}V_{\beta',\rho'}\big]-\\
        &\quad\;\;\,\mathbb{E}_{V\in \mathrm{O}(K)}\big[V_{\alpha,\rho}V_{\beta,\rho}\big]\mathbb{E}_{V\in \mathrm{O}(K)}\big[V_{\alpha',\rho'}V_{\beta',\rho'}\big]\Big]\\
        &=\sum_{\rho}\frac{s^4_{\rho}}{K^2}\sum_{\nu,n}\iota_{\nu}(\boldsymbol{\theta})\,\iota_{n}(\boldsymbol{\theta})\sum_{\nu',n'}\iota_{\nu'}(\boldsymbol{\theta})\,\iota_{n'}(\boldsymbol{\theta})\,\times\\
        &\quad\;\sum_{\alpha,\beta}\Big([B_j^{\top}]_{\nu,\alpha}
        [B_k]_{\beta,n}[B_j^{\top}]_{\nu',\alpha}
        [B_k]_{\beta,n'}\,+\\
        &\quad\quad\quad\;\;[B_j^{\top}]_{\nu,\alpha}
        [B_k]_{\beta,n}[B_j^{\top}]_{\nu',\beta}
        [B_k]_{\alpha,n'}\Big)+\mathcal{O}(K^{-3})\\
        &=\frac{\mathrm{tr}(S^4)}{K^2}\,\Big[\big(\boldsymbol{\iota}(\boldsymbol{\theta})\big|B_j^{\top}B_k\big|\boldsymbol{\iota}(\boldsymbol{\theta})\big)^2+\\
        &\quad\quad\quad\quad\quad\big(\boldsymbol{\iota}(\boldsymbol{\theta})\big|B_j^{\top}B_j\big|\boldsymbol{\iota}(\boldsymbol{\theta})\big)\big(\boldsymbol{\iota}(\boldsymbol{\theta})\big|B_k^{\top}B_k\big|\boldsymbol{\iota}(\boldsymbol{\theta})\big)\Big]+\\
        &\quad\quad\quad\quad\quad\mathcal{O}(K^{-3}) \,\,.
    \end{split}
\end{equation}
Thus we have
\begin{equation}
    \mathbb{E}_{V\in \mathrm{O}(K)}\big[F_{j,k}(\boldsymbol{\theta})\big]=\frac{\mathrm{tr}(S^2)}{K}\,\big(\boldsymbol{\iota}(\boldsymbol{\theta})\big|B_j^{\top}B_k\big|\boldsymbol{\iota}(\boldsymbol{\theta})\big) \,\,,
\end{equation}
and
\begin{equation}
    \begin{split}
        \mathrm{Var}_{V\in \mathrm{O}(K)}\big[F_{j,k}(\boldsymbol{\theta})\big]&=\frac{\mathrm{tr}(S^4)}{K^2}\,\Big[\big(\boldsymbol{\iota}(\boldsymbol{\theta})\big|B_j^{\top}B_k\big|\boldsymbol{\iota}(\boldsymbol{\theta})\big)^2+\\
        &\quad\quad\quad\quad\quad\big(\boldsymbol{\iota}(\boldsymbol{\theta})\big|B_j^{\top}B_j\big|\boldsymbol{\iota}(\boldsymbol{\theta})\big)\big(\boldsymbol{\iota}(\boldsymbol{\theta})\big|B_k^{\top}B_k\big|\boldsymbol{\iota}(\boldsymbol{\theta})\big)\Big]+\\
        &\quad\quad\quad\quad\quad\mathcal{O}(K^{-3}) \,\,.
    \end{split}
\end{equation}
We now examine the expectation value $\big(\boldsymbol{\iota}(\boldsymbol{\theta})\big|B_j^{\top}B_k\big|\boldsymbol{\iota}(\boldsymbol{\theta})\big)$, to be able to make further statements about $F_{j,k}(\boldsymbol{\theta})$.
\begin{equation}
    \begin{split}
        \big(\boldsymbol{\iota}(\boldsymbol{\theta})\big|B_j^{\top}B_k\big|\boldsymbol{\iota}(\boldsymbol{\theta})\big)&=\bigg(\sum_{\nu_j,\nu_j'}\beta^{(j)}_{\nu_j,\nu_j'}\,\iota^{(j)}_{\nu_j}(\theta_j)\,\iota^{(j)}_{\nu_j'}(\theta_j)\bigg)\bigg(\sum_{\nu_k,\nu_k'}\beta^{(k)}_{\nu_k,\nu_k'}\,\iota^{(k)}_{\nu_k}(\theta_k)\,\iota^{(k)}_{\nu_k'}(\theta_k)\bigg)\times\\
        &\quad\;\;\prod_{m\neq j,k}\bigg(\sum_{\nu_m}\iota^{(m)}_{\nu_m}(\theta_m)\,\iota^{(m)}_{\nu_m}(\theta_m)\bigg) \,\,.
    \end{split}
\end{equation}
Recall the normalization of the basis functions $\iota^{(m)}_{\nu_m}(\theta_m)$, i.e., 
\begin{equation}
    \mathbb{E}_{\theta_m}\Big[\iota^{(m)}_{\nu_m}(\theta_m)\,\iota^{(m)}_{\nu_m'}(\theta_m)\Big]=\delta_{\nu_m,\nu_m'} \,\,.
\end{equation}
Then, since $B_j^{\top}B_k$ is an off-diagonal matrix for $j\neq k$, the expectation value $\big(\boldsymbol{\iota}(\boldsymbol{\theta})\big|B_j^{\top}B_k\big|\boldsymbol{\iota}(\boldsymbol{\theta})\big)$ is suppressed in expectation over $\boldsymbol{\theta}$ in this case. On the other hand, for $j=k$, $B_j^{\top}B_j$ is diagonal and positive semi-definite, hence the expectation value $\big(\boldsymbol{\iota}(\boldsymbol{\theta})\big|B_j^{\top}B_k\big|\boldsymbol{\iota}(\boldsymbol{\theta})\big)$ has a magnitude scaling as $\mathcal{O}(K)$. Summarizing, in expectation over $V\in \mathrm{O}(K)$ and $\boldsymbol{\theta}$ we have
\begin{equation}
    \begin{split}
        &\mathbb{E}\big[F_{j,k}\big]\in
        \begin{cases}
        \mathcal{O}(1)\,\mathrm{tr}(S^2)\;,\quad\mathrm{for}\;j=k\\
        \mathcal{O}(K^{-1})\,\mathrm{tr}(S^2)\;,\quad\mathrm{for}\;j\neq k
        \end{cases}\\
        &\mathrm{Var}\big[F_{j,k}\big]\in
        \mathcal{O}(1)\,\mathrm{tr}(S^4) \,\,.
    \end{split}
\end{equation}

\section{Details on construction of tensorized models} \label{app:details_tensorized_models}
Here we provide more details on the construction of the tensorized models used in the main text. This is based on the tensor network (TN) decomposition of the structure constants $\Gamma$, starting from the following approximation
\begin{equation}
    \Gamma_{\mu,\nu}\approx\sum_{\rho}\sum_{\sigma=1}^{\chi}U_{\mu,\rho}\,T_{\rho,\sigma}\,s_{\sigma}\big[V^{\top}\big]_{\sigma,\nu} \,\,.
\end{equation}
which differs from Eq.~\eqref{eq:SVD_gamma_S} by the presence of an isometry $T$, which is a linear mapping from the $D$-dimensional input functions' space to a $\chi$-dimensional reduced space, building an internal TN representation of the input functions' space. 
We decompose the matrix $V$ as a tensor-train \cite{Oseledets2011,Schollwoeck2011}
\begin{equation}
    V_{\nu,\sigma}\approx\sum_{a_1,...,a_{M-1}=1}^{\chi}\mathcal{V}^{[1]\,\nu_1}_{\sigma,a_1}\,\mathcal{V}^{[2]\,\nu_2}_{a_1,a_2}\,...\mathcal{V}^{[M]\,\nu_M}_{a_{M-1},1} \,\,,
\end{equation}
where $\mathcal{V}^{[m]}$ are rank-3 tensors satisfying the right-normalization condition
\begin{equation}
    \sum_{\nu_m=1}^{\tilde{d}}\sum_{a_m=1}^{\chi}\mathcal{V}^{[m]\,\nu_m}_{a_{m-1},a_m}\,\mathcal{V}^{[m]\,\nu_m}_{a_{m-1}',a_m}=\delta_{a_{m-1}',a_{m-1}} \,\,,
\end{equation}
in order for $V$ to have orthonormal columns. The TN decomposition of $V$ admits the following graphical representation
\begin{align}
    V_{\nu,\sigma}=\;
    \begin{tikzpicture} [baseline={([yshift=+0ex]current bounding box.center)}]
        \node[] (rho) at (0,-1) {$\sigma$};
        \node[rectangle,minimum width=10em,draw] (V) at (2.5,-1) {$V$};
        \node[] (nu1) at (1,0) {$\nu_1$};
        \node[] (nu2) at (2,0) {$\nu_2$};
        \node[] (dots) at (3,0) {$\cdots$};
        \node[] (nuM) at (4,0) {$\nu_M$};
        \draw [thick] (rho) -- (V);
        \draw [thick] (nu1) -- (nu1 |- V.north);
        \draw [thick] (nu2) -- (nu2 |- V.north);
        \draw [thick] (nuM) -- (nuM |- V.north);
    \end{tikzpicture}
    \;\approx\;
    \begin{tikzpicture} [baseline={([yshift=+0ex]current bounding box.center)}]
        \node[] (rho) at (0,-1) {$\sigma$};
        \node[rectangle, draw] (V1) at (1,-1) {$\mathcal{V}^{[1]}$};
        \node[rectangle, draw] (V2) at (2.1,-1) {$\mathcal{V}^{[2]}$};
        \node[] (Vdots) at (3.2,-1) {$\cdots$};
        \node[rectangle, draw] (VM) at (4.3,-1) {$\mathcal{V}^{[M]}$};
        \node[] (nu1) at (1,0) {$\nu_1$};
        \node[] (nu2) at (2.1,0) {$\nu_2$};
        \node[] (dots) at (3.2,0) {$\cdots$};
        \node[] (nuM) at (4.3,0) {$\nu_M$};
        \draw [thick] (rho) -- (V1) -- (V2) -- (Vdots) -- (VM);
        \draw [thick] (nu1) -- (nu1 |- V1.north);
        \draw [thick] (nu2) -- (nu2 |- V2.north);
        \draw [thick] (nuM) -- (nuM |- VM.north);
    \end{tikzpicture}
\end{align}
In order to construct a random right-normalized tensor train representing $V$, it is sufficient to generate random tensors $\mathcal{V}^{[m]}$ by reshaping randomly generated matrices $\mathbb{V}^{[m]}$ with orthonormal columns (i.e., satisfying $\mathbb{V}^{[m]\,\top}\,\mathbb{V}^{[m]}=I$) with elements $\mathbb{V}^{[m]}_{(\nu_m,a_m),a_{m-1}}=\mathcal{V}^{[m]\,\nu_m}_{a_{m-1},a_m}$. These, by construction, satisfy the above right-normalization condition.
The orthogonal matrix $U$ can be decomposed as an orthogonal matrix product operator (MPO) \cite{Pirvu2010,Hubig2017,Styliaris2025}
\begin{equation} 
    U_{\mu,\rho}\approx\sum_{a_1,...,a_{N-1}=1}^{\chi}\mathcal{U}^{[1]\,\mu_1,\rho_1}_{1,a_1}\,\mathcal{U}^{[2]\,\mu_2,\rho_2}_{a_1,a_2}\,...\,\mathcal{U}^{[M]\,\mu_N,\rho_N}_{a_{N-1},1} \,\,,
\end{equation}
with $\mathcal{U}^{[n]}$ being rank-4 tensors constrained to yield an orthogonal $U$. The TN decomposition of $U$ has the following diagrammatic representation
\begin{align}
    U_{\mu,\rho}=\;
    \begin{tikzpicture} [baseline={([yshift=+0ex]current bounding box.center)}]
        \node[] (mu1) at (0,2) {$\mu_1$};
        \node[] (mu2) at (0,1) {$\mu_2$};
        \node[] (mudots) at (0,0) {$\vdots$};
        \node[] (muN) at (0,-1) {$\mu_N$};
        \node[rectangle,minimum height=10em,draw] (U) at (1,0.5) {$U$};
        \node[] (rho1) at (2,2) {$\rho_1$};
        \node[] (rho2) at (2,1) {$\rho_2$};
        \node[] (rhodots) at (2,0) {$\vdots$};
        \node[] (rhoN) at (2,-1) {$\rho_N$};
        \draw [thick] (mu1) -- (mu1 -| U.west);
        \draw [thick] (mu2) -- (mu2 -| U.west);
        \draw [thick] (muN) -- (muN -| U.west);
        \draw [thick] (rho1) -- (rho1 -| U.east);
        \draw [thick] (rho2) -- (rho2 -| U.east);
        \draw [thick] (rhoN) -- (rhoN -| U.east);
    \end{tikzpicture}
    \;\approx\;
    \begin{tikzpicture} [baseline={([yshift=+0ex]current bounding box.center)}]
        \node[] (mu1) at (0,2) {$\mu_1$};
        \node[] (mu2) at (0,1) {$\mu_2$};
        \node[] (mudots) at (0,0) {$\vdots$};
        \node[] (muN) at (0,-1) {$\mu_N$};
        \node[rectangle, draw] (U1) at (1,2) {$\mathcal{U}^{[1]}$};
        \node[rectangle, draw] (U2) at (1,1) {$\mathcal{U}^{[2]}$};
        \node[] (Udots) at (1,0) {$\vdots$};
        \node[rectangle, draw] (UN) at (1,-1) {$\mathcal{U}^{[N]}$};
        \node[] (rho1) at (2,2) {$\rho_1$};
        \node[] (rho2) at (2,1) {$\rho_2$};
        \node[] (rhodots) at (2,0) {$\vdots$};
        \node[] (rhoN) at (2,-1) {$\rho_N$};
        \draw [thick] (U1) -- (U2) -- (Udots) -- (UN);
        \draw [thick] (mu1) -- (mu1 -| U1.west);
        \draw [thick] (mu2) -- (mu2 -| U2.west);
        \draw [thick] (muN) -- (muN -| UN.west);
        \draw [thick] (rho1) -- (rho1 -| U1.east);
        \draw [thick] (rho2) -- (rho2 -| U2.east);
        \draw [thick] (rhoN) -- (rhoN -| UN.east);
    \end{tikzpicture}
\end{align}
In our numerical examples, for constructing random orthogonal MPOs we restrict to the case of the matrix $U$ having a so-called `staircase' structure, which graphically corresponds to
\begin{align}
    U_{\mu,\rho}=\;
    \begin{tikzpicture} [baseline={([yshift=+0ex]current bounding box.center)}]
        \node[] (mu1) at (0,2) {$\mu_1$};
        \node[] (mu2) at (0,1) {$\mu_2$};
        \node[] (mu3) at (0,0) {$\mu_3$};
        \node[] (mu4) at (0,-1) {$\mu_4$};
        \node[rectangle,minimum height=10em,draw] (U) at (1,0.5) {$U$};
        \node[] (rho1) at (2,2) {$\rho_1$};
        \node[] (rho2) at (2,1) {$\rho_2$};
        \node[] (rho3) at (2,0) {$\rho_3$};
        \node[] (rho4) at (2,-1) {$\rho_4$};
        \draw [thick] (mu1) -- (mu1 -| U.west);
        \draw [thick] (mu2) -- (mu2 -| U.west);
        \draw [thick] (mu3) -- (mu3 -| U.west);
        \draw [thick] (mu4) -- (mu4 -| U.west);
        \draw [thick] (rho1) -- (rho1 -| U.east);
        \draw [thick] (rho2) -- (rho2 -| U.east);
        \draw [thick] (rho3) -- (rho3 -| U.east);
        \draw [thick] (rho4) -- (rho4 -| U.east);
    \end{tikzpicture}
    \;\approx\;
    \begin{tikzpicture} [baseline={([yshift=+0ex]current bounding box.center)}]
        \node[] (mu1) at (0,2) {$\mu_1$};
        \node[] (mu2) at (0,1) {$\mu_2$};
        \node[] (mu3) at (0,0) {$\mu_3$};
        \node[] (mu4) at (0,-1) {$\mu_4$};
        \node[rectangle, minimum height=4em, draw] (U1) at (1,1.5) {$U^{[1,2]}$};
        \node[rectangle, minimum height=4em, draw] (U2) at (2.2,0.5) {$U^{[2,3]}$};
        \node[rectangle, minimum height=4em, draw] (U3) at (3.4,-0.5) {$U^{[3,4]}$};
        \node[] (rho1) at (4.5,2) {$\rho_1$};
        \node[] (rho2) at (4.5,1) {$\rho_2$};
        \node[] (rho3) at (4.5,0) {$\rho_3$};
        \node[] (rho4) at (4.5,-1) {$\rho_4$};
        \draw [thick] (mu1) -- (mu1 -| U1.west);
        \draw [thick] (mu2) -- (mu2 -| U1.west);
        \draw [thick] (mu3) -- (mu3 -| U2.west);
        \draw [thick] (mu4) -- (mu4 -| U3.west);
        %\draw [thick] (U1.east) -- (U1.east -| U2.west);
        \draw [thick] (1.53, 1.0) -- (1.68,1.0);
        %\draw [thick] (U2.east) -- (U2.east -| U3.west);
        \draw [thick] (2.73, 0.0) -- (2.88,0.0);
        \draw [thick] (rho1) -- (rho1 -| U1.east);
        \draw [thick] (rho2) -- (rho2 -| U2.east);
        \draw [thick] (rho3) -- (rho3 -| U3.east);
        \draw [thick] (rho4) -- (rho4 -| U3.east);
    \end{tikzpicture}
\end{align}
with $U^{[n,n+1]}$ being $d^2\times d^2$ random orthogonal matrices that can be decomposed as 2-site orthogonal MPOs with bond dimension $\chi=d^2$ via SVD as follows
\begin{equation}
\begin{split}
    U^{[n,n+1]}_{(\mu_n,\rho_n),(\mu_{n+1},\rho_{n+1})}&=\sum_{a_n=1}^{\chi}\tilde{U}_{(\mu_n,\rho_n),a_n}\,\tilde{S}_{a_n,a_n}\,\big[\tilde{V}^{\top}\big]_{a_n,(\mu_{n+1},\rho_{n+1})}\\
    &\equiv\sum_{a_n=1}^{\chi}\tilde{\mathcal{U}}^{[n]\,\mu_n,\rho_n}_{1,a_n}\,\tilde{\mathcal{U}}^{[n]\,\mu_{n+1},\rho_{n+1}}_{a_n,1} \,\,,
\end{split}
\end{equation}
with $\tilde{\mathcal{U}}^{[n]\,\mu_n,\rho_n}_{1,a_n}=\tilde{U}_{(\mu_n,\rho_n),a_n}\,\tilde{S}_{a_n,a_n}$ and $\tilde{\mathcal{U}}^{[n]\,\mu_{n+1},\rho_{n+1}}_{a_n,1}=\big[\tilde{V}^{\top}\big]_{a_n,(\mu_{n+1},\rho_{n+1})}$. The tensors $\mathcal{U}^{[n]}$ are then simply constructed as follows
\begin{equation}
    \mathcal{U}^{[n]\,\mu_n,\rho_n}_{a_{n-1},a_n}=\sum_{\rho_{n}'=1}^{d}\tilde{\mathcal{U}}^{[n-1]\,\mu_n,\rho_n'}_{a_{n-1},1}\,\tilde{\mathcal{U}}^{[n]\,\rho_{n}',\rho_{n}}_{1,a_n} \,\,.
\end{equation}
Finally, the isometry $T$ can be decomposed as a tree tensor network (TTN) \cite{Shi2006,Tagliacozzo2009,Murg2010}
\begin{equation}
\begin{split}
    T_{\rho,\sigma}\approx\sum_{\ell=1}^{\log_2N-1}\sum_{\tau=1}^{N/2^{\ell+1}}\sum_{o^{\ell,\tau}_1,o^{\ell,\tau}_2=1}^{\chi}
    &\mathcal{T}^{[2\tau-1]\,o^{\ell,\tau}_1}_{[\ell]\,o^{\ell-1,2\tau-1}_1,o^{\ell-1,2\tau-1}_2}\,
    \mathcal{T}^{[2\tau]\,o^{\ell,\tau}_2}_{[\ell]\,o^{\ell-1,2\tau}_1,o^{\ell-1,2\tau}_2}\,\times\\
    &\mathcal{T}^{[\tau]\,o^{\ell+1,\lceil\tau/2\rceil}_{\lfloor\tau/2\rfloor+1}}_{[\ell+1]\,o^{\ell,\tau}_1,o^{\ell,\tau}_2}\,\,,
\end{split}
\end{equation}
with $o^{0,\tau}_1=\rho_{2\tau-1}$, $o^{0,\tau}_2=\rho_{2\tau}$, $o^{\log_2N,1}_{1}=\sigma$. The tensors $\mathcal{T}^{[\tau]}_{[\ell]}$ are isometric rank-3 tensors satisfying
\begin{equation}
    \sum_{i_1,i_2=1}^{\chi}\mathcal{T}^{[\tau]\,o}_{[\ell]\,i_1,i_2}\,\mathcal{T}^{[\tau]\,o'}_{[\ell]\,i_1,i_2}=\delta_{o',o} \,\,,
\end{equation}
in order for $T$ to be an isometry. The TN decomposition of $T$ has the following diagrammatic representation
\begin{align}
    T_{\rho,\sigma}\approx\;
    \begin{tikzpicture} [baseline={([yshift=+0ex]current bounding box.center)}]
        \node[] (rho1) at (0,3) {$\rho_1$};
        \node[] (rho2) at (0,2) {$\rho_2$};
        \node[] (rho3) at (0,1) {$\rho_3$};
        \node[] (rho4) at (0,0) {$\rho_4$};
        \node[] (rho5) at (0,-1) {$\rho_5$};
        \node[] (rho6) at (0,-2) {$\rho_6$};
        \node[] (rho7) at (0,-3) {$\rho_7$};
        \node[] (rho8) at (0,-4) {$\rho_8$};
        \node[trapezium, minimum height=8mm, shape border rotate=270, draw](T11) at (1,2.5) {$\mathcal{T}^{[1]}_{[1]}$};
        \node[trapezium, minimum height=8mm, shape border rotate=270, draw](T12) at (1,0.5) {$\mathcal{T}^{[2]}_{[1]}$};
        \node[trapezium, minimum height=8mm, shape border rotate=270, draw](T13) at (1,-1.5) {$\mathcal{T}^{[3]}_{[1]}$};
        \node[trapezium, minimum height=8mm, shape border rotate=270, draw](T14) at (1,-3.5) {$\mathcal{T}^{[4]}_{[1]}$};
        \node[trapezium, minimum height=8mm, shape border rotate=270, draw](T21) at (2.5,1.0) {$\mathcal{T}^{[1]}_{[2]}$};
        \node[trapezium, minimum height=8mm, shape border rotate=270, draw](T22) at (2.5,-2.0) {$\mathcal{T}^{[2]}_{[2]}$};
        \node[trapezium, minimum height=8mm, shape border rotate=270, draw](T31) at (4,-0.5) {$\mathcal{T}^{[1]}_{[3]}$};
        \node[] (sigma) at (5,-0.5) {$\sigma$};
        \draw [thick] (rho1) -- (rho1 -| T11.west);
        \draw [thick] (rho2) -- (rho2 -| T11.west);
        \draw [thick] (rho3) -- (rho3 -| T12.west);
        \draw [thick] (rho4) -- (rho4 -| T12.west);
        \draw [thick] (rho5) -- (rho5 -| T13.west);
        \draw [thick] (rho6) -- (rho6 -| T13.west);
        \draw [thick] (rho7) -- (rho7 -| T14.west);
        \draw [thick] (rho8) -- (rho8 -| T14.west);
        \draw [thick] (sigma) -- (sigma -| T31.east);
        \draw [thick] (T11.east) -- (T21);
        \draw [thick] (T12) -- (T12 -| T21.west);
        \draw [thick] (T13) -- (T13 -| T22.west);
        \draw [thick] (T14.east) -- (T22);
        \draw [thick] (T21.east) -- (T31);
        \draw [thick] (T22.east) -- (T31);
    \end{tikzpicture}
\end{align}
Generating random isometries $\mathcal{T}_{[\ell]}^{[\tau]}$ can be done by simply reshaping a randomly generated $\chi^2\times\chi$ matrix $\mathbb{T}_{[\ell]}^{[\tau]}$ with $\chi$ orthonormal columns.

\section{Construction of biased and unbiased models} \label{app:biased_unbiased_models_S}
Here we provide details on the construction of biased and unbiased models used in the main text, for both non-tensorized and tensorized models. We define the data-generating function $y(\boldsymbol{x})$ to be
\begin{equation}
    y(\boldsymbol{x})=\sum_{\mu=1}^{D}\sum_{\nu=1}^{K}e_{\mu}(\boldsymbol{x})\,\iota_{\nu}(\boldsymbol{\theta}^*)\sum_{\rho=1}^{R}s_{\rho}\,U^{\mathrm{(d)}}_{\mu,\rho}\,\big[V^{\mathrm{(d)}\,\top}\big]_{\rho,\nu}\,\,,
    \label{eq:data_gen_function_S}
\end{equation}
with $R<D$, $\boldsymbol{\theta}^*$ a given parameter configuration, and $V^{\mathrm{(d)}}$ satisfying the property $\sum_{\nu}V^{\mathrm{(d)}}_{\nu,\rho}\,\iota_{\nu}(\boldsymbol{\theta}^*)=0$ for $\rho=R+1,...,K$.
We generate the matrix $V^{\mathrm{(d)}}$ as $V^{\mathrm{(d)}}=\big[\tilde{V}\quad\tilde{W}\big]$, i.e., by horizontally stacking a $K\times R$ matrix $\tilde{V}$ with randomly drawn orthonormal columns, and a $K\times(D-R)$ matrix $\tilde{W}$ constructed via Gram-Schmidt orthogonalization in order to satisfy the constraints
\begin{equation}
\begin{cases}
    \tilde{V}^{\top}\tilde{W} = 0 \\
    \tilde{W}^{\top}\tilde{V} = 0 \\
    \tilde{W}^{\top}\big|\boldsymbol{\iota}(\boldsymbol{\theta}^*)\big)=0\;,\;\mathrm{i.e.},\;\sum_{\nu}\tilde{W}_{\nu,\sigma}\,\iota_{\nu}(\boldsymbol{\theta}^*)=0\;\;\forall\;\sigma=1,...,D-R \,\,.
\end{cases}
\end{equation}
As discussed in the main text, the construction of \emph{biased} models, i.e., models that for $\boldsymbol{\theta}=\boldsymbol{\theta}^*$ exactly match the data-generating function $y(\boldsymbol{x})$, is achieved by setting $U=U^{\mathrm{(d)}}$ and $V=V^{\mathrm{(d)}}$ in their structure constants (see Eq.~\eqref{eq:SVD_gamma_S}), while \emph{unbiased} models are constructed by randomly drawing $U$ and $V$ independently of $U^{\mathrm{(d)}}$ and $V^{\mathrm{(d)}}$.
The values $s_{\rho}$ ($\rho=1,...,D$) of the correlation spectrum are chosen to be $s_{\rho}=1/\sqrt{D}$, i.e., a uniform correlation spectrum with normalization $\sum_{\rho}s_{\rho}^2=1$. This uniform choice is performed in order to have that \emph{full} models $f^{\mathrm{(f)}}_{\boldsymbol{\theta}}(\boldsymbol{x})$ specified by the structure constants
\begin{equation}
    \Gamma^{\mathrm{(f)}}_{\mu,\nu}=\sum_{\rho=1}^{D}U_{\mu,\rho}\,s_{\rho}\,\big[V^{\top}\big]_{\rho,\nu} \,\,,
    \label{eq:full_regr_model_gamma_S}
\end{equation}
have, with high probability, a high effective dimension (since $\sum_{\rho}s_{\rho}^4=1/D$). \emph{Cutoff} models $f^{\mathrm{(c)}}_{\boldsymbol{\theta}}(\boldsymbol{x})$ specified by the structure constants
\begin{equation}
    \Gamma^{\mathrm{(c)}}_{\mu,\nu}=\sum_{\rho=1}^{R}U_{\mu,\rho}\,s_{\rho}\,\big[V^{\top}\big]_{\rho,\nu}+\sum_{\rho=R+1}^{D}U_{\mu,\rho}\,\mathrm{e}^{-\frac{\rho-R}{\xi}}s_{\rho}\,\big[V^{\top}\big]_{\rho,\nu} \,\,,
    \label{eq:cut_regr_model_gamma_S}
\end{equation}
with a positive decay rate $\xi$, have instead a lower effective dimension than full models, with high probability.

The construction of biased models for tensorized models follows the same idea, with suitable modifications to accommodate for the tensor structure of the matrix of singular vectors $V$. The idea is again to find a matrix $\tilde{W}$ such that $\sum_{\nu}\tilde{W}_{\nu,\sigma}\,\iota_{\nu}(\boldsymbol{\theta}^*)=0\;\;\forall\;\sigma=1,...,D-R$, where now $\tilde{W}$ has the following tensor-train decomposition
\begin{equation}
    \tilde{W}_{(\nu_1,...,\nu_M),\sigma}=\sum_{a_1,...,a_{M-1}}\mathcal{W}^{[1]\,\nu_1}_{\sigma,a_1}\,\mathcal{W}^{[2]\,\nu_2}_{a_1,a_2}\,...\,\mathcal{W}^{[M]\,\nu_M}_{a_{M-1},1} \,\,.
\end{equation}
The condition that $\big|\boldsymbol{\iota}(\boldsymbol{\theta})\big)$ is in the kernel of $\tilde{W}$ can be rewritten as
\begin{equation}
    \sum_{a_1,...,a_{M-1}}\prod_{m=1}^M\bigg[\sum_{\nu_m}\mathcal{W}^{[m]\,\nu_m}_{a_{m-1},a_m}\,\iota^{(m)}_{\nu_m}(\theta^*_m)\bigg]=0 \,\,,
\end{equation}
with $a_0\equiv\sigma$ and $a_M\equiv 1$. We now construct the matrices $\mathcal{Q}^{[m]}(\theta^*_m)$ with elements
\begin{equation}
    \big[\mathcal{Q}^{[m]}(\theta^*_m)\big]_{a_{m-1},a_m}\equiv\sum_{\nu_m}\mathcal{W}^{[m]\,\nu_m}_{a_{m-1},a_m}\,\iota^{(m)}_{\nu_m}(\theta^*_m) \,\,,
\end{equation}
and the vector $\boldsymbol{q}(\boldsymbol{\theta}^*_{2\to M})$ with elements
\begin{equation}
    q_{a_1}(\boldsymbol{\theta}^*_{2\to M})\equiv\big[\mathcal{Q}^{[2]}(\theta^*_2)\,...\mathcal{Q}^{[M]}(\theta^*_M)\big]_{a_1,1} \,\,.
\end{equation}
Using these, we can turn the conditions on $\tilde{W}$ into the following conditions for the single tensor $\mathcal{W}^{[1]}$
\begin{equation}
\begin{cases}
    \sum_{a_1}\sum_{\nu_1}\mathcal{W}^{[1]\,\nu_1}_{\sigma,a_1}\,\mathcal{W}^{[1]\,\nu_1}_{\sigma',a_1}=\delta_{\sigma,\sigma'}\quad\text{from right-normalization condition} \\
    \sum_{a_1}\sum_{\nu_1}\mathcal{W}^{[1]\,\nu_1}_{\sigma,a_1}\,q_{a_1}(\boldsymbol{\theta}^*_{2\to M})\,\iota^{(1)}_{\nu_1}(\theta^*_1)=0 \,\,,
\end{cases}
\end{equation}
under the assumption that all other tensors $\mathcal{W}^{[m>1]}$ are already right-normalized. These conditions on the tensor $\mathcal{W}^{[1]}$ can be easily cast in matrix form, by grouping the indices $\gamma_1\equiv(a_1,\nu_1)$, reshaping $\mathcal{W}^{[1]}$ into a matrix $\mathbb{W}^{[1]}$ with elements
\begin{equation}
    \mathbb{W}^{[1]}_{\sigma,\gamma_1}=\mathbb{W}^{[1]}_{\sigma,(a_1,\nu_1)}=\mathcal{W}^{[1]\,\nu_1}_{\sigma,a_1} \,\,,
\end{equation}
and constructing the vector $\mathbf{v}(\boldsymbol{\theta}^*)$ as tensor product $\boldsymbol{q}(\boldsymbol{\theta}^*_{2\to M})\otimes\boldsymbol{\iota}^{(1)}(\theta^*_1)$, with elements
\begin{equation}
    \mathrm{v}_{\gamma_1}(\boldsymbol{\theta}^*)=\mathrm{v}_{(a_1,\nu_1)}(\boldsymbol{\theta}^*)=q_{a_1}(\boldsymbol{\theta}^*_{2\to M})\,\iota^{(1)}_{\nu_1}(\theta^*_1) \,\,.
\end{equation}
In matrix form
\begin{equation}
\begin{cases}
    \mathbb{W}^{[1]}\,\mathbb{W}^{[1]\,\top}=I\quad\text{from right-normalization condition} \\
    \mathbb{W}^{[1]}\,\mathbf{v}(\boldsymbol{\theta}^*)=\mathbf{0} \,\,.
\end{cases}
\end{equation}
Using this, the construction of fully biased tensorized models is summarized in the following steps
\begin{enumerate}
    \item Choose a (random) parameter configuration $\boldsymbol{\theta}^*$.
    \item Construct a set of (random) right-normalized tensors $\{\mathcal{W}^{[2]\,\nu_2}_{a_1,a_2}\,,...,\mathcal{W}^{[M]\,\nu_M}_{a_{M-1},1}\}$ and compute the vector $\mathbf{v}(\boldsymbol{\theta}^*)=\boldsymbol{q}(\boldsymbol{\theta}^*_{2\to M})\otimes\boldsymbol{\iota}^{(1)}(\theta^*_1)$ as described before.
    \item Via Gram-Schmidt orthogonalization, generate a random matrix $\mathbb{W}^{[1]}$ whose columns are normalized, mutually orthogonal and all orthogonal to $\mathbf{v}(\boldsymbol{\theta}^*)$. This matrix satisfies the above conditions by construction.
    \item Reshape $\mathbb{W}^{[1]}$ to the order-three tensor $\mathcal{W}^{[1]}$.
    \item The matrix $\tilde{W}_{(\nu_1,...,\nu_M),\sigma}=\sum_{a_1,...,a_{M-1}}\mathcal{W}^{[1]\,\nu_1}_{\sigma,a_1}\,\mathcal{W}^{[2]\,\nu_2}_{a_1,a_2}\,...\,\mathcal{W}^{[M]\,\nu_M}_{a_{M-1},1}$ satisfies the desired orthogonality conditions for constructing a biased model.
\end{enumerate}
Given the tensor-train representation 
\begin{equation}
    V^{\mathrm{(d)}}_{(\nu_1,...,\nu_M),\rho}=\sum_{a_1,...,a_{M-1}}\mathcal{V}^{[1]\,\nu_1}_{\rho,a_1}\,\mathcal{V}^{[2]\,\nu_2}_{a_1,a_2}\,...\,\mathcal{V}^{[M]\,\nu_M}_{a_{M-1},1} \,\,,
\end{equation}
the construction of partially biased tensorized models is obtained by perturbing the tensors $\mathcal{V}^{[m]}$. Specifically we construct $V^{\mathrm{(d)}}_{\epsilon}$ in the data-generating function from tensors $\mathcal{V}^{[m]}_{\epsilon}$ that are obtained by reshaping the orthogonal matrices $\mathbb{V}^{[m]}_{\epsilon}=\mathrm{ortho}(\mathbb{V}^{[m]}+\epsilon\,\mathbb{G})$, where $\mathbb{G}$ is a matrix with Gaussian entries.

\section{Structure constants and neural tangent kernel} \label{app:relation_with_NTK}
In this section, we connect our analysis of Fourier regression models with the neural tangent kernel (NTK) \cite{Jacot2018}, showing how it is influenced by the structure constants of the model. We start by setting the notation used throughout this and the next section. We write the model expression as follows:
\begin{equation}
    f_{\boldsymbol{\theta}}(\boldsymbol{x})=\boldsymbol{e}(\boldsymbol{x})^\top \,\Gamma\,\boldsymbol{\iota}(\boldsymbol{\theta})
    \label{eq:model-vector}
\end{equation}
where we introduce the vectors
\begin{equation}
    \boldsymbol{e}(\boldsymbol{x})
    =\bigl(e_1(\boldsymbol{x}),\ldots,e_D(\boldsymbol{x})\bigr)^\top\in\mathbb{R}^D,
    \qquad
    \boldsymbol{\iota}(\boldsymbol{\theta})
    =\bigl(\iota_1(\boldsymbol{\theta}),\ldots,\iota_K(\boldsymbol{\theta})\bigr)^\top\in\mathbb{R}^K\,\,.
    \label{eq:feature-vectors}
\end{equation}
We denote the SVD of the structure constants as $\Gamma=USV^\top$. We introduce the parameter-basis Jacobian as the $K\times M$ matrix with elements
\begin{equation}
\left[J(\boldsymbol{\theta})\right]_{\nu,j}=\frac{\partial \iota_{\nu}(\boldsymbol{\theta})}{\partial\theta_j}=\sum_{\kappa=1}^K\left[B_j\right]_{\nu,\kappa}\iota_{\kappa}(\boldsymbol{\theta}) \,\,,
\label{eq:param-jacobian}
\end{equation}
with the $K\times K$ derivative tensor $B_j$ introduced in Section \ref{app:FIM_diagramm_expr_deriv}. This allows us to express the derivatives as
\begin{equation}
\frac{\partial f_{\boldsymbol{\theta}}(\boldsymbol{x})}{\partial\theta_j}=\left[\boldsymbol{e}(\boldsymbol{x})^\top\,\Gamma\,J(\boldsymbol{\theta})\right]_j \,\,.
\label{eq:param-derivs}
\end{equation}
We also introduce the $D\times M$ matrix $G(\boldsymbol{\theta})$ defined as
\begin{equation}
    G(\boldsymbol{\theta})=\Gamma\,J(\boldsymbol{\theta}) \,\,,
\end{equation}
with columns $\left[\boldsymbol{g}^1(\boldsymbol{\theta}) \,\dots\,\boldsymbol{g}^M(\boldsymbol{\theta})\right]$ with $\boldsymbol{g}^j(\boldsymbol{\theta})\in\mathbb{R}^D$ representing the coefficient vector in input space of the model derivative w.r.t.~the $j$-th parameter.

\subsection{Neural tangent kernel: definition, decomposition and relation to FIM}
\label{app:subs:ntk_and_fim}
For a differentiable parameterized model $f_{\boldsymbol{\theta}}(\boldsymbol{x})$, the neural tangent kernel (NTK) at parameter value $\boldsymbol{\theta}$ is \cite{Jacot2018}
\begin{equation}
    K_{\boldsymbol{\theta}}(\boldsymbol{x},\boldsymbol{x}')
    =\nabla_{\boldsymbol{\theta}}f_{\boldsymbol{\theta}}(\boldsymbol{x})^\top
     \nabla_{\boldsymbol{\theta}}f_{\boldsymbol{\theta}}(\boldsymbol{x}')
    =\sum_{j=1}^{M}
      \frac{\partial f_{\boldsymbol{\theta}}(\boldsymbol{x})}{\partial\theta_j}
      \frac{\partial f_{\boldsymbol{\theta}}(\boldsymbol{x}')}{\partial\theta_j}.
    \label{eq:ntk-definition}
\end{equation}
Here $\boldsymbol{x}$ and $\boldsymbol{x}'$ are two inputs, and $\nabla_{\boldsymbol{\theta}}f_{\boldsymbol{\theta}}(\boldsymbol{x})$ is the $M$-dimensional gradient of the model output with respect to the parameters. The NTK measures how gradient-based training changes function values. In the lazy-training or linearized regime, modes with larger NTK eigenvalues are learned faster under gradient flow \cite{Jacot2018,Fort2020,Loo2022,Amini2022}. For a model of the form of Eq.~\eqref{eq:model-vector} it is easy to show that FIM and NTK can be expressed as
\begin{align}
&F(\boldsymbol{\theta})=G(\boldsymbol{\theta})^\top G(\boldsymbol{\theta}) \,\,,\\
&K_{\boldsymbol{\theta}}(\boldsymbol{x},\boldsymbol{x}')=\boldsymbol{e}(\boldsymbol{x})^\top \,G(\boldsymbol{\theta})G(\boldsymbol{\theta})^\top\,\boldsymbol{e}(\boldsymbol{x}')\equiv\boldsymbol{e}(\boldsymbol{x})^\top \,\mathcal{K}(\boldsymbol{\theta})\,\boldsymbol{e}(\boldsymbol{x}') \,\,.
\end{align}
Hence, the non-zero spectrum of the \emph{population NTK matrix} $\mathcal{K}(\boldsymbol{\theta})$ is identical to that of the FIM. Denoting with $G(\boldsymbol{\theta})=\mathcal{U}(\boldsymbol{\theta})\Sigma(\boldsymbol{\theta})\mathcal{V}(\boldsymbol{\theta})^\top$ the SVD of $G(\boldsymbol{\theta})$, with $\Sigma(\boldsymbol{\theta})$ the diagonal matrix of singular values shared between the FIM and $\mathcal{K}(\boldsymbol{\theta})$, the NTK can be written as
\begin{equation}
    K_{\boldsymbol{\theta}}(\boldsymbol{x},\boldsymbol{x}')=\boldsymbol{e}^{\mathcal{U}_{\boldsymbol{\theta}}}(\boldsymbol{x})^\top \,\Sigma(\boldsymbol{\theta})^2\,\boldsymbol{e}^{\mathcal{U}_{\boldsymbol{\theta}}}(\boldsymbol{x}') \,\,,
\end{equation}
with $\boldsymbol{e}^{\mathcal{U}_{\boldsymbol{\theta}}}(\boldsymbol{x})=\mathcal{U}(\boldsymbol{\theta})^\top\,\boldsymbol{e}(\boldsymbol{x})$ denoting the \emph{active input functions} of the model at parameter point $\boldsymbol{\theta}$. The observation that $\Sigma(\boldsymbol{\theta})^2$ is precisely the FIM spectrum is very important, as it connects the model capacity measured by the ED, for which we derived explicit bounds and spectral properties based on the structure constants (on average in the parameter space), with the model training dynamics that, at least at late times, can be approximately characterized by the NTK.

At this point, it is important to distinguish between the two cases:
\begin{itemize}
    \item \emph{Under-parameterized regime} $M<D$. In this regime, the NTK matrix $\mathcal{K}(\boldsymbol{\theta})$ is singular with at least $D-M$ zero eigenvalues. In this regime, the number of independent directions in input function space that can be explored us upper-bounded by the number of independent model parameters.
    \item \emph{Over-parameterized regime} $M\geq D$. In this regime, the model has in principle enough parameters to explore all $D$ independent directions in the input function space. Whether this can effectively be achieved depends on the model details, and in particular on the structure constants, and is measured by the ED.
\end{itemize}
In the next section we discuss how these regimes can qualitatively influence the training behavior of a model in a simple example

\subsection{Average form of NTK}
\label{app:subs:ntk_avgs}
To move a step closer to how the structure constants influence the properties of the NTK, we can consider its average over the parameter space, which is calculated as follows
\begin{align}
&\mathbb{E}_{\boldsymbol{\theta}\in\Theta}[K_{\boldsymbol{\theta}}(\boldsymbol{x},\boldsymbol{x}')]=\frac{1}{|\Theta|}\int_\Theta K_{\boldsymbol{\theta}}(\boldsymbol{x},\boldsymbol{x}')\mathrm{d}\boldsymbol{\theta}\\
&=\sum_{\substack{\mu,\kappa,\nu,j \\ \mu',\kappa',\nu'}}e_{\mu}(\boldsymbol{x})^\top \,\Gamma_{\mu,\kappa}\,[B_j]_{\kappa,\nu}\,\mathbb{E}_{\boldsymbol{\theta}\in\Theta}\left[\iota_{\nu}(\boldsymbol{\theta}) \iota_{\nu'}(\boldsymbol{\theta})\right]\,[B_j]_{\kappa',\nu'}\,\Gamma_{\mu',\kappa'}\,e_{\mu'}(\boldsymbol{x}')\\
&=\sum_{\substack{\mu,\kappa \\ \mu',\kappa'}}e_{\mu}(\boldsymbol{x})^\top \,\Gamma_{\mu,\kappa}\,\left[\sum_jB_jB_j^\top\right]_{\kappa,\kappa'}\,\Gamma_{\mu',\kappa'}\,e_{\mu'}(\boldsymbol{x}')\\
&=\sum_{\substack{\mu,\kappa,\rho \\ \mu',\rho'}}e_{\mu}(\boldsymbol{x})^\top \,U_{\mu,\rho}\,S_{\rho,\rho}\,V^\top_{\rho,\kappa}\,B^2_{\kappa,\kappa}\,V_{\kappa,\rho'}\,S_{\rho',\rho'}\,U^\top_{\rho',\mu'}\,e_{\mu'}(\boldsymbol{x}') \,\,,
\end{align}
where in the third line we used $\mathbb{E}_{\boldsymbol{\theta}\in\Theta}\left[\iota_{\nu}(\boldsymbol{\theta}) \iota_{\nu'}(\boldsymbol{\theta})\right]=\delta_{\nu,\nu'}$ and in the last line we set $B^2_{\kappa,\kappa}=\left[\sum_jB_jB_j^\top\right]_{\kappa,\kappa'}$, using the fact that $B_jB_j^\top$ is a diagonal and positive semi-definite matrix. To make further progress, we can again average over the orthogonal group $V\in\mathrm{O}(K)$ obtaining
\begin{align}
&\mathbb{E}_{V\in\mathrm{O}(K)}\mathbb{E}_{\boldsymbol{\theta}\in\Theta}[K_{\boldsymbol{\theta}}(\boldsymbol{x},\boldsymbol{x}')]\\
&=\frac{\sum_{\kappa}B^2_{\kappa,\kappa}}{K}\sum_{\mu,\rho,\mu'}e_{\mu}(\boldsymbol{x})^\top \,U_{\mu,\rho}\,S^2_{\rho,\rho}\,U^\top_{\rho,\mu'}\,e_{\mu'}(\boldsymbol{x}')\\
&=\frac{\sum_{\kappa}B^2_{\kappa,\kappa}}{K}\,\boldsymbol{e}^U(\boldsymbol{x})^\top \,S^2\,\boldsymbol{e}^U(\boldsymbol{x}') \,\,.
\label{eq:avg_NTK_eigen}
\end{align}
Here $\frac{\sum_{\kappa}B^2_{\kappa,\kappa}}{K}$ is a prefactor of $O(1)$, and we see that on average the active input functions the NTK aligns with are the input basis functions dictated by the structure constants $\Gamma$, i.e., its left singular vectors. This is an important observation for the analysis of the training dynamics performed in the next section. In particular, it shows how $\Gamma$ controls the eigenfunctions and the spectrum of the NTK. Using results from classical ML literature, the NTK can be shown to control the dynamics of the training \cite{Jacot2018,Fort2020,Loo2022,Amini2022}. Thus, we can use the above relation, although true only in an approximate sense, to relate the training dynamics to the structure constants $\Gamma$. This ultimately enables us to uncover qualitative differences between the cases where $\Gamma$ is biased towards the data generating function $g(\boldsymbol{x})$ or unbiased, as well as differences between models with hight and low ED, which is controlled by the singular spectrum $S$.

\section{Late-time analysis of training dynamics} \label{app:training_dynamics_analysis}
In this section, we show how model-task alignment can accelerate gradient-descent training of Fourier regression models in a late-training, fixed-NTK regime \cite{Fort2020,Loo2022}. To do this, we develop a simplified model allowing us to draw qualitative conclusions on the behavior of the training dynamics for biased and unbiased models, in high and low ED regimes. In particular, we show that the speed of convergence of training towards the target is strongly influenced by the overlap between the target and the NTK eigenfunctions, and by the magnitude of the related NTK eigenvalues.

We start by setting some useful notation prior to our derivation. We consider the following training setup. We denote the training dataset with
\begin{equation}
    \mathcal{D}_{\mathfrak{n}_{\mathrm{train}}}=\{(\boldsymbol{x}_n,g_n)\}_{n=1}^{\mathfrak{n}_{\mathrm{train}}},
    \qquad
    g_n=g(\boldsymbol{x}_n)\,\,,
    \label{eq:datatrain}
\end{equation}
where $\boldsymbol{x}_n\in\mathcal{X}$ are samples in the input space $\mathcal{X}\subset\mathbb{R}^N$ and $g(\boldsymbol{x})$ is the target or data-generating function. We use the $\underline{\cdot}$ notation to denote vectors of training samples, i.e., $\underline{h}=\bigl(h(\boldsymbol{x}_1),\ldots,h(\boldsymbol{x}_{\mathfrak{n}_{\mathrm{train}}})\bigr)^\top\in\mathbb{R}^{\mathfrak{n}_{\mathrm{train}}}$. The empirical residual vector are defined as
\begin{equation}
    \underline{r}(\boldsymbol{\theta})=\underline{f}({\boldsymbol{\theta}})-\underline{g},
    \qquad
    r_n(\boldsymbol{\theta})=f_{\boldsymbol{\theta}}(\boldsymbol{x}_n)-g(\boldsymbol{x}_n)\,\,,
    \label{eq:empirical-residual}
\end{equation}
It is also useful to define the matrix of data feature vectors $\Phi_{\mathfrak{n}_{\mathrm{train}}}\in\mathbb{R}^{\mathfrak{n}_{\mathrm{train}}\times D}$ with elements
\begin{equation}
    [\Phi_{\mathfrak{n}_{\mathrm{train}}}]_{n,\mu}=e_{\mu}(\boldsymbol{x}_n) \,\,,
\end{equation}
the empirical data Jacobian matrix $\mathcal{J}_{\mathfrak{n}_{\mathrm{train}}}(\boldsymbol{\theta})=\Phi_{\mathfrak{n}_{\mathrm{train}}}\Gamma J(\boldsymbol{\theta})\in\mathbb{R}^{\mathfrak{n}_{\mathrm{train}}\times M}$ with elements
\begin{equation}
    \mathcal{J}_{\mathfrak{n}_{\mathrm{train}}}(\boldsymbol{\theta})_{n,j}=\frac{\partial f_{\boldsymbol{\theta}}(\boldsymbol{x}_n)}{\partial\theta_j}=\left[\boldsymbol{e}(\boldsymbol{x}_n)^\top \,\Gamma\,J(\boldsymbol{\theta})\right]_j \,\,,
\end{equation}
and the \emph{empirical NTK matrix}
\begin{align}
    K_{\mathfrak{n}_{\mathrm{train}}}(\boldsymbol{\theta})&=\frac{1}{{\mathfrak{n}_{\mathrm{train}}}}\mathcal{J}_{\mathfrak{n}_{\mathrm{train}}}(\boldsymbol{\theta})\mathcal{J}_{\mathfrak{n}_{\mathrm{train}}}(\boldsymbol{\theta})^\top=\frac{1}{{\mathfrak{n}_{\mathrm{train}}}}\Phi_{\mathfrak{n}_{\mathrm{train}}}\mathcal{K}(\boldsymbol{\theta})\Phi_{\mathfrak{n}_{\mathrm{train}}}^\top\,\,.
    \label{eq:empirical_NTK}
\end{align}

Using these definitions, we can write the gradient of the MSE loss function $\mathcal L_{\mathfrak{n}_{\mathrm{train}}}(\boldsymbol{\theta})=\frac{1}{{\mathfrak{n}_{\mathrm{train}}}}\sum_{n=1}^{{\mathfrak{n}_{\mathrm{train}}}}r_n(\boldsymbol{\theta})^2$ yielding the update rule of the parameter vector in gradient descent, which in continuous time becomes
\begin{equation}
\frac{\mathrm{d}\boldsymbol{\theta}_t}{\mathrm{d}t}=-\nabla_{\boldsymbol{\theta}}\mathcal L_{\mathfrak{n}_{\mathrm{train}}}(\boldsymbol{\theta}_t)
=-\frac{2}{{\mathfrak{n}_{\mathrm{train}}}}\,\mathcal{J}_{\mathfrak{n}_{\mathrm{train}}}(\boldsymbol{\theta}_t)^\top\,\underline{r}(\boldsymbol{\theta}_t) \,\,,
\label{eq:param-grad-descent}
\end{equation}
which results in the updates for the residuals
\begin{equation}
\frac{\mathrm{d}\underline{r}(\boldsymbol{\theta}_t)}{\mathrm{d}t}=-2K_{\mathfrak{n}_{\mathrm{train}}}(\boldsymbol{\theta}_t)\,\underline{r}(\boldsymbol{\theta}_t) \,\,.
\end{equation}
In the late-training regime, it has been shown that the NTK variations significantly slow down \cite{Fort2020,Loo2022}. In this regime, to make further progress in analytically understanding the effects of a bias in the model training, one can replace the NTK $K_{\mathfrak{n}_{\mathrm{train}}}(\boldsymbol{\theta}_t)$ with a fixed value $K_{\mathfrak{n}_{\mathrm{train}}}(\overline{\boldsymbol{\theta}})\equiv\overline{K}_{\mathfrak{n}_{\mathrm{train}}}$. Using this we can therefore write the residuals evolution as
\begin{equation}
\underline{r}(t)=\exp\bigl(-2\overline{K}_{\mathfrak{n}_{\mathrm{train}}}t\bigr)\,\underline{r}(0) \,\,.
\end{equation}
Expanding in the $\overline{K}_{\mathfrak{n}_{\mathrm{train}}}$ eigenvectors $\overline{K}_{\mathfrak{n}_{\mathrm{train}}}\,\underline{v}_a=\overline{\lambda}_a\,\underline{v}_a$ gives
\begin{align}
\underline{r}(t)&=\sum_a\exp\bigl(-2\overline{\lambda}_at\bigr)\,\bigl(\underline{r}(0)^\top\underline{v}_a\bigr)\,\underline{v}_a\\
&=\sum_a\exp\bigl(-2\overline{\lambda}_at\bigr)\,\bigl(\underline{f}(\boldsymbol{\theta}_0)^\top\underline{v}_a-\underline{g}^\top\underline{v}_a\bigr)\,\underline{v}_a \,\,.
\label{eq:evo_residuals_1}
\end{align}
For wide neural networks or linear models, this expression is exact \cite{Jacot2018}, otherwise it is an approximation valid only for short times in the late-training regime.

Again, to make further progress on the analysis of the training dynamics, we restrict to the case of \emph{representative sampling}, which corresponds to the case where the scalar product of two samples vectors well approximates the inner product in input space, that is, $\frac{1}{\mathfrak{n}_{\mathrm{train}}}\underline{h}^\top\underline{f}\approx\bigl\langle h,f\bigr\rangle_{\mathcal{X}}$. Under this condition, $\frac{1}{\mathfrak{n}_{\mathrm{train}}}\left[\Phi_{\mathfrak{n}_{\mathrm{train}}}^\top\Phi_{\mathfrak{n}_{\mathrm{train}}}\right]_{\mu,\mu'}\approx \bigl\langle e_{\mu},e_{\mu'}\bigr\rangle_{\mathcal{X}}=\delta_{\mu,\mu'}$. Under this condition, one can easily see from Eq.~\eqref{eq:empirical_NTK} that the spectrum of $\overline{K}_{\mathfrak{n}_{\mathrm{train}}}$ becomes well approximated by the spectrum of $\overline{\mathcal{K}}=\mathcal{K}(\overline{\boldsymbol{\theta}})$ (and equivalently of the FIM $\overline{F}=F(\overline{\boldsymbol{\theta}})$). Thus, with the diagonal representation $\overline{\mathcal{K}}=\overline{\mathcal{U}}\,\overline{\Sigma}^2\,\overline{\mathcal{U}}^\top$ with $\overline{\Sigma}^2=\mathrm{diag}\left(\lambda_1,...,\lambda_D\right)$ we can write the residuals evolution in function space as
\begin{equation}
    r_{t}(\boldsymbol{x})=\sum_{k=1}^D\mathrm{e}^{-2\lambda_k t}\left\langle f_0-g,\,\overline{u}_k\right\rangle_{\mathcal{X}}\,\overline{u}_k(\boldsymbol{x})
\end{equation}
with $\overline{u}_k(\boldsymbol{x})=\left[\overline{\mathcal{U}}^\top\boldsymbol{e}(\boldsymbol{x})\right]_k$. At this point, we recall the result of Eq.~\eqref{eq:avg_NTK_eigen}. Although valid only in an average sense, it can guide our analysis further in relating the NTK eigen-decomposition with the structure constants $\Gamma$. In the following, for a qualitative analysis of the ED-bias interplay, we therefore consider the following scenario:
\begin{align}
& \overline{\Sigma}^2\approx S^2\,,\quad \overline{\mathcal{U}}\approx U\,\,,
\end{align}
that is, we set the eigen-decomposition of the NTK matrix in input space $\overline{\mathcal{K}}$ to be the approximately equal to that of the structure constants $\Gamma\Gamma^\top$ (up to overall multiplicative factor of the eigenvalues).

\subsection{Data-generating function}
We now consider the data-generating function and the model to have the form
\begin{align}
    &g(\boldsymbol{x})=\boldsymbol{e}(\boldsymbol{x})^\top U^{(\mathrm{d})}S^{(\mathrm{d})}V^{(\mathrm{d})\,\top}\boldsymbol{\iota}(\boldsymbol{\theta}^*)\\
    &f_{\boldsymbol{\theta}}(\boldsymbol{x})=\boldsymbol{e}(\boldsymbol{x})^\top USV^{\,\top}\boldsymbol{\iota}(\boldsymbol{\theta})
\end{align}
which gives the overlaps
\begin{align}
    &\left\langle g,\,\overline{u}_k\right\rangle_{\mathcal{X}}=\left[\overline{\mathcal{U}}^\top U^{(\mathrm{d})}S^{(\mathrm{d})}V^{(\mathrm{d})\,\top}\boldsymbol{\iota}(\boldsymbol{\theta}^*)\right]_k \,\,,\\
    &\left\langle f_0,\,\overline{u}_k\right\rangle_{\mathcal{X}}=\left[\overline{\mathcal{U}}^\top USV^{\top}\boldsymbol{\iota}(\boldsymbol{\theta}_0)\right]_k\,\,.
    \label{eq:f0_uk_ovlp}
\end{align}
As in the main text, we consider the case of a data generating function with
\begin{equation}
    S^{(\mathrm{d})}_{\rho,\rho}=0\quad\text{for }\rho>R \,\,,
\end{equation}
which we can use to define the \emph{data-relevant} modes as
\begin{align}
w^{(\mathrm{r})}_{a}=\left[U^{(\mathrm{d})\,\top}\boldsymbol{e}\right]_a\,\, \text{ for } a\le R \,\,,
\end{align}
and the \emph{data-irrelevant} modes $w^{(\mathrm{i})}_{b}$, for $b>R$, as some orthogonal complement, forming the orthonormal basis $\{w^{(\mathrm{r})}_{1},...,w^{(\mathrm{r})}_{R},w^{(\mathrm{i})}_{R+1},...,w^{(\mathrm{i})}_{D}\}$. Therefore we have the following expansion (dropping the dependence on $\boldsymbol{x}$ for notational convenience)
\begin{align}
\overline{u}_k&=\sum_{a=1}^R \bigl\langle w^{(\mathrm{r})}_{a},\overline{u}_k\bigr\rangle_{\mathcal{X}}w^{(\mathrm{r})}_{a}+\sum_{b=R+1}^D\bigl\langle w^{(\mathrm{i})}_{b},\overline{u}_k\bigr\rangle_{\mathcal{X}}w^{(\mathrm{i})}_{b}\\&=\sum_{a=1}^R c_k^{(\mathrm{r}),a}\,w^{(\mathrm{r})}_{a}+\sum_{b=R+1}^D c_k^{(\mathrm{i}),b}\,w^{(\mathrm{i})}_{b}\,\,.
\end{align}
which we can use to express the evolution of the residuals along the data-relevant and data-irrelevant components as follows
The residual in the relevant directions is
\begin{align}
  r^{(\mathrm{r})}_a(t)
  &=\sum_{k=1}^{D} \e^{-2\lambda_k t}\,c_k^{(\mathrm{r}),a}\,\Xinner{f_0-g}{\overline{u}_k}\\
  &=\sum_{k=1}^{D} \e^{-2\lambda_k t}\,c_k^{(\mathrm{r}),a}\left[\sum_{a'=1}^R c_k^{(\mathrm{r}),a'}h^{(\mathrm{r})}_{a'} +\sum_{b'=R+1}^D c_k^{(\mathrm{i}),b'}h^{(\mathrm{i})}_{b'}\right]\,\,,\\
  \label{eq:relevant-residual}
\end{align}
and
\begin{align}
  r^{(\mathrm{i})}_b(t)
  &=\sum_{k=1}^{D} \e^{-2\lambda_k t}\,c_k^{(\mathrm{i}),b}\,\Xinner{f_0-g}{\overline{u}_k}\\
  &=\sum_{k=1}^{D} \e^{-2\lambda_k t}\,c_k^{(\mathrm{i}),b}\left[\sum_{a'=1}^R c_k^{(\mathrm{r}),a'}h^{(\mathrm{r})}_{a'} +\sum_{b'=R+1}^D c_k^{(\mathrm{i}),b'}h^{(\mathrm{i})}_{b'}\right]\,\,,\\
  \label{eq:irrelevant-residual}
\end{align}
with $h^{(\mathrm{r})}_{a'}=\Xinner{f_0-g}{w^{(\mathrm{r})}_{a'}}$ and $h^{(\mathrm{i})}_{b'}=\Xinner{f_0-g}{w^{(\mathrm{i})}_{b'}}=\Xinner{f_0}{w^{(\mathrm{i})}_{b'}}$. These expressions constitute the starting points for our comparisons between biased and unbiased models with high and low ED.

\subsection{Unbiased models}
In the unbiased case, the overlap between $w^{(\mathrm{r})}_{a}$ and $\overline{u}_k$ is effectively random. Hence, we can compute the expected value over random realizations of $\overline{u}_k$ (using the same techniques from random matrix theory as used before) of the squares $\left|r^{(\mathrm{r})}_a(t)\right|^2$ and $\left|r^{(\mathrm{i})}_b(t)\right|^2$. We obtain
\begin{align}
\mathbb E_{\overline{u}}\left[\left|r_m(t)\right|^2\right]=\left(\frac{1}{D}\sum_{k=1}^D\e^{-2\lambda_k t}\right)^2\left|h_m\right|^2 + O(D^{-1})\,\,,
\end{align}
with $r_m=r^{(\mathrm{r})}_a,r^{(\mathrm{i})}_b$ and $h_m=h^{(\mathrm{r})}_a,h^{(\mathrm{i})}_b$.

We now distinguish between the two cases of under- and over-parameterized models. In the case of an under-parameterized model ($M<D$), at least $D-M$ NTK eigenvalues are $0$. In this case, the above expectation values cannot decay during training, which represents the fact that if the data-generating function has components along the null-space of the NTK, the difference in these directions cannot be reduced, since the model cannot effectively move along them.
The over-parameterized case $M\geq D$ is a bit more subtle. To be able to draw qualitative conclusions, we make the simplifying assumption of a two-step spectrum, i.e.,
\begin{align}
    & \lambda_{k}=\lambda_0\,, \text{ for }k\le R\,\,,\\
    & \lambda_{k}=\gamma\lambda_0\,\text{ with }0<\gamma<1\,, \text{ for }k>R\,\,,
    \label{eq:two-step-spectrum}
\end{align}
and set $R=\alpha D$ with $0<\alpha<1$, which gives
\begin{align}
\mathbb E_{\overline{u}}\left[\left|r_m(t)\right|^2\right]\approx\left(\alpha\,\e^{-2\lambda_0 t}+(1-\alpha)\,\e^{-2\gamma\lambda_0 t}\right)^2\left|h_m\right|^2\,\,.
\label{eq:unbiased-squared-residualevo}
\end{align}
To infer the speed of convergence for target-relevant and irrelevant modes we consider the instantaneous decay rate:
\begin{align}
\varrho^{(\mathrm{unbiased})}(t)&=-\frac{\mathrm{d}}{\mathrm{d}t}\log\left(\mathbb E_{\overline{u}}\left[\left|r_m(t)\right|^2\right]\right)\\
&=4\gamma\lambda_0+4(1-\gamma)\lambda_0\frac{\alpha\,\e^{-2\lambda_0(1-\gamma)t}}{1-\alpha+\alpha\,\e^{-2\lambda_0(1-\gamma)t}} \,\,.
\end{align}
The initial and the final rates are:
\begin{align}
&\varrho^{(\mathrm{unbiased})}(0)=4\lambda_0\left[\gamma+\alpha(1-\gamma)\right] \,\,,\\
&\varrho^{(\mathrm{unbiased})}(\infty)=4\gamma\lambda_0 \,\,,
\end{align}
for both target-relevant and irrelevant modes. Note that $\gamma$ and $\alpha$ describe the shape of the eigenvalues of the NTK, which is controlled by the shape of the singular values of $\Gamma$. Hence, in we can qualitatively associate higher values of $\gamma$ or $\alpha$ with larger ED of the models, i.e., a larger effective spectral width for the active modes. Note that in this unbiased case, increasing the ED by increasing $\gamma$ or $\alpha$ indeed increases the decay rates, effectively resulting in a more effective training dynamics, consistent with our observations in the main text. As a final note, it is important to mention that while these derivation give a qualitative account of the effects of ED in training, they do not offer a quantitative account of the training dynamics. In particular, while from the previous analysis it appears that in the over-parameterized regime the residuals can decay to 0, it is important to stress that in reality the model is constrained to move on a parameterized \emph{manifold} in input space: even in the over-parameterized regime there could be points in the $\mathbb{R}^D$ that cannot effectively be reached. If this manifold happens to not contain, at least approximately, the data-generating function, the model residuals cannot obviously decay to zero.

\subsection{Biased architecture}
In the biased case, the target-relevant modes $w^{(\mathrm{r})}_{a}$ have significant overlap with the first $R$ NTK eigenfunctions. We therefore set
\begin{align}
&c_k^{\mathrm{(r)},a}\approx\delta_{k,a}\,\text{ for }k\le R\,\,,\quad c_k^{\mathrm{(r)},a}=O(\epsilon)\,\text{ for }k> R\,\,,\\
&c_k^{\mathrm{(i)},b}=O(\epsilon)\,\text{ for }k\le R\,\,,\quad c_k^{\mathrm{(i)},b}\approx\delta_{k,b}\,\text{ for }k> R\,\,,
\label{eq:biased-coeffs}
\end{align}
with $\epsilon\ll 1$. We expand the the squares $\left|r^{(\mathrm{r})}_a(t)\right|^2$ and $\left|r^{(\mathrm{i})}_b(t)\right|^2$ and keep only the $O(1)$ terms, and consider the two-step spectrum scenario of Eq.~\eqref{eq:two-step-spectrum}, which ultimately yields
\begin{align}
&\left|r^{(\mathrm{r})}_a(t)\right|^2=\e^{-4\lambda_0t}\left|\Xinner{f_0-g}{w^{(\mathrm{r})}_a}\right|^2 + O(\epsilon)\,\,,\\
&\left|r^{(\mathrm{i})}_b(t)\right|^2=\e^{-4\gamma\lambda_0t}\left|\Xinner{f_0}{w^{(\mathrm{i})}_b}\right|^2 + O(\epsilon)\,\,.
\label{eq:biased-squared-residualevo}
\end{align}
From the above equations it is already possible to see that the decay rate along the target-relevant directions is always faster than that for unbiased models, since $\varrho^{(\mathrm{unbiased})}(t)<4\lambda_0$. Furthermore, from Eq.~\eqref{eq:f0_uk_ovlp} and under the toy-model approximations made in this section, we see that $\left|\Xinner{f_0}{w^{(\mathrm{i})}_b}\right|^2\approx\gamma\lambda_0\left[V^\top\boldsymbol{\iota}(\boldsymbol{\theta}_0)\right]^2_b$, i.e., in the biased case the model structure induces a reduction of the overlap along the irrelevant directions, and this reduction becomes stronger the lower the $\gamma$ (hence the ED) becomes. Thus, in this simplified model, the decay rate along data-relevant directions is the same for high- and low-ED models, whereas the decay rate along the data-irrelevant directions is smaller for low-ED ones. However, for low-ED models the overlap $\Xinner{f_0}{w^{(\mathrm{i})}_b}$ is structurally suppressed, which means there exists a finite training time for which the residual is indeed smaller for low-ED models than for high-ED ones. This, at least qualitatively, shows a mechanism by which a lower ED, combined with a model alignment with the given regression task, can concentrate the model along the data-relevant subspace of the input function space, and thus result in a better training than a model with higher ED.

The discussion above assumed all $D$ NTK eigenvalues to be $>0$, hence implicitly referred to an over-parameterized regime $M\geq D$. An intuition of the behavior in the case $M<D$ can be gained by choosing $R<D-M$ and setting the last $D-M$ eigenvalues $\lambda_{k>M}=0$. In this case, for the strongly aligned models of this section we can still consider the data-relevant modes $w^{(\mathrm{r})}_a$ to have significant overlap with the first $R$ NTK eigenvalues, and the NTK null-space to be approximately a subspace of the span of the data-irrelevant modes $w^{(\mathrm{i})}_b$. In this situation, there is a subset of residuals with constant amplitude over the training time as $\left|r^{(\mathrm{i})}_{b>M}(t)\right|^2=\left|\Xinner{f_0}{w^{(\mathrm{i})}_{b>M}}\right|^2 + O(\epsilon)$. Also in this situation, the overlap $\Xinner{f_0}{w^{(\mathrm{i})}_{b>M}}$ is suppressed for low-ED models, as they effectively concentrate along the data-relevant directions. This means that the residual along these components cannot decay to zero, and hence it remains large at all times for large ED models. Adding more parameters then ``unfreezes" these components: the residuals acquire then an exponential decay which is faster for models with higher ED, and this then reduces the MSE gap between high- and low-ED models, as observed in the main text.

\subsection{Summary of training analysis}
In summary, our toy model example qualitatively captures the ED-bias interplay in training that we observe in the main text, via the link offered by the NTK. Specifically, we showed the following points:
\begin{itemize}
\item For biased models, the eigenvectors (in the function input space $\mathbb{R}^D$) of the NTK with higher eigenvalues align with the subspace where the data-generating function lives. Equivalently, it is more likely for biased models to find points in the accessible manifold where the high-weight NTK eigenfunctions align with the data-generating function. This results in a rapid decay of the residuals along the data-relevant subspace. The residuals along the data-irrelevant subspace are typically suppressed for biased models, and are more suppressed the lower the ED of the model is, since a lower ED imposes a smaller spectral weight of the corresponding directions. This in turn implies that for finite training times, the MSE can be lower for lower ED models compared to higher ED ones.
\item For unbiased models, the eigenvectors of the NTK are effectively randomly oriented w.r.t.~the subspace where the data-generating function lives. Hence, all NTK eigenvectors become effectively ``spread" across the data-relevant and data-irrelevant subspaces. This spreading of the high-weight NTK eigenvalues on data-irrelevant directions effectively lowers the decay rate of the residuals. Increasing the ED has the overall effect of increasing the non-leading NTK eigenvalues (which potentially carry a significant overlap with data-relevant directions), hence increases the training speed. Furthermore, the accessible manifold of an unbiased model typically does not capture the data-generating function, meaning that the model residuals cannot decay to zero.
\item Transitioning from the under-parameterized regime $M<D$ to the over-parameterized regime $M\geq D$ effectively ``unfreezes" NTK modes previously belonging to the NTK null space. In the biased case, this gives the residuals along these modes a way of decaying, with the decay along these modes being faster for high ED models.
\item Finally, the above analysis was carried out in the limit of representative sampling, where the empirical NTK matrix can effectively be replaced by the NTK defined in function space. For a finite $\mathfrak{n}_{\mathrm{train}}$ not sufficiently resolving the NTK, the empirical NTK controlling the data-resolved residual dynamics deviates from $\mathcal{K}(\boldsymbol{\theta})$. Our analysis, based on the relation between $\mathcal{K}(\boldsymbol{\theta})$ and the model structure constants, needs to be revised in this regime. Qualitatively, for low $\mathfrak{n}_{\mathrm{train}}$ the NTK eigenfunctions cannot align with the data, which effectively slows down the convergence of the training and reduces the gap between the higher and lower ED models. For unbiased models, the NTK eigenfunctions are already somewhat ``random" w.r.t.~the data, hence no significant difference is observed.
\end{itemize}

\section{Further numerical results on effective dimension} \label{app:eff_dim_more_numerics}
\begin{figure*}
    \centering
    \includegraphics[width=0.5\linewidth]{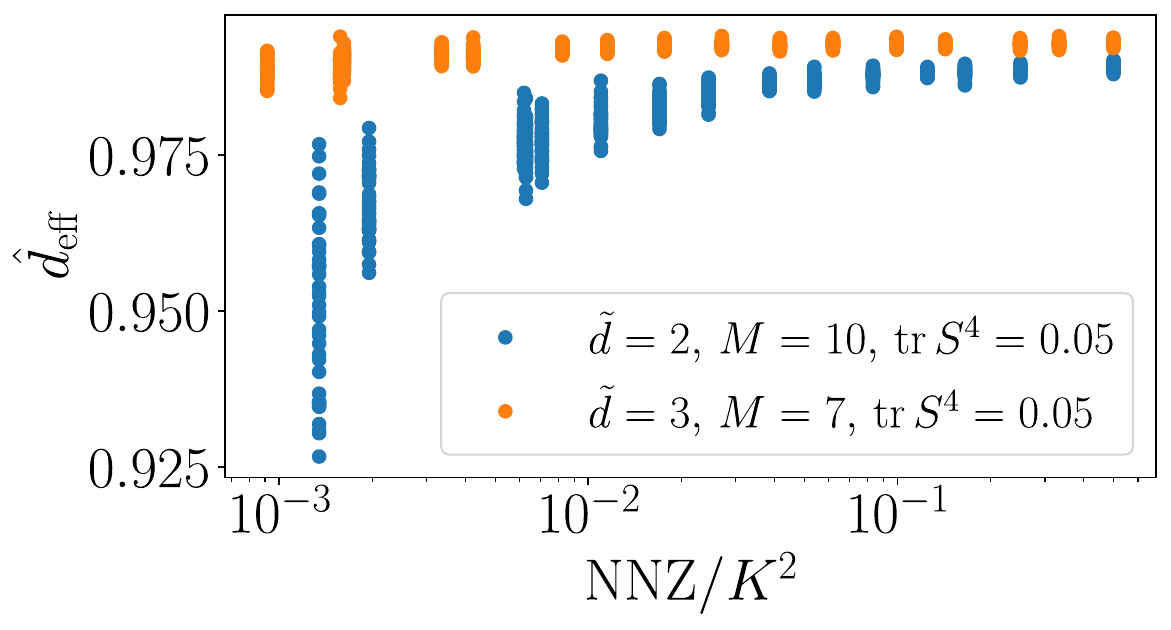}
    \caption{Scaling of normalized ED with the sparsity of the orthogonal matrix $V$. On the $x$ axis, $\mathrm{NNZ}$ corresponds to the number of non-zero elements of $V$ (the lower $\mathrm{NNZ}$, the more sparse $V$ is). Each point corresponds to a random model realization, i.e., a random $\Gamma$ uniformly drawn from $[-1,+1]^{D\times K}$. For every value of $\mathrm{NNZ}$, $40$ model realizations are drawn. The normalized ED is computed using $150$ parameters samples for estimating the normalized FIM.}
    \label{figapp:ED_fullmodels_scaling_sparsity} 
\end{figure*}
\begin{figure*}
    \centering
    \includegraphics[width=\linewidth]{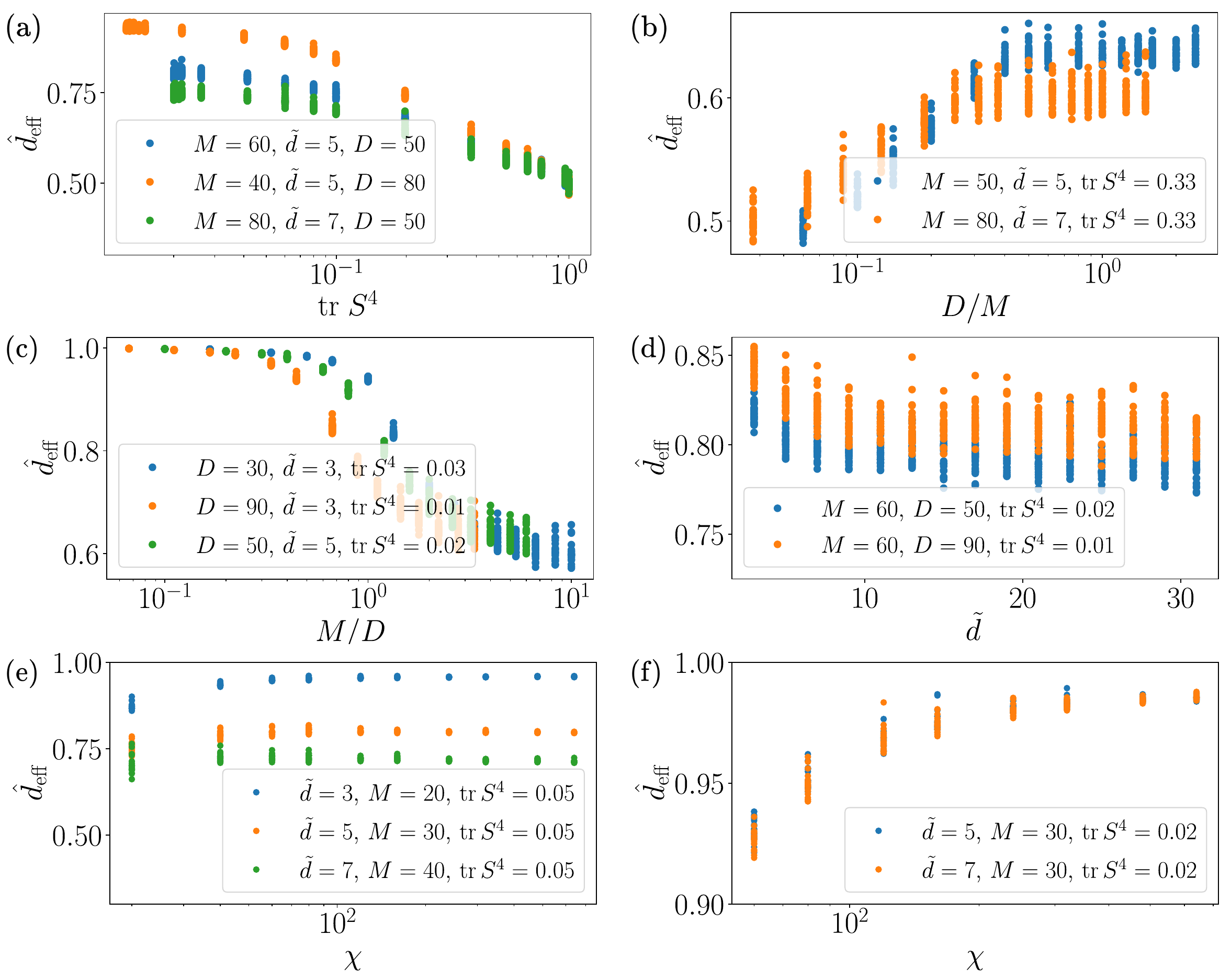}
    \caption{(a) Scaling of $\hat{d}_{\mathrm{eff}}$ with the purity $\mathrm{tr}(S^4)$ of the correlation spectrum ($\chi=100$). (b) Scaling of $\hat{d}_{\mathrm{eff}}$ with the ratio $D/M$. (c) Scaling of $\hat{d}_{\mathrm{eff}}$ with the ratio $M/D$. (d) Scaling of $\hat{d}_{\mathrm{eff}}$ with $\tilde{d}$. (e)-(f) Scaling of $\hat{d}_{\mathrm{eff}}$ with the bond dimension $\chi$. Each point corresponds to a random model realization, i.e., a random generation of $V$ as a right-normalized tensor train. For every value on the $x$ axis, $30$ model realizations are drawn. The normalized ED is computed using $200$ parameters samples for estimating the normalized FIM.}
    \label{figapp:ED_TNmodels_scalings} 
\end{figure*}
Here we provide further numerical results on the dependence of the ED on various model characteristics. The results presented here confirm those shown in the main text, namely that the ED being primarily controlled by the number of input basis functions $D$ when $D<M$, and by $\mathrm{tr}(S^4)$ in the regime $D>M$.

In Fig.~\ref{figapp:ED_fullmodels_scaling_sparsity} we show how the normalized ED depends on the sparsity of the matrix $V$ containing the right singular vectors of the structure constants $\Gamma$. Two different model classes are shown: $\tilde{\mathcal{B}}_m=\{\cos\theta_m,\sin\theta_m\}$ (i.e., $\tilde{d}=2$) and $\tilde{\mathcal{B}}_m=\{1,\cos\theta_m,\sin\theta_m\}$ (i.e., $\tilde{d}=3$). From these results, it is evident that $\hat{d}_{\mathrm{eff}}$ does not strongly depends on the sparsity of $V$.

In Fig.~\ref{figapp:ED_TNmodels_scalings} we show the dependence of the normalized ED on various model characteristics, in the case of tensorized structure constants, that is, with $V$ decomposed as a tensor train. As expected, the ED monotonically decreases in expectation for increasing $\mathrm{tr}(S^4)$ (see panel (a)), and in the overparameterized regime $M>D$ its value also strongly depends on $D$ (see panels (b) and (c)), as expected from the bounds described in Section \ref{app:ED_bound}. The other model characteristics such as $\tilde{d}$ and the bond dimension $\chi$ have little influence on the ED (see panels (d), (e) and (f)), further confirming our expectations that $D>M$ the ED is mostly controlled by $\mathrm{tr}(S^4)$.

\section{Further numerical results on training regression models} \label{app:training_more_numerics}
Here we show further numerical results on the interplay of ED and model bias and their effect on training regression models with gradient descent methods. 
The experiments performed here are analogous to those presented in the main text. we perform several training experiments with randomly drawn data-generating function $y(\boldsymbol{x})$ and structure constants $\Gamma$. For chosen dimensions $D$ and $K$ we draw random instances of $y(\boldsymbol{x})$, and many random instances of models (specified by $\Gamma$) with different degree of bias towards $y(\boldsymbol{x})$. For any given degree of bias, we train several random instances of full models (Eq.~\eqref{eq:full_regr_model_gamma_S}) and cutoff models (Eq.~\eqref{eq:cut_regr_model_gamma_S}), in order to compare the training dynamics of models with higher and lower ED, respectively. For any given degree of bias we consider the minimum MSE attained during training as a proxy for the training quality, denoted with $\mathrm{MSE}_{\mathrm{min}}^{\mathrm{full}}$ and $\mathrm{MSE}_{\mathrm{min}}^{\mathrm{cut}}$ for full and cutoff models, respectively, and study the difference
\begin{equation}
    \Delta_{\mathrm{f-c}}\mathrm{MSE}_{\mathrm{min}}=\mathrm{MSE}_{\mathrm{min}}^{\mathrm{full}}-\mathrm{MSE}_{\mathrm{min}}^{\mathrm{cut}} \,\,,
    \label{eq:DeltaMSE_fullminuscut_S}
\end{equation}
as a function of the difference in the ED between full and cutoff models, i.e., $\hat{d}_{\mathrm{eff}}^{\mathrm{full}}-\hat{d}_{\mathrm{eff}}^{\mathrm{cut}}$. A positive value of $\Delta_{\mathrm{f-c}}\mathrm{MSE}_{\mathrm{min}}$ implies that the full model (with higher ED) trains to a higher MSE compared to the cutoff one, i.e., the model with lower ED model has a better training performance. Conversely, a negative $\Delta_{\mathrm{f-c}}\mathrm{MSE}_{\mathrm{min}}$ implies a better performance of models with higher ED.

\begin{figure*}
    \centering
    \includegraphics[width=\linewidth]{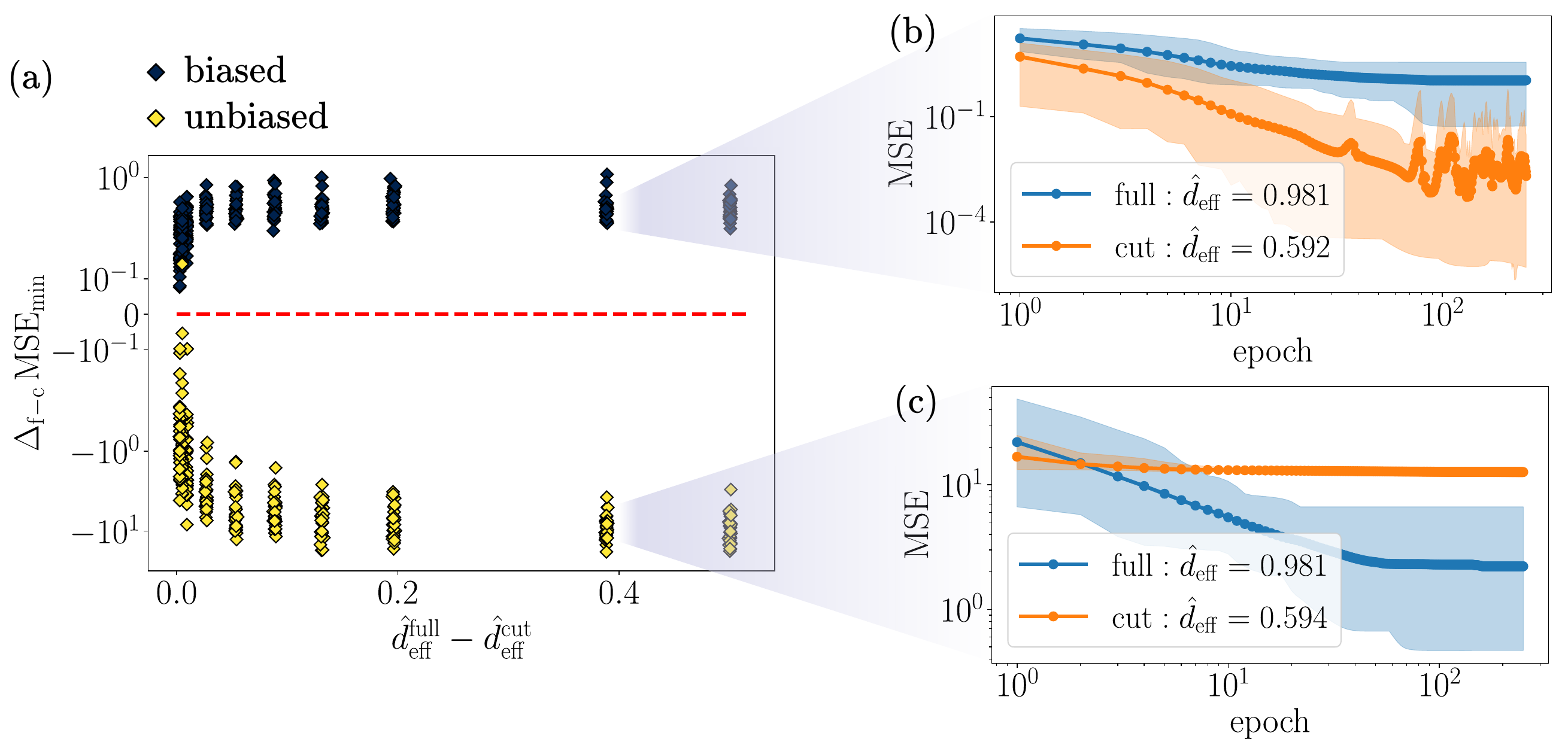}
    \caption{(a) $\Delta_{\mathrm{f-c}}\mathrm{MSE}_{\mathrm{min}}$ for different values of the difference $\hat{d}_{\mathrm{eff}}^{\mathrm{full}}-\hat{d}_{\mathrm{eff}}^{\mathrm{cut}}$. Each point corresponds to $\Delta_{\mathrm{f-c}}\mathrm{MSE}_{\mathrm{min}}$ averaged over $30$ training instances, for a single random model realization. The red line serves as a guide for the eye for zero $\mathrm{MSE}$ difference. (b) Training curves for a random biased model realization. (c) Training curves for a random unbiased model realization. In (b)-(c), the full model is in blue and the cutoff model in orange, and the shading corresponds to the spread over $30$ training instances. 
    For these plots, $N=1$, $\Omega=\{1,...,9\}$ ($d=19$), $\tilde{\mathcal{B}}_m=\{\cos\theta_m,\sin\theta_m\}$ ($\tilde{d}=2$), $M=12$, $R=4$, $\mathfrak{n}_{\mathrm{train}}=30$ with a batch size of $5$.}
    \label{figapp:MSE_BiasUnbias_wTrain_full_Dloc2} 
\end{figure*}

\begin{figure*}
    \centering
    \includegraphics[width=\linewidth]{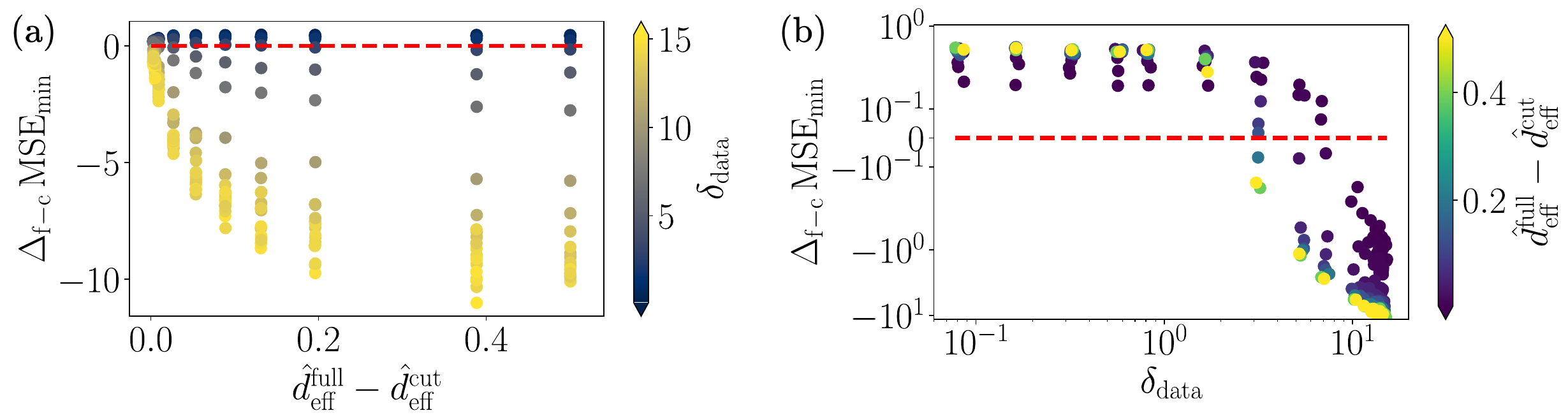}
    \caption{(a) $\Delta_{\mathrm{f-c}}\mathrm{MSE}_{\mathrm{min}}$ for different values of the difference $\hat{d}_{\mathrm{eff}}^{\mathrm{full}}-\hat{d}_{\mathrm{eff}}^{\mathrm{cut}}$. Each point corresponds to $\Delta_{\mathrm{f-c}}\mathrm{MSE}_{\mathrm{min}}$ averaged over $30$ training instances for $30$ random model realization. (b) Same as panel (a) but resolved as a function of $\delta_{\mathrm{data}}$. The red line serves as a guide for the eye for zero $\mathrm{MSE}$ difference. 
    For these plots, $N=1$, $\Omega=\{1,...,9\}$ ($d=19$), $\tilde{\mathcal{B}}_m=\{\cos\theta_m,\sin\theta_m\}$ ($\tilde{d}=2$), $M=12$, $R=4$, $\mathfrak{n}_{\mathrm{train}}=30$ with a batch size of $5$.}
    \label{figapp:MSE_AllBiases_full_Dloc2} 
\end{figure*}

\begin{figure*}
    \centering
    \includegraphics[width=\linewidth]{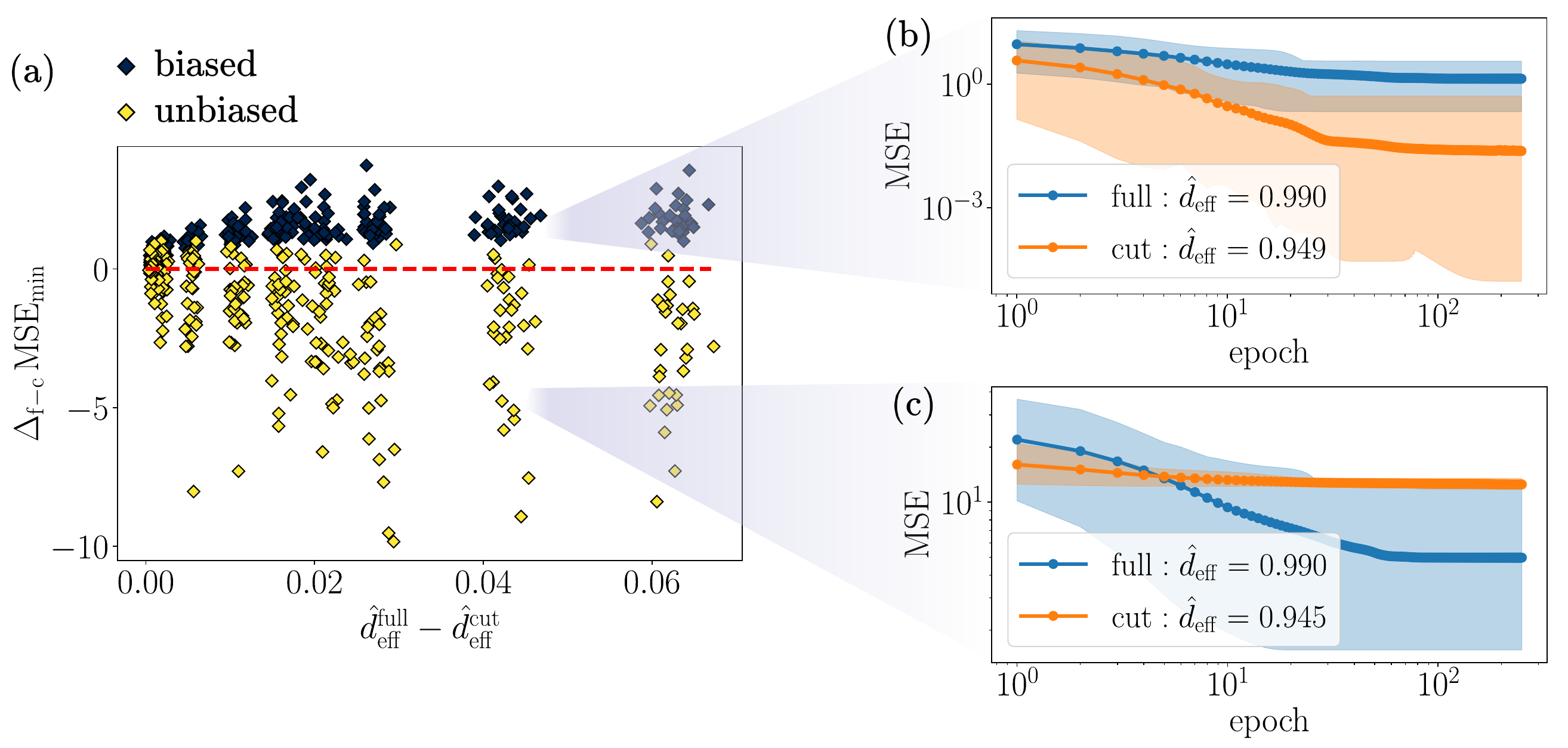}
    \caption{(a) $\Delta_{\mathrm{f-c}}\mathrm{MSE}_{\mathrm{min}}$ for different values of the difference $\hat{d}_{\mathrm{eff}}^{\mathrm{full}}-\hat{d}_{\mathrm{eff}}^{\mathrm{cut}}$. Each point corresponds to $\Delta_{\mathrm{f-c}}\mathrm{MSE}_{\mathrm{min}}$ averaged over $30$ training instances, for a single random model realization. The red line serves as a guide for the eye for zero $\mathrm{MSE}$ difference. (b) Training curves for a random biased model realization. (c) Training curves for a random unbiased model realization. In (b)-(c), the full model is in blue and the cutoff model in orange, and the shading corresponds to the spread over $30$ training instances. 
    For these plots, $N=1$, $\Omega=\{1,...,6\}$ ($d=13$), $\tilde{\Omega}=\{1,2\}$ ($\tilde{d}=5$), $M=5$, $R=4$, $\mathfrak{n}_{\mathrm{train}}=25$ with a batch size of $5$.}
    \label{figapp:MSE_BiasUnbias_wTrain_full_Dloc5} 
\end{figure*}

\begin{figure*}
    \centering
    \includegraphics[width=\linewidth]{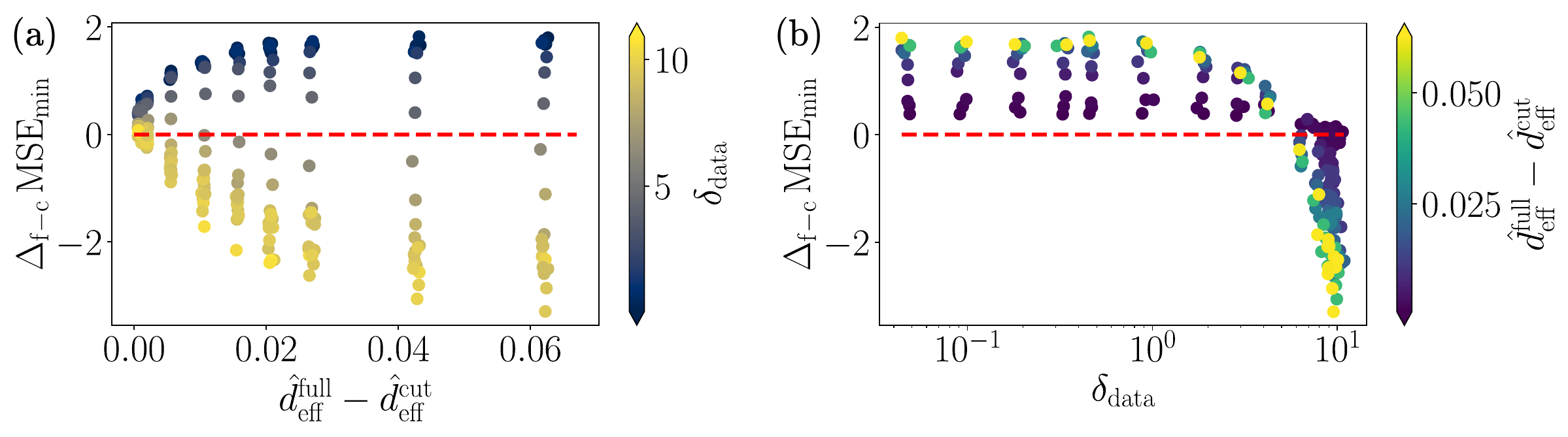}
    \caption{(a) $\Delta_{\mathrm{f-c}}\mathrm{MSE}_{\mathrm{min}}$ for different values of the difference $\hat{d}_{\mathrm{eff}}^{\mathrm{full}}-\hat{d}_{\mathrm{eff}}^{\mathrm{cut}}$. Each point corresponds to $\Delta_{\mathrm{f-c}}\mathrm{MSE}_{\mathrm{min}}$ averaged over $30$ training instances for $30$ random model realization. (b) Same as panel (a) but resolved as a function of $\delta_{\mathrm{data}}$. The red line serves as a guide for the eye for zero $\mathrm{MSE}$ difference. 
    For these plots, $N=1$, $\Omega=\{1,...,6\}$ ($d=13$), $\tilde{\Omega}=\{1,2\}$ ($\tilde{d}=5$), $M=5$, $R=4$, $\mathfrak{n}_{\mathrm{train}}=25$ with a batch size of $5$.}
    \label{figapp:MSE_AllBiases_full_Dloc5} 
\end{figure*}

Overall, the additional result presented here confirm and corroborate the findings in the main text. The results for models with full (i.e., non-tensorized) structure constants are shown in Figs.~\ref{figapp:MSE_BiasUnbias_wTrain_full_Dloc2}, \ref{figapp:MSE_AllBiases_full_Dloc2}, \ref{figapp:MSE_BiasUnbias_wTrain_full_Dloc5} and \ref{figapp:MSE_AllBiases_full_Dloc5}. Qualitatively, the behavior of $\Delta_{\mathrm{f-c}}\mathrm{MSE}_{\mathrm{min}}$ as a function of the ED difference $\hat{d}_{\mathrm{eff}}^{\mathrm{full}}-\hat{d}_{\mathrm{eff}}^{\mathrm{cut}}$ is the same and consistent with what presented in the main text irrespectively of the model specification (see Figs.~\ref{figapp:MSE_BiasUnbias_wTrain_full_Dloc2} and \ref{figapp:MSE_AllBiases_full_Dloc2} for the case of $\tilde{d}=2$, i.e., $\tilde{\mathcal{B}}_m=\{\cos\theta_m,\sin\theta_m\}$, and Figs.~\ref{figapp:MSE_BiasUnbias_wTrain_full_Dloc5} and \ref{figapp:MSE_AllBiases_full_Dloc5} for the case $\tilde{d}=5$, i.e., $\tilde{\mathcal{B}}_m=\{1,\cos\theta_m,\cos2\theta_m,\sin\theta_m,\sin2\theta_m\}$).

\begin{figure*}
    \centering
    \includegraphics[width=\linewidth]{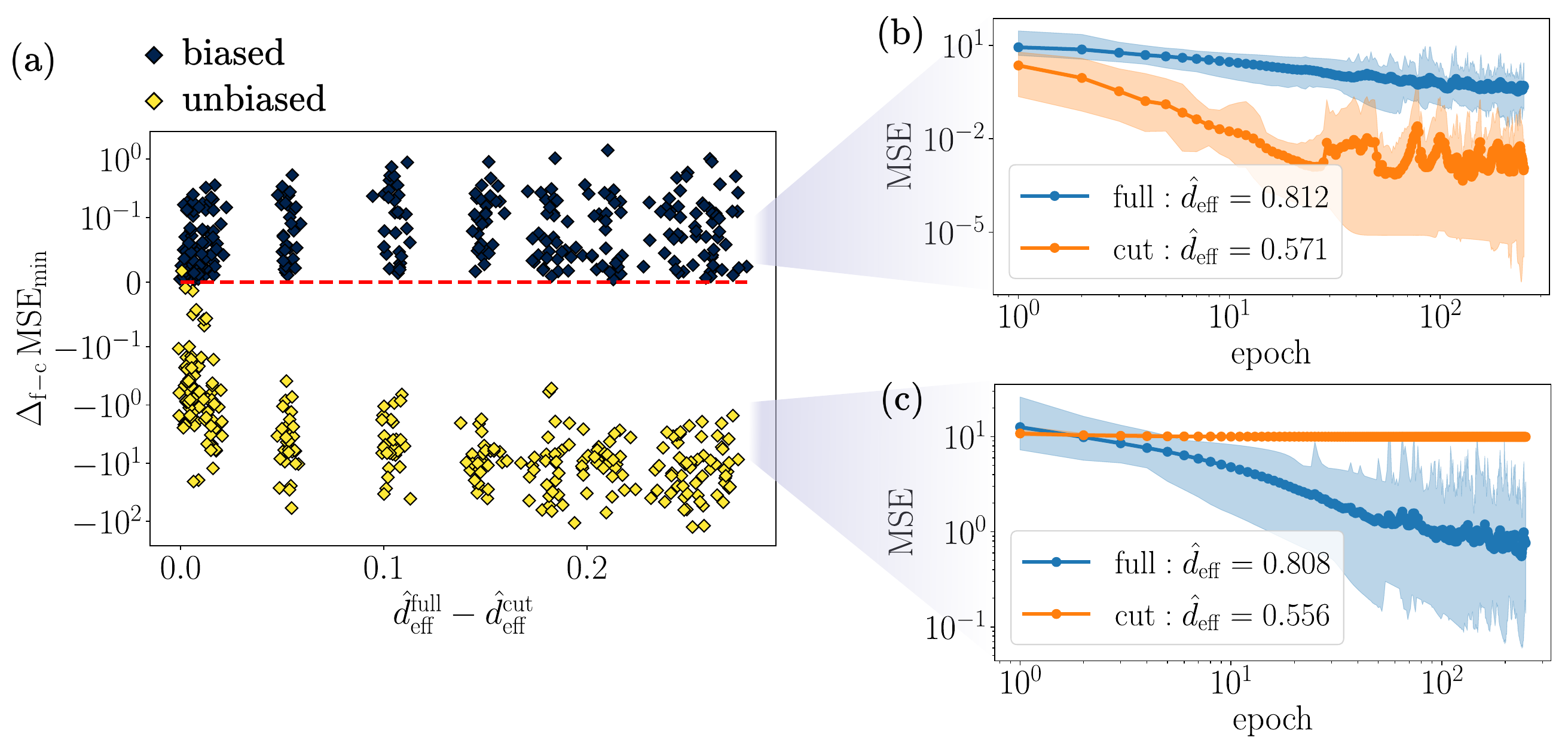}
    \caption{(a) $\Delta_{\mathrm{f-c}}\mathrm{MSE}_{\mathrm{min}}$ for different values of the difference $\hat{d}_{\mathrm{eff}}^{\mathrm{full}}-\hat{d}_{\mathrm{eff}}^{\mathrm{cut}}$. Each point corresponds to $\Delta_{\mathrm{f-c}}\mathrm{MSE}_{\mathrm{min}}$ averaged over $30$ training instances, for a single random model realization, i.e., a random right-normalized tensor train representing $V$. The red line serves as a guide for the eye for zero $\mathrm{MSE}$ difference. (b) Training curves for a random biased model realization. (c) Training curves for a random unbiased model realization. In (b)-(c), the full model is in blue and the cutoff model in orange, and the shading corresponds to the spread over $30$ training instances. 
    For these plots, $N=1$, $\Omega=\{1,...,13\}$ ($d=27$), $\tilde{\Omega}=\{1,2\}$ ($\tilde{d}=5$), $M=36$, $R=4$, $\chi=60$, $\mathfrak{n}_{\mathrm{train}}=30$ with a batch size of $5$.}
    \label{figapp:MSE_BiasUnbias_wTrain_TN_Dloc5} 
\end{figure*}

\begin{figure*}
    \centering
    \includegraphics[width=\linewidth]{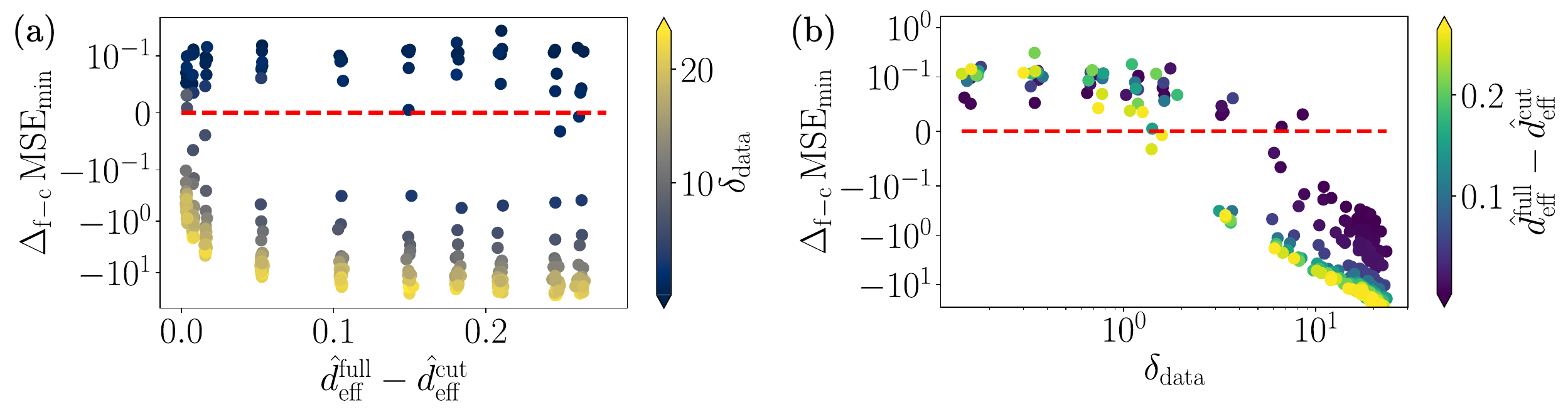}
    \caption{(a) $\Delta_{\mathrm{f-c}}\mathrm{MSE}_{\mathrm{min}}$ for different values of the difference $\hat{d}_{\mathrm{eff}}^{\mathrm{full}}-\hat{d}_{\mathrm{eff}}^{\mathrm{cut}}$. Each point corresponds to $\Delta_{\mathrm{f-c}}\mathrm{MSE}_{\mathrm{min}}$ averaged over $30$ training instances for $30$ random model realizations, i.e., random right-normalized tensor trains representing $V$. (b) Same as panel (a) but resolved as a function of $\delta_{\mathrm{data}}$. The red line serves as a guide for the eye for zero $\mathrm{MSE}$ difference. 
    For these plots, $N=1$, $\Omega=\{1,...,13\}$ ($d=27$), $\tilde{\Omega}=\{1,2\}$ ($\tilde{d}=5$), $M=36$, $R=4$, $\chi=60$, $\mathfrak{n}_{\mathrm{train}}=30$ with a batch size of $5$.}
    \label{figapp:MSE_AllBiases_TN_Dloc5} 
\end{figure*}

\begin{figure*}
    \centering
    \includegraphics[width=\linewidth]{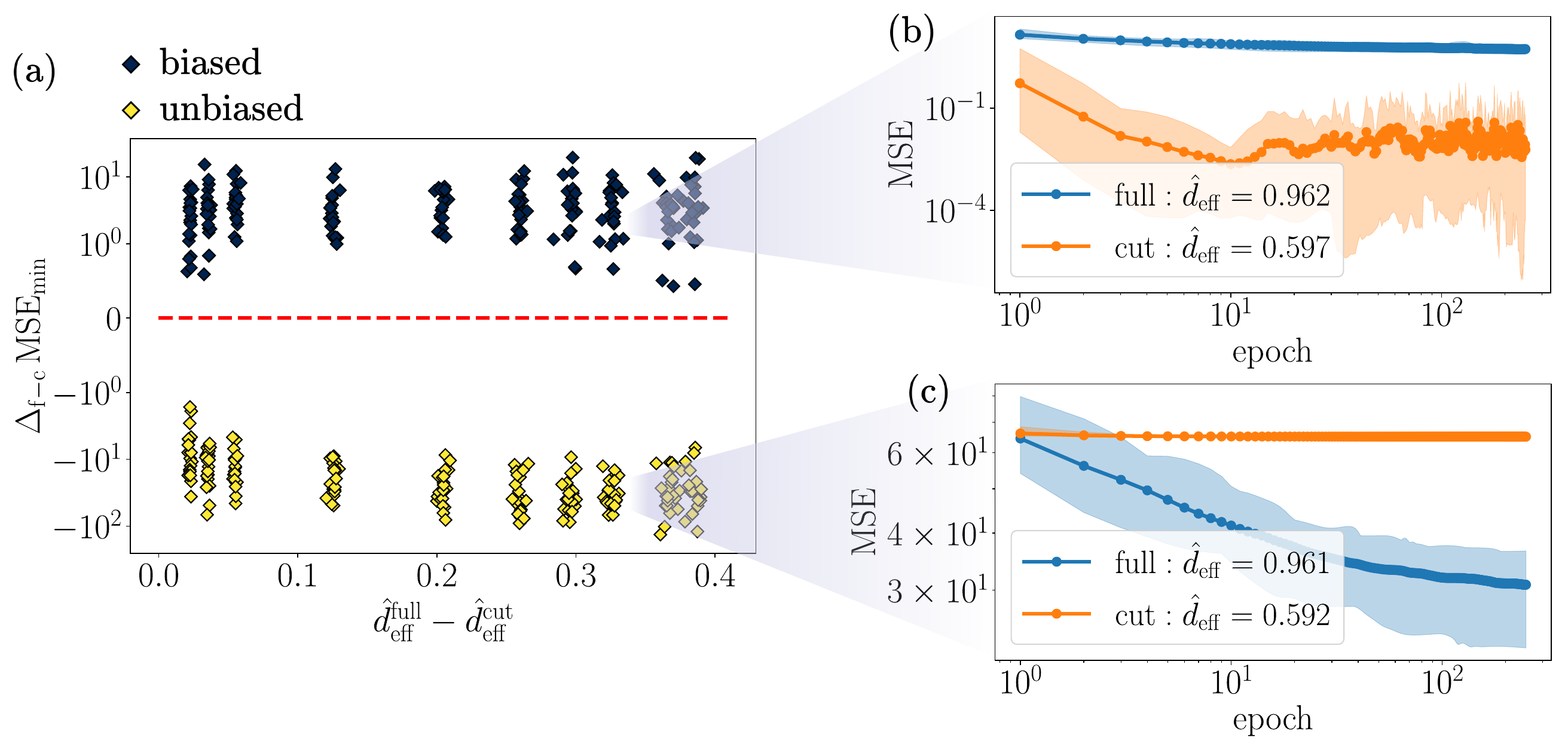}
    \caption{(a) $\Delta_{\mathrm{f-c}}\mathrm{MSE}_{\mathrm{min}}$ for different values of the difference $\hat{d}_{\mathrm{eff}}^{\mathrm{full}}-\hat{d}_{\mathrm{eff}}^{\mathrm{cut}}$. Each point corresponds to $\Delta_{\mathrm{f-c}}\mathrm{MSE}_{\mathrm{min}}$ averaged over $30$ training instances, for a single random model realization, i.e., a random right-normalized tensor train representing $V$. The red line serves as a guide for the eye for zero $\mathrm{MSE}$ difference. (b) Training curves for a random biased model realization. (c) Training curves for a random unbiased model realization. In (b)-(c), the full model is in blue and the cutoff model in orange, and the shading corresponds to the spread over $30$ training instances. 
    For these plots, $N=2$, $\Omega=\{1,...,5\}$ ($d=11$), $\tilde{\Omega}=\{1\}$ ($\tilde{d}=3$), $M=32$, $R=7$, $\chi=100$, $\mathfrak{n}_{\mathrm{train}}=225$ with a batch size of $5$.}
    \label{figapp:MSE_BiasUnbias_wTrain_TN_2D_Dloc3} 
\end{figure*}

The corresponding results for tensorized models (where only $V$ is decomposed as a tensor train) are shown in Figs.~\ref{figapp:MSE_BiasUnbias_wTrain_TN_Dloc5} and \ref{figapp:MSE_AllBiases_TN_Dloc5} for $\tilde{d}=5$, while Fig.~\ref{figapp:MSE_BiasUnbias_wTrain_TN_2D_Dloc3} shows results for $\tilde{d}=3$ and two input features $N=2$. 




\end{document}